\documentclass[11pt]{article}

\usepackage[final]{acl}
\usepackage{times}
\usepackage{latexsym}

\usepackage[T1]{fontenc}
\usepackage[utf8]{inputenc}

\usepackage{microtype}

\usepackage{inconsolata}

\usepackage{graphicx}

\usepackage{amssymb}
\usepackage{amsmath}
\usepackage{algorithm}
\usepackage{algpseudocode}
\usepackage[dvipsnames]{xcolor}

\usepackage{fvextra}
\usepackage{booktabs}   % \toprule, \midrule, \bottomrule
\usepackage{arydshln}   % \hdashline, \addlinespace
\usepackage{tabularx}   % tabularx environment
\usepackage{multirow}   % \multirow
\usepackage{pifont}
\usepackage{float}
\usepackage{placeins}
\usepackage{enumitem}
\usepackage[table]{xcolor}

\newcommand{\tthink}[1]{\textcolor{RoyalBlue}{\texttt{#1}}}
\newcommand{\tsearch}[1]{\textcolor{Orange}{\texttt{#1}}}
\newcommand{\tinfo}[1]{\textcolor{ForestGreen}{\texttt{#1}}}
\newcommand{\tanswer}[1]{\textcolor{Red}{\texttt{#1}}}
\newcommand{\treason}[1]{\textcolor{Rhodamine}{\texttt{#1}}}
\newcommand{\tuser}[1]{\textcolor{Gray}{\texttt{#1}}}

\title{\textsc{SteerSearch}: Interactive Search Reasoning with Mid-Response Steering}

\author{
Udit Thakur$^{1,2}$\thanks{Work done at eBay, as a thesis project.} \:\:\:\:\:\:\:\:
Baohao Liao$^{1,2}$\thanks{Project lead, correspondence to: \texttt{\href{mailto:b.liao@uva.nl}{b.liao@uva.nl}}.} \:\:\:\:\:\:\:\:
Matteo Nulli$^{2}$ \:\:\:\:\:\:\:\:
Stefan Vasilev$^{2}$ \\
\textbf{Seyyed Hadi Hashemi$^{2}$} \:\:\:\:\:\:
\textbf{Shahram Khadivi$^{2}$} \:\:\:\:\:\:
\textbf{Christof Monz$^{1}$} \\
$^{1}$Language Technology Lab, University of Amsterdam \:\:\:\:\: $^{2}$eBay
}

\begin{document}
\maketitle

% \begin{abstract}
% Search-augmented language models are increasingly used for long-horizon tasks, yet most open training methods assume a fixed user goal throughout the search trajectory. We study \textit{mid-response steering}, where a user intervenes during reasoning or retrieval and redirects the model objective. We identify \textit{answer-target anchoring}, where fixed-query search agents retrieve evidence for the updated request but still answer an earlier question. We introduce \textsc{SteerSearch}, an RL framework that trains on search trajectories with injected user interventions, together with Contrastive Context Reward (CCR) that supervises target shifts after each intervention. We also construct \textsc{SteerSearch-Bench} from multi-hop QA datasets to evaluate this setting. Without CCR, \textsc{SteerSearch} improves average steering accuracy over Search-R1 from 11.1\% to 31.3\% on Qwen2.5-3B and from 36.6\% to 44.6\% on Qwen3-4B, while preserving standard search-reasoning performance. The gains also extend across larger Qwen2.5 models and natural multi-turn conversations.
% \end{abstract}

\begin{abstract}
Search-augmented language models increasingly serve as long-horizon research agents, yet most open training methods assume a fixed user goal throughout search. We study \textit{mid-response steering}, where users redirect a model during ongoing reasoning or retrieval. We identify \textit{answer-target anchoring}, a failure mode in which fixed-query search agents retrieve evidence for the updated request but still answer an earlier question. We introduce \textsc{SteerSearch}, an RL framework that trains on interactive search trajectories with injected interventions to enable context-preserving target reassignment, and \textsc{SteerSearch-Bench}, a benchmark adapted from multi-hop QA datasets. We also investigate Contrastive Context Reward (CCR), an auxiliary signal for aligning post-intervention reasoning with the updated goal. Without CCR, \textsc{SteerSearch} improves average steering accuracy over Search-R1 from 11.1\% to 31.3\% on Qwen2.5-3B and from 36.6\% to 44.6\% on Qwen3-4B, while preserving standard single-query performance. Gains also extend to larger Qwen2.5 models and natural multi-turn conversations.
\end{abstract}

\section{Introduction}
\label{sec:introduction}

\begin{figure*}[t]
    \centering
    \includegraphics[width=0.86\textwidth]{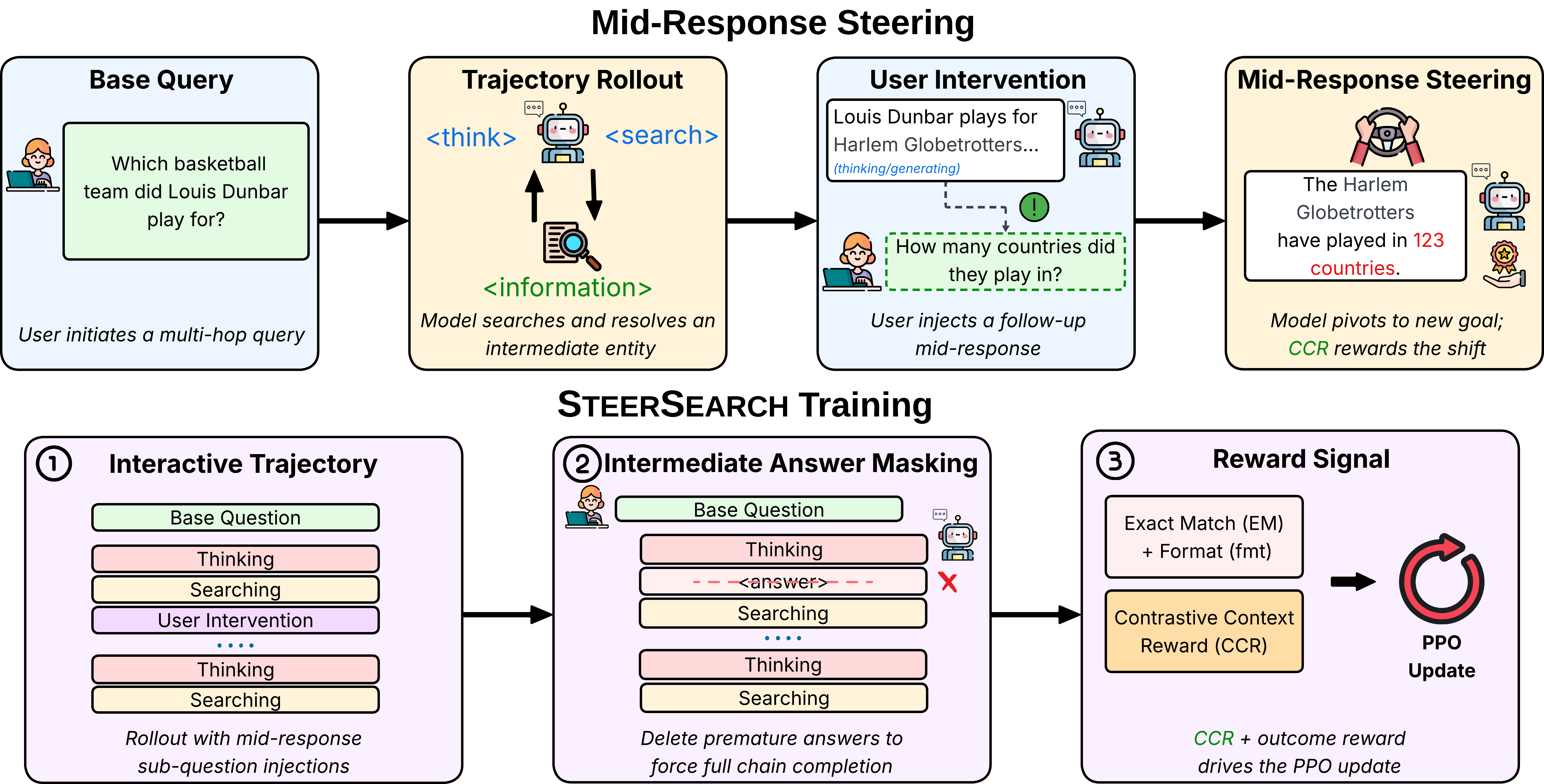}
    \vspace{-2mm}
    % \caption{
    %   \textbf{Overview of \textsc{SteerSearch}.} \textbf{Top}: During inference, the model receives a base query, conducts a search-and-reason trajectory, and is interrupted mid-response with an injected sub-question. It must pivot its answer target while preserving all prior retrieved evidence. \textbf{Bottom}: \textsc{SteerSearch} teaches the model with this behavior by (A) rolling out interactive trajectories with mid-response injections, (B) deleting premature answers in non-final phases to prevent shortcutting, and (C) rewarding the policy with standard outcome signals plus a Contrastive Context Reward (CCR) that measures whether the model's reasoning shifts toward the newly injected sub-question.
    %   }
    \caption{
    \textbf{Overview of \textsc{SteerSearch}.}
    \textbf{Top}: Given a base query, the model follows a search-and-reason trajectory before receiving a mid-response sub-question injection, requiring it to pivot its answer target while preserving retrieved evidence.
    \textbf{Bottom}: Training comprises (A) interactive rollouts with mid-response injections, (B) deletion of premature answers in non-final phases to prevent shortcutting, and (C) standard outcome rewards plus a Contrastive Context Reward (CCR) measuring whether reasoning shifts toward the injected sub-question.
    }
    \label{fig:overview}
    % \vspace{-4mm}
\end{figure*}

\begin{figure*}[t]
    \centering
    \includegraphics[width=0.86\textwidth]{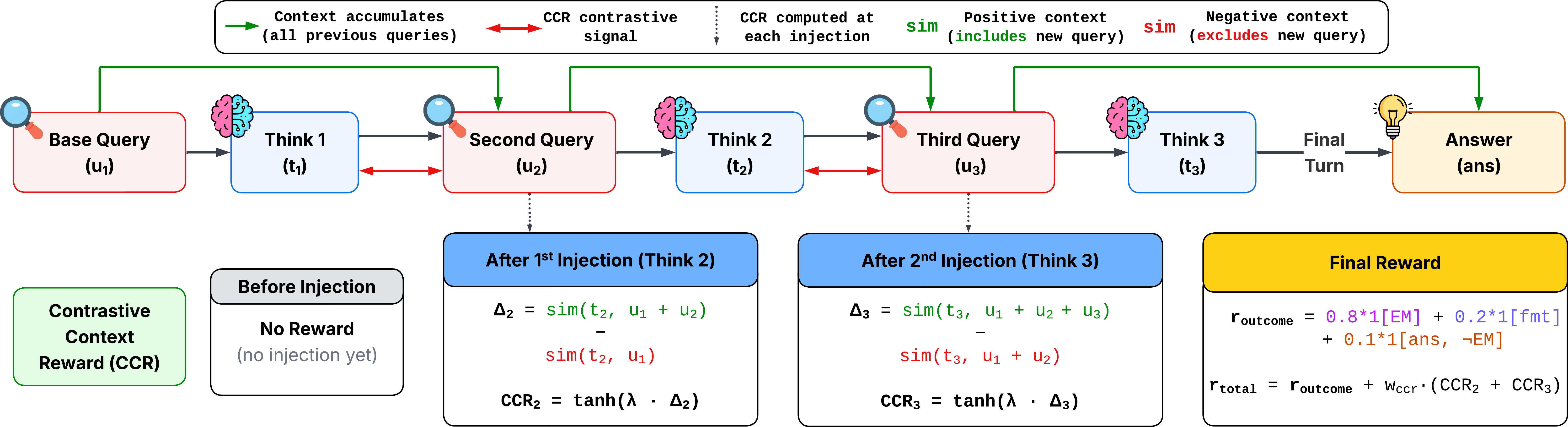}
    \vspace{-2mm}
    \caption{\textbf{Contrastive Context Reward (CCR).} After each mid-response injection, CCR contrasts the model's reasoning against the cumulative user-goal context before and after the intervention, producing an auxiliary signal for target reassignment.}
    \label{fig:ccr-diagram}
\end{figure*}

Search-augmented language models are increasingly being deployed as long-horizon research agents rather than one-shot question answerers \cite{openai2025deepresearchsystemcard, anthropic2025multiagentresearch, google2025geminideepresearch, zheng2025deepresearcherscalingdeepresearch, wu-etal-2025-webwalker, miromind2026mirothinker}. Systems such as Deep Research \cite{openai2025deepresearchsystemcard} plan, search, and synthesize information over multiple steps, shifting search-augmented reasoning toward extended interactive workflows. Recent agentic interfaces \cite{anthropic_claude_code, openai_codex} allow users to interrupt long-running responses and refine the request without restarting or losing progress. This emerging interaction pattern introduces a new requirement for search agents: they must continue from accumulated evidence while adapting to user instructions that arrive after reasoning or retrieval has already begun.

However, most open search-reasoning training methods still assume a fixed-goal trajectory \cite{jin2025searchr1trainingllmsreason, chen2025researchlearningreasonsearch, zheng2025deepresearcherscalingdeepresearch, wu-etal-2025-webwalker, li2026openresearcher, wu2025mmsearchr1incentivizinglmmssearch}. Methods such as Search-R1 \cite{jin2025searchr1trainingllmsreason} and ReSearch \cite{chen2025researchlearningreasonsearch} train language models to interleave reasoning with search, allowing retrieved evidence to influence subsequent generations, but the user goal is typically fixed at the start of the rollout. Even when multiple searches are allowed, the model is optimized to answer the original question rather than to handle a new user turn that changes the operative target during generation. This mismatch gives rise to a distinct failure mode for interactive research agents, which we call \textit{answer-target anchoring}. When a model trained under a fixed-goal trajectory is placed in an interrupted interaction setting, it may retrieve evidence that is sufficient for the user's updated request, but still produce an answer to an earlier question because its reasoning state remains anchored to the original target. For example, after identifying the Harlem Globetrotters as the team Louis Dunbar played for, a user may interrupt with a dependent follow-up asking how many countries that team has played in. Answering the follow-up requires preserving the intermediate entity from the first phase, but also reassigning the answer target from ``which team'' to ``how many countries?''. Standard search-reasoning RL provides little direct supervision for this kind of context-preserving goal shift.

In this work, we introduce \textsc{SteerSearch}, a reinforcement learning framework for mid-response steering in search-augmented reasoning, as illustrated in Figure~\ref{fig:overview}. Rather than optimizing only uninterrupted fixed-goal trajectories, \textsc{SteerSearch} trains models on interactive trajectories where user interventions are injected during ongoing search and reasoning. The model must preserve useful evidence from earlier steps while reassigning its answer target to the latest user goal. To support this behavior, \textsc{SteerSearch} combines randomized intervention timing, intermediate-answer masking, and Contrastive Context Reward (CCR), a new technique encouraging the model's reasoning to pivot toward newly injected user goals (Figure~\ref{fig:ccr-diagram}).

To evaluate this setting, we construct \textsc{SteerSearch-Bench}, a mid-response steering benchmark adapted from state of the art multi-hop benchmarks~\cite{yang-etal-2018-hotpotqa, ho-etal-2020-constructing, trivedi-etal-2022-musique, press2023measuringnarrowingcompositionalitygap}, where each original question is converted into a sequence of injected user interventions. Experiments on two instruction-tuned models show that \textsc{SteerSearch} substantially improves mid-response steering on average \(\sim14\) percentage points over multi-hop, while preserving single-query search reasoning performance. CCR further improves steering accuracy, with gains varying across model and reward-model configurations. We additionally demonstrate generalization across model scales, with \textsc{SteerSearch} consistently outperforming Search-R1 on the 7B and 14B variants of Qwen2.5, as well as across model families on Gemma3-4B.

Our contributions are as follows:
% \begin{itemize}[nosep, leftmargin=*]
\begin{itemize}[nosep, leftmargin=*, itemsep=1.5pt]
    \item We formulate \textit{mid-response steering} for search-augmented reasoning and identify \textit{answer-target anchoring}, a failure mode where fixed-goal search agents retrieve evidence for an updated request but still answer an earlier question.
    \item We propose \textsc{SteerSearch}, an RL framework that trains models on interactive search trajectories with injected user interventions, encouraging context-preserving target reassignment during generation. We also introduce CCR, an auxiliary reward for aligning post-intervention reasoning with the updated user goal.
    \item We introduce \textsc{SteerSearch-Bench}, a benchmark for evaluating whether search agents can follow mid-response user interventions while reusing accumulated evidence.
    \item We show that \textsc{SteerSearch} improves average accuracy on \textsc{SteerSearch-Bench} by 8.0--20.2 percentage points over Search-R1, while preserving performance on standard search-reasoning benchmarks.
\end{itemize}

\begin{figure*}[t]
% \begin{figure}[t]
  \centering
  \includegraphics[width=0.8\textwidth]{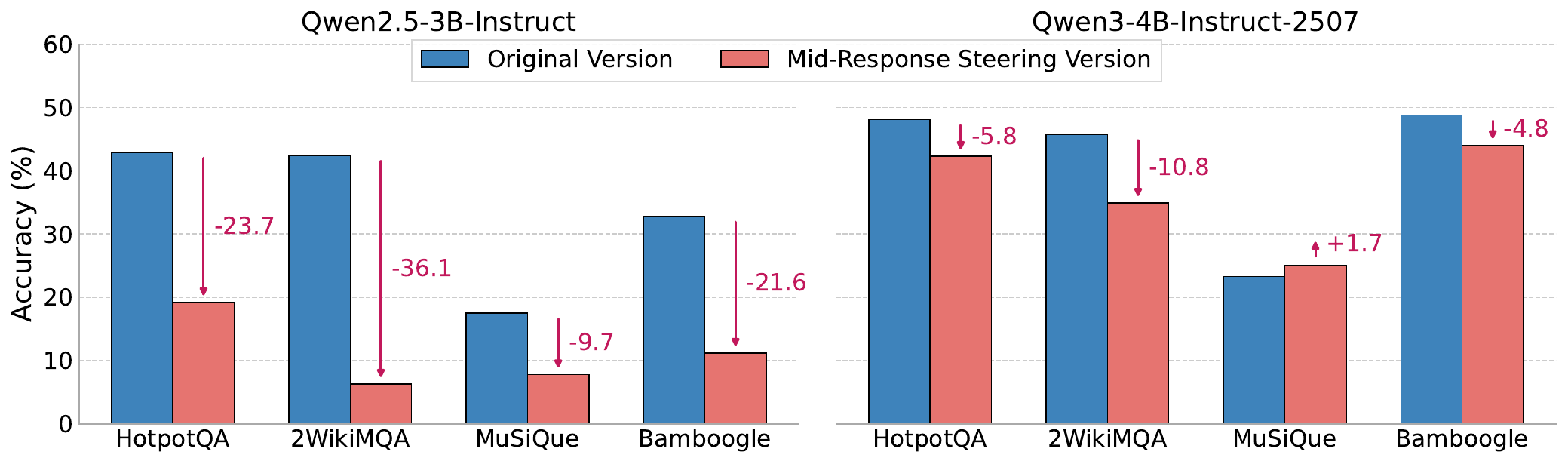}
  \vspace{-3mm}
  \caption{\textbf{LLMs performance drop on mid-response steering.} Performance of LLMs trained only with single-query search on the original multi-hop questions and their mid-response steering variants. The steering variants decompose each original single query into a sequence of injected user queries, simulating user's interruptions.}
  \label{fig:performance_drop}
  % \vspace{-4mm}
% \end{figure}
\end{figure*}
\section{Related Work}

\textbf{Search-augmented reasoning with reinforcement learning.}
Recent work trains language models to interleave reasoning with external search. Search-R1~\cite{jin2025searchr1trainingllmsreason} and ReSearch~\cite{chen2025researchlearningreasonsearch} formulate retrieval as part of the model's action space, allowing the policy to issue search queries, read returned evidence, and continue reasoning before producing an answer. Follow-up work shows that search-reasoning agents are sensitive to reward design, initialization, and retrieval environments~\cite{jin2025empiricalstudyreinforcementlearning}, while related RL-based approaches extend search reasoning to multimodal search and long-horizon web research~\cite{wu2025mmsearchr1incentivizinglmmssearch,narayan2025deepmmsearchr1empoweringmultimodalllms,zheng2025deepresearcherscalingdeepresearch,li2026openresearcher,liao2026self}.
However, these methods typically optimize uninterrupted trajectories toward a fixed user goal, whereas \textsc{SteerSearch} studies search-reasoning after a new user turn arrives mid-response.

\textbf{Multi-hop reasoning and query decomposition.}
Multi-hop question answering benchmarks evaluate whether models can combine evidence across multiple facts or documents. HotpotQA~\cite{yang-etal-2018-hotpotqa}, 2WikiMultiHopQA~\cite{ho-etal-2020-constructing}, and MuSiQue~\cite{trivedi-etal-2022-musique} construct questions whose answers require linked reasoning steps, often with supporting evidence or structured reasoning paths. Prompting methods such as Self-Ask~\cite{press2023measuringnarrowingcompositionalitygap} encourage models to decompose a question into intermediate sub-questions, and recent training approaches such as ParallelSearch~\cite{zhao2025parallelsearchtrainllmsdecompose} learn to decompose complex queries into sub-queries that can be searched independently.
Unlike static query decomposition, \textsc{SteerSearch} treats sub-questions as user interventions injected into an ongoing trajectory, requiring the model to incorporate the new goal without discarding prior context.

\textbf{Interactive and long-horizon research agents.}
Recent systems move search-augmented language models toward long-horizon research workflows. Deployed systems such as ChatGPT Deep Research, Gemini Deep Research, and Claude Research browse, inspect sources, and synthesize findings over extended interactions~\cite{openai2025deepresearchsystemcard,google2025geminideepresearch,anthropic2025multiagentresearch}. Web-agent benchmarks further evaluate models' ability to navigate, gather information, and solve tasks across multiple steps~\cite{liao2026multi,liao2025reward,mialon2024gaia,wei2025browsecomp,zhou2025browsecomp}. Recent ChatGPT interfaces also expose mid-response course adjustment, where users can refine a request while a long-running response is still being produced~\cite{openai2026gpt54}.

Together, these systems and benchmarks motivate a training problem not captured by standard search-reasoning evaluations: agents must remain responsive to user steering after reasoning and retrieval have already begun.

\section{Preliminaries}
\label{sec:preliminary}

\subsection{Mid-Response Steering}
\label{sec:mid_response_steering}

We study search-augmented reasoning where the user may intervene while the model is still reasoning or retrieving. Unlike standard single-query search, where the answer target remains fixed throughout the trajectory, mid-response steering allows a new user goal to arrive after partial reasoning and retrieval have already occurred. The model must therefore adapt to the latest goal while preserving useful information from the existing trajectory.

Let $u=\langle u_1,\ldots,u_K\rangle$ denote an ordered sequence of user goals, where $u_1$ is the initial query and $u_k$ for $k>1$ is a mid-response intervention. After receiving $u_k$, the policy $\pi_\theta$ produces model moves $m_{k,t}$, which may contain reasoning, a search query, or an answer. A search query produces a retriever observation $o_{k,t}=R(m_{k,t})$. The resulting interactive trajectory is

\[
\tau =
\Big\langle
u_k,\{(m_{k,t},o_{k,t})\}_{t=1}^{T_k}
\Big\rangle_{k=1}^{K}.
\]

For $k<K$, the next goal $u_{k+1}$ may be injected before the model finishes responding to $u_k$, after reasoning or retrieval has already modified the trajectory. User interventions and retrieved observations are supplied by the environment, while model moves are generated by the policy.

The objective is to answer the terminal user goal while conditioning on the accumulated interaction history:
\[
\max_{\pi_\theta}
\mathbb{E}_{(u,a^\ast)\sim\mathcal{D},\;
\tau\sim\pi_\theta(\cdot\mid u,R)}
\left[
r(a^\ast,\tau)
\right],
\]

where $a^\ast$ is the gold answer to the final goal $u_K$. Mid-response steering therefore requires \textit{context-preserving target reassignment}: after receiving $u_{k+1}$, the model should reuse relevant evidence from earlier phases while treating the new intervention as the operative answer target.

\subsection{Answer-Target Anchoring}
\label{sec:answer_target_anchoring}

A natural question is whether models trained only for single-query search can directly handle mid-response steering. We evaluate a Search-R1-style model~\cite{jin2025searchr1trainingllmsreason} on both the original multi-hop questions and their mid-response steering variants. As shown in Figure~\ref{fig:performance_drop}, performance drops substantially when the same reasoning problem is presented through sequential user interventions, indicating that strong fixed-query search ability does not directly transfer to a changing answer target.

\begin{figure*}[t]
  \centering
  \includegraphics[width=0.9\textwidth]{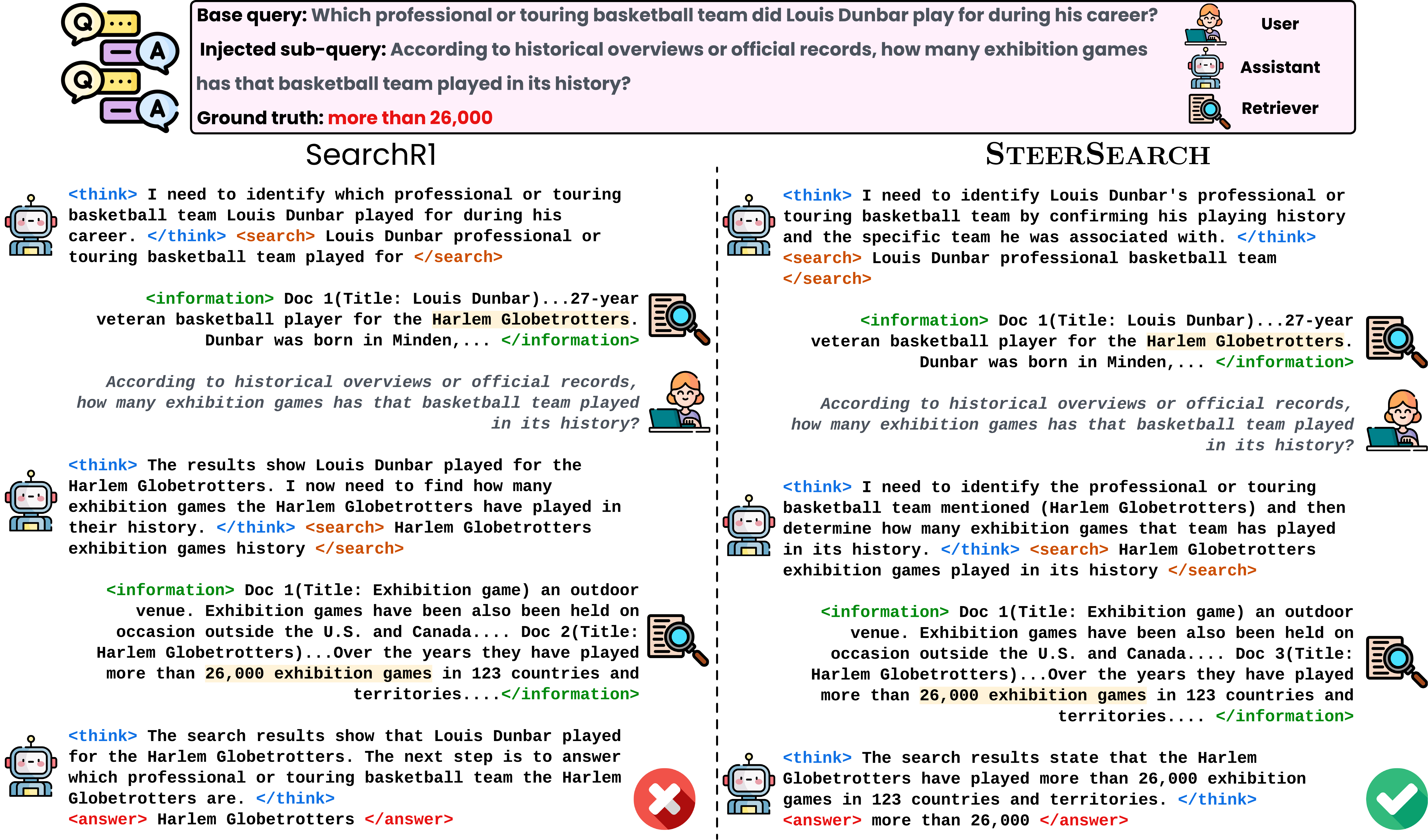}
  \vspace{-2mm}
  % \caption{\textbf{Comparison of Search-R1 and \textsc{SteerSearch} on the same mid-response steering trajectory.} Both retrieve evidence containing the updated answer, but Search-R1 remains anchored to the base question and answers with the intermediate entity. \textsc{SteerSearch} treats the injection as operative goal and produces the correct answer.}
  \caption{\textbf{Search-R1 vs. \textsc{SteerSearch} on the same steering trajectory.} Both retrieve relevant evidence, but Search-R1 remains anchored to the base question, while \textsc{SteerSearch} follows the injected query correctly.}
  \label{fig:trajectory-comparison}
\end{figure*}

We identify a common failure mode behind this degradation, which we call \textit{answer-target anchoring}. As illustrated in Figure~\ref{fig:trajectory-comparison}, the model may respond locally to a new user query by reasoning about it and retrieving the required evidence, yet still produce a final answer corresponding to an earlier goal. The failure is therefore distinct from retrieval error: the relevant evidence may already be present, while the model fails to update which user goal controls the final response.

This phenomenon is related to long-context robustness issues such as \textit{lost-in-the-middle}~\cite{liu2024lost, chen-etal-2024-sifo, liao2025lost}, but differs in the source of the failure. Mid-response steering requires the model not only to use information from a growing context, but to reassign the operative answer target after a user interruption while preserving useful prior evidence. This motivates training directly on trajectories in which the target changes during generation.
\section{\textsc{SteerSearch}}
\label{sec:steersearch}

Motivated by answer-target anchoring (Section~\ref{sec:answer_target_anchoring}) in single-query search models under user intervention, we present \textsc{SteerSearch}, a reinforcement learning framework for context-preserving target reassignment under mid-response steering.

\subsection{Overview}
\label{subsec:steersearch-overview}

\textsc{SteerSearch} trains the policy on the interactive trajectory formulation defined in Section~\ref{sec:mid_response_steering}, exposing it during training to the same type of target changes encountered when a user redirects an ongoing search. Its key departures from standard fixed-query search training are: (1) \textit{randomized phase budgets}, which vary intervention timing; (2) \textit{intermediate answer removal}, which prevents premature answers from anchoring subsequent reasoning; and (3) a \textit{Contrastive Context Reward} (CCR), which provides direct supervision for post-intervention target shifts.

A rollout begins with the initial user goal and alternates between model reasoning and retrieval. Before the terminal phase, the next user goal is injected after a randomized interaction budget while the accumulated reasoning, search queries, and retrieved evidence remain available. If the model produces an answer in a non-final phase, the premature answer is masked before continuing to the next intervention while useful preceding reasoning and retrieval are retained. The model must therefore continue from the existing context while changing which user goal controls its subsequent reasoning and final answer.

This differs from Search-R1~\cite{jin2025searchr1trainingllmsreason}, where a single answer target remains fixed throughout the rollout. Rather than prescribing a new reasoning strategy, \textsc{SteerSearch} changes the interaction experienced during RL so that target reassignment becomes part of the learned policy. Figure~\ref{fig:overview} summarizes the framework, while the complete rollout mechanics and configuration are provided in Section~\ref{sec:rollout-config} and Algorithm~\ref{alg:steersearch}.

\subsection{Contrastive Context Reward}
\label{sec:reward}

The terminal outcome reward indicates whether the final answer is correct but does not directly supervise whether the model updates its reasoning when a new user goal is introduced. As Figure~\ref{fig:trajectory-comparison} illustrates, a model can retrieve evidence for an intervention while remaining anchored to an earlier answer target. CCR provides an auxiliary signal that measures whether post-intervention reasoning moves toward the updated user context (Figure~\ref{fig:ccr-diagram}).

Let $\phi(\cdot)$ be a frozen text encoder producing normalized embeddings, with cosine similarity
\[
\mathrm{sim}(x,y)=\phi(x)^\top\phi(y).
\]
For an intervention $u_{k+1}$, let $H_{k+1}$ denote the reasoning blocks scored after the intervention. We compare each $h\in H_{k+1}$ against a positive context $p_{k+1}$ containing the updated user goal and a negative context $n_k$ representing the context before that intervention:
\[
\Delta_{k+1}(h)
=
\mathrm{sim}(h,p_{k+1})
-
\mathrm{sim}(h,n_k).
\]
Positive values indicate greater alignment with the updated goal, while negative values indicate continued alignment with the preceding context.

The auxiliary reward for an intervention phase is
\[
r^{\mathrm{ccr}}_{k+1}
=
\tanh\!\left(
\lambda
\frac{1}{|H_{k+1}|}
\sum_{h\in H_{k+1}}
\Delta_{k+1}(h)
\right),
\]
where $\lambda$ controls the sharpness of the signal. CCR augments the terminal outcome reward as
\[
r_{\mathrm{total}}(\tau,a^\ast)
=
r(\tau,a^\ast)
+
w_{\mathrm{ccr}}
\sum_{k=1}^{K-1}
r^{\mathrm{ccr}}_{k+1},
\]
where $w_{\mathrm{ccr}}$ controls its contribution.

Our main CCR configuration uses cumulative user context, terminal reward assignment, and \texttt{Qwen3-Embedding-0.6B}~\cite{zhang2025qwen3embeddingadvancingtext} as the frozen similarity encoder. We additionally evaluate alternative context, reasoning, assignment, and encoder choices, including E5~\cite{wang2024textembeddingsweaklysupervisedcontrastive}. We also compare CCR with IGPO~\cite{wang2026informationgainbasedpolicyoptimization}, a dense reward based on marginal gains in ground-truth confidence. IGPO assumes a fixed answer target, whereas under mid-response steering the operative goal may change after each intervention. Full configuration details and ablations are provided in Appendix~\ref{app:appendix_ccr}.
\begin{table*}[h]
\centering
\small
\resizebox{\textwidth}{!}{
\begin{tabular}{lccccccccc}
\toprule
\multirow{3}{*}{\textbf{Method}} & \multicolumn{3}{c}{\textbf{General QA}} & \multicolumn{4}{c}{\textbf{Multi-Hop QA}} & \multirow{3}{*}{\textbf{Avg.}} \\
\cmidrule(lr){2-4}\cmidrule(lr){5-8}
& \textbf{NQ$^\dagger$} & \textbf{TriviaQA$^*$} & \textbf{PopQA$^*$} & \textbf{HotpotQA$^\dagger$} & \textbf{2WikiMQA$^*$} & \textbf{MuSiQue$^*$} & \textbf{Bamboogle$^*$} & \\
\midrule
\multicolumn{9}{l}{\textit{Qwen2.5-3B-Instruct}} \\
Vanilla & 0.216 & 0.415 & 0.259 & 0.188 & 0.175 & 0.044 & 0.168 & 0.209 \\
CoT & 0.023 & 0.032 & 0.005 & 0.021 & 0.021 & 0.002 & 0.000 & 0.015 \\
IRCoT & 0.111 & 0.312 & 0.200 & 0.164 & 0.171 & 0.067 & 0.240 & 0.181 \\
Search-o1 & 0.238 & 0.472 & 0.262 & 0.221 & 0.218 & 0.054 & 0.320 & 0.255 \\
RAG & 0.348 & 0.544 & 0.387 & 0.255 & 0.226 & 0.047 & 0.080 & 0.270 \\
SFT & 0.249 & 0.292 & 0.104 & 0.186 & 0.248 & 0.044 & 0.112 & 0.176 \\
R1-instruct & 0.210 & 0.449 & 0.171 & 0.208 & 0.275 & 0.060 & 0.192 & 0.224 \\
Rejection Sampling & 0.294 & 0.488 & 0.332 & 0.240 & 0.233 & 0.059 & 0.210 & 0.265 \\
Search-R1 & 0.447 & 0.602 & 0.433 & 0.429 & 0.424 & 0.175 & 0.328 & 0.405 \\
\rowcolor{orange!8}
\textsc{SteerSearch} & 0.450 & 0.615 & 0.442 & \textbf{0.442} & 0.409 & \textbf{0.209} & \textbf{0.384} & \textbf{0.422} \\
\rowcolor{blue!8}
\textsc{SteerSearch}+CCR \textit{(QwenEmb)} & \textbf{0.456} & \textbf{0.621} & \textbf{0.447} & 0.437 & \textbf{0.426} & 0.165 & 0.328 & 0.411 \\
\midrule
\multicolumn{9}{l}{\textit{Qwen3-4B-Instruct-2507}} \\
Vanilla & 0.283 & 0.497 & 0.317 & 0.244 & 0.270 & 0.060 & 0.376 & 0.292 \\
Search-R1 & \textbf{0.496} & 0.660 & \textbf{0.487} & \textbf{0.481} & 0.457 & 0.233 & 0.488 & 0.472 \\
\rowcolor{orange!8}
\textsc{SteerSearch} & 0.491 & 0.652 & 0.479 & 0.475 & \textbf{0.498} & \textbf{0.239} & 0.480 & 0.473 \\
\rowcolor{blue!8}
\textsc{SteerSearch}+CCR \textit{(QwenEmb)} & 0.487 & \textbf{0.664} & 0.480 & 0.480 & \textbf{0.498} & 0.220 & \textbf{0.512} & \textbf{0.477} \\
\bottomrule
\end{tabular}
}
% \caption{\textbf{Exact match on the Standard Benchmark.} $\dagger$~in-domain, $*$~out-of-domain, \textbf{Bold} = best per model among fine-tuned methods, and Vanilla = no additional training. \textsc{SteerSearch} uses only the terminal outcome reward. \textsc{SteerSearch}+CCR \textit{(QwenEmb)} uses cumulative-context CCR with terminal assignment and \texttt{Qwen3-Embedding-0.6B} as the frozen CCR similarity encoder; document retrieval remains unchanged.}
\vspace{-3mm}
\caption{\textbf{Exact match on the Standard Benchmark.}
$\dagger$~in-domain, $*$~out-of-domain, \textbf{Bold} = best per model among fine-tuned methods.
Vanilla = no additional training.
\textsc{SteerSearch} uses terminal outcome reward, while \textsc{SteerSearch}+CCR applies the contrastive context reward on top of it.}

\label{tab:searchr1_bench}
\end{table*}
\section{Experiments}
\label{sec:experiments}

% This section evaluates whether training on mid-response steering trajectories improves interactive search reasoning while preserving standard retrieval-augmented question answering. We compare \textsc{SteerSearch} with Search-R1 under matched training and evaluation settings, and further evaluate generalization across natural multi-turn conversations, model sizes, and model families. Detailed experimental setup and implementation details are provided in Appendix~\ref{sec:setup}.

\textbf{Models and Baselines.}
Our primary experiments use Qwen2.5-3B-Instruct~\cite{qwen2025qwen25technicalreport} and Qwen3-4B-Instruct-2507~\cite{yang2025qwen3technicalreport}. We additionally evaluate Qwen2.5-7B-Instruct, Qwen2.5-14B-Instruct, and Gemma3-4B-Instruct~\cite{gemmateam2025gemma3technicalreport} in the text-only setting to study model generalization. Our primary baseline is Search-R1~\cite{jin2025searchr1trainingllmsreason}, which trains the same base models for fixed-query iterative reasoning and retrieval but does not include mid-response interventions. Search-R1 and \textsc{SteerSearch} use the same retriever, PPO objective, reward format, and primary training budget, isolating the effect of intervention-based training. 
% We also report the original instruction-tuned model without additional SFT or RL training as Vanilla where applicable.

\textbf{Training.}
Both methods are initialized from supervised fine-tuning on 4K GPT-5.4-generated HotpotQA trajectories, followed by PPO training. The primary experiments use 40K RL samples. Search-R1 is trained only on fixed-query trajectories, while \textsc{SteerSearch} uses a mixture of fixed-query and multi-phase trajectories containing injected sub-questions. The \textsc{SteerSearch} trajectories are constructed from multi-hop questions and expose the model to decomposition, hint steering, and disambiguation interventions. Unless otherwise stated, all primary results use the SFT-4K/PPO-40K configuration. Detailed training data composition, rollout configuration, and optimization settings are provided in Appendix~\ref{sec:setup}.

\textbf{Evaluation.}
We evaluate two complementary settings. The Standard Benchmark measures single-query retrieval-augmented question answering across seven datasets and tests whether steering training preserves ordinary search-reasoning ability. \textsc{SteerSearch-Bench} evaluates the target interactive setting by injecting sub-questions during an ongoing search trajectory and scoring only the final answer. We report exact match (EM) as the primary metric, following~\citet{jin2025searchr1trainingllmsreason} while also reporting semantic evaluation using GPT-5.5 as a judge. We additionally evaluate \textsc{SteerSearch} on natural multi-turn conversations from TopiOCQA \citet{adlakha-etal-2022-topiocqa} to test generalization beyond synthetically generated interventions. Benchmark construction and evaluation details are provided in Appendix~\ref{sec:benchmarks}.

\subsection{Main Results}
\label{subsec:main_results}

\begin{table*}[!t]
\centering\small
\begin{tabular}{lcccccc}
\toprule
\textbf{Method} & \textbf{HotpotQA$^\dagger$} & \textbf{2WikiMQA$^*$} & \textbf{MuSiQue$^*$} & \textbf{Bamboogle$^*$} & \textbf{\textbf{Avg.}} \\
\midrule
\multicolumn{6}{l}{\textit{Qwen2.5-3B-Instruct}} \\
Vanilla & 0.101 & 0.054 & 0.039 & 0.080 & 0.068 \\
Search-R1 & 0.192 & 0.063 & 0.078 & 0.112 & 0.111 \\
\rowcolor{orange!8}
\textsc{SteerSearch} & 0.376 & 0.260 & 0.241 & 0.376 & 0.313 \\
\rowcolor{blue!8}
\textsc{SteerSearch}+CCR \textit{(QwenEmb)} & \textbf{0.435} & \textbf{0.391} & \textbf{0.264} & \textbf{0.496} & \textbf{0.396} \\
\midrule
\multicolumn{6}{l}{\textit{Qwen3-4B-Instruct-2507}} \\
Vanilla & 0.171 & 0.173 & 0.068 & 0.096 & 0.127 \\
Search-R1 & 0.423 & 0.349 & 0.250 & 0.440 & 0.366 \\
\rowcolor{orange!8}
\textsc{SteerSearch} & 0.472 & 0.457 & 0.305 & \textbf{0.552} & 0.446 \\
\rowcolor{blue!8}
\textsc{SteerSearch}+CCR \textit{(QwenEmb)} & \textbf{0.477} & \textbf{0.474} & \textbf{0.308} & 0.544 & \textbf{0.451} \\
\bottomrule
\end{tabular}
\vspace{-3mm}
% \caption{\textbf{Exact match on \textsc{SteerSearch-Bench}} with mid-response sub-question injection. $\dagger$~in-domain, $*$~out-of-domain, \textbf{Bold} = best per model among fine-tuned methods, and Vanilla = no additional training. \textsc{SteerSearch}+CCR \textit{(QwenEmb)} uses cumulative-context CCR with terminal assignment and \texttt{Qwen3-Embedding-0.6B} as the frozen CCR similarity encoder; document retrieval remains unchanged.}
\caption{\textbf{Performance on \textsc{SteerSearch-Bench}}, a new benchmark for mid-response steering. $\dagger$~in-domain, $*$~out-of-domain, \textbf{Bold} = best per model among fine-tuned methods, and Vanilla = no additional training.}
\label{tab:steer_bench}
\end{table*}

\textbf{Standard Benchmark.}
Table~\ref{tab:searchr1_bench} shows that training on mid-response steering trajectories does not degrade standard single-query search-reasoning ability. This is important because \textsc{SteerSearch} changes the training distribution substantially: instead of optimizing only uninterrupted question-answering trajectories, the model is exposed to injected sub-questions, multi-phase rollouts, and intermediate answer masking. For Qwen2.5, the gains are not restricted to HotpotQA, which supplies the SFT trajectories, but extend to several out-of-domain datasets. The curriculum therefore remains compatible with ordinary iterative retrieval outside the training distribution. For Qwen3, Search-R1 and \textsc{SteerSearch} obtain nearly identical average performance, making preservation of standard search behavior the relevant result rather than improvement on this setting.

\textbf{Mid-Response Steering.}
Table~\ref{tab:steer_bench} shows a substantially different pattern on \textsc{SteerSearch-Bench}. Search-R1 improves over Vanilla but transfers only partially to interrupted interactions. For Qwen2.5, average EM increases from 6.8\% for Vanilla to 11.1\% for Search-R1, while \textsc{SteerSearch} reaches 31.3\%. The improvement is consistent across all datasets despite differences in their multi-hop structure. Qwen3 begins from a considerably stronger Search-R1 baseline, reaching 36.6\% average EM, but \textsc{SteerSearch} still improves performance to 44.6\%. Ordinary search training therefore provides some ability to continue reasoning after an intervention, particularly for the stronger model, but does not reliably train the model to change which user request controls the final answer. The remaining gap separates target reassignment from the retrieval and multi-hop reasoning capabilities already learned by Search-R1. GPT-5.5 based semantic evaluation preserves the same separation between Search-R1 and \textsc{SteerSearch}, indicating that the gains are not driven by exact-match sensitivity to answer surface form (Table~\ref{tab:semantic_eval}).

\textbf{Natural Multi-Turn Generalization.}
To test whether the learned steering behavior extends beyond synthetically generated intervention chains, we evaluate the same Search-R1 and \textsc{SteerSearch} checkpoints without CCR on human-authored TopiOCQA conversations. We apply the same interaction protocol by injecting each subsequent user turn during the ongoing search trajectory. As shown in Appendix~\ref{subsec:topiocqa} (Table~\ref{tab:topiocqa}), \textsc{SteerSearch} improves over Search-R1 for both models across EM, token-level F1, and GPT-5.5 semantic judging. These gains provide external evidence that target reassignment is not restricted to the \textsc{SteerSearch-Bench} generation pipeline.

\textbf{Contrastive Context Reward.}
The main tables report CCR with terminal assignment and \texttt{Qwen3-Embedding-0.6B} as the frozen CCR similarity encoder. Among the E5-based variants, cumulative-terminal CCR performs best overall and achieves the strongest average performance on \textsc{SteerSearch-Bench}. Using Qwen3-Embedding further improves steering performance for both models, while its effect on the Standard Benchmark is mixed. Full configuration details, encoder comparisons, and the IGPO~\cite{wang2026informationgainbasedpolicyoptimization} baseline are provided in Appendix~\ref{app:appendix_ccr} and Tables~\ref{tab:ccr_ablation_ss}--\ref{tab:ccr_ablation_sr1}.

\vspace{-2mm}
\begin{figure}[h]
\centering
\includegraphics[width=\columnwidth]{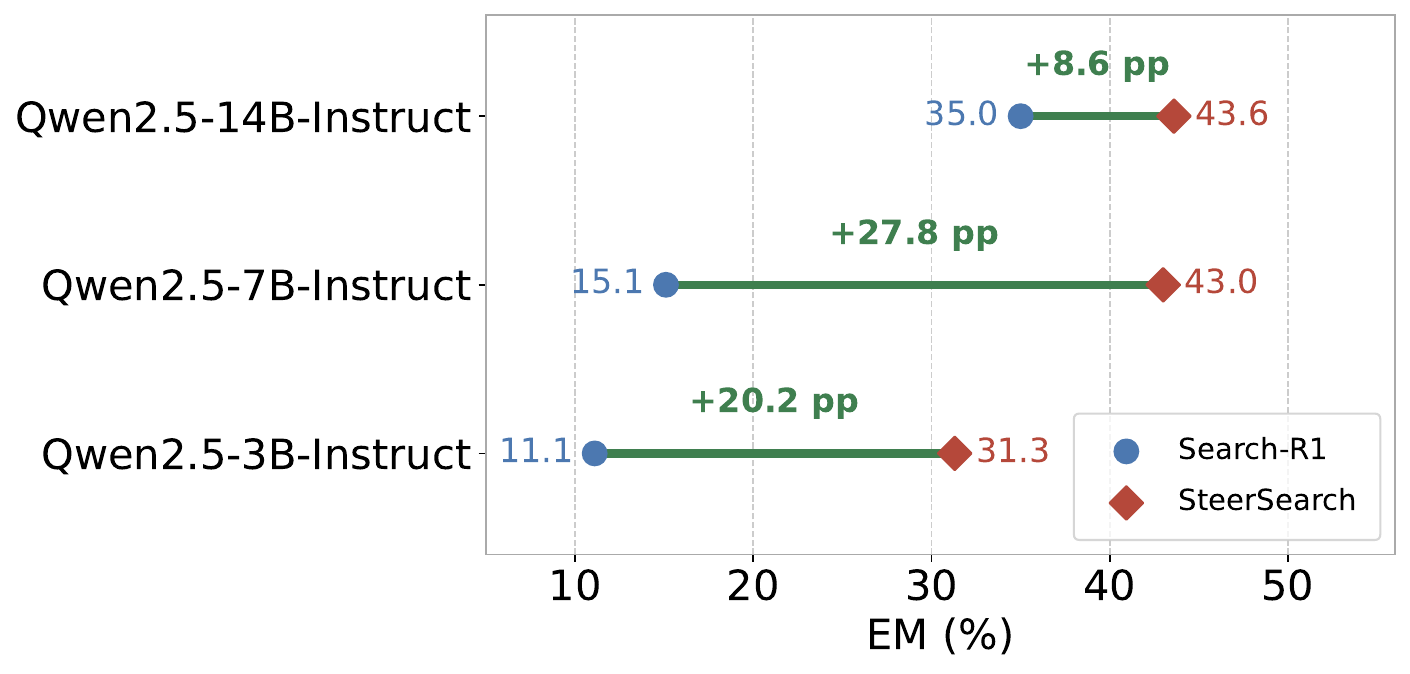}
\vspace{-8mm}
\caption{\textbf{\textsc{SteerSearch} vs.\ Search-R1} on \textsc{SteerSearch-Bench} across Qwen2.5 model sizes.}
\label{fig:model-scaling}
\vspace{-3mm}
\end{figure}

\textbf{Model Generalization.}
We evaluate \textsc{SteerSearch} across larger Qwen2.5 models and use Gemma3-4B-Instruct as a cross-family check. Figure~\ref{fig:model-scaling} shows that \textsc{SteerSearch} consistently improves \textsc{SteerSearch-Bench} performance over Search-R1 across Qwen2.5 model sizes, although the gap narrows at 14B as the Search-R1 baseline becomes stronger. The corresponding Standard Benchmark results show that these gains remain concentrated in the interrupted interaction setting. Gemma3-4B-Instruct reinforces this result beyond the Qwen family: \textsc{SteerSearch} improves over Search-R1 on both evaluation settings, while CCR provides a further substantial improvement, yielding the strongest results in both settings. Detailed per-dataset results and the CCR comparison are reported in Appendix~\ref{subsec:cross-family-generalization}, Tables~\ref{tab:gemma_ss_bench} and~\ref{tab:gemma_sr1_bench}.

\begin{table*}[h]
\centering\small
\setlength{\tabcolsep}{2pt}
\resizebox{\textwidth}{!}{
\begin{tabular}{lcccccccccccc}
\toprule
\textbf{Dataset} & \multicolumn{6}{c}{\textbf{\textit{Qwen2.5-3B-Instruct}}} & \multicolumn{6}{c}{\textbf{\textit{Qwen3-4B-Instruct-2507}}} \\
\cmidrule(lr){2-7}\cmidrule(lr){8-13}
 & \multicolumn{2}{c}{\textbf{\textsc{SteerSearch-Bench}}} & \multicolumn{2}{c}{\textbf{First Sub-Question}} & \multicolumn{2}{c}{\textbf{Concatenated}} & \multicolumn{2}{c}{\textbf{\textsc{SteerSearch-Bench}}} & \multicolumn{2}{c}{\textbf{First Sub-Question}} & \multicolumn{2}{c}{\textbf{Concatenated}} \\
\cmidrule(lr){2-3}\cmidrule(lr){4-5}\cmidrule(lr){6-7}\cmidrule(lr){8-9}\cmidrule(lr){10-11}\cmidrule(lr){12-13}
 & \textbf{Search-R1} & \textbf{\textsc{SteerSearch}} & \textbf{Search-R1} & \textbf{\textsc{SteerSearch}} & \textbf{Search-R1} & \textbf{\textsc{SteerSearch}} & \textbf{Search-R1} & \textbf{\textsc{SteerSearch}} & \textbf{Search-R1} & \textbf{\textsc{SteerSearch}} & \textbf{Search-R1} & \textbf{\textsc{SteerSearch}} \\
\midrule
HotpotQA$^\dagger$ & 0.192 & \cellcolor{orange!8}\textbf{0.376} & 0.512\hspace{3pt}{\scriptsize\textcolor{ForestGreen}{$\uparrow$32.0 pp}} & \cellcolor{orange!8}\textbf{0.564}\hspace{3pt}{\scriptsize\textcolor{ForestGreen}{$\uparrow$18.9 pp}} & 0.267\hspace{3pt}{\scriptsize\textcolor{ForestGreen}{$\uparrow$7.5 pp}} & \cellcolor{orange!8}\textbf{0.374}\hspace{3pt}{\scriptsize\textcolor{BrickRed}{$\downarrow$0.2 pp}} & 0.423 & \cellcolor{blue!8}\textbf{0.472} & \cellcolor{blue!8}\textbf{0.616}\hspace{3pt}{\scriptsize\textcolor{ForestGreen}{$\uparrow$19.3 pp}} & 0.610\hspace{3pt}{\scriptsize\textcolor{ForestGreen}{$\uparrow$13.9 pp}} & 0.426\hspace{3pt}{\scriptsize\textcolor{ForestGreen}{$\uparrow$0.3 pp}} & \cellcolor{blue!8}\textbf{0.448}\hspace{3pt}{\scriptsize\textcolor{BrickRed}{$\downarrow$2.4 pp}} \\
2WikiMQA$^*$ & 0.063 & \cellcolor{orange!8}\textbf{0.260} & \cellcolor{orange!8}\textbf{0.516}\hspace{3pt}{\scriptsize\textcolor{ForestGreen}{$\uparrow$45.3 pp}} & \cellcolor{orange!8}\textbf{0.516}\hspace{3pt}{\scriptsize\textcolor{ForestGreen}{$\uparrow$25.5 pp}} & 0.179\hspace{3pt}{\scriptsize\textcolor{ForestGreen}{$\uparrow$11.6 pp}} & \cellcolor{orange!8}\textbf{0.284}\hspace{3pt}{\scriptsize\textcolor{ForestGreen}{$\uparrow$2.4 pp}} & 0.349 & \cellcolor{blue!8}\textbf{0.457} & 0.554\hspace{3pt}{\scriptsize\textcolor{ForestGreen}{$\uparrow$20.5 pp}} & \cellcolor{blue!8}\textbf{0.564}\hspace{3pt}{\scriptsize\textcolor{ForestGreen}{$\uparrow$10.7 pp}} & 0.382\hspace{3pt}{\scriptsize\textcolor{ForestGreen}{$\uparrow$3.3 pp}} & \cellcolor{blue!8}\textbf{0.402}\hspace{3pt}{\scriptsize\textcolor{BrickRed}{$\downarrow$5.5 pp}} \\
MuSiQue$^*$ & 0.078 & \cellcolor{orange!8}\textbf{0.241} & 0.564\hspace{3pt}{\scriptsize\textcolor{ForestGreen}{$\uparrow$48.7 pp}} & \cellcolor{orange!8}\textbf{0.586}\hspace{3pt}{\scriptsize\textcolor{ForestGreen}{$\uparrow$34.5 pp}} & 0.100\hspace{3pt}{\scriptsize\textcolor{ForestGreen}{$\uparrow$2.2 pp}} & \cellcolor{orange!8}\textbf{0.206}\hspace{3pt}{\scriptsize\textcolor{BrickRed}{$\downarrow$3.5 pp}} & 0.250 & \cellcolor{blue!8}\textbf{0.305} & 0.622\hspace{3pt}{\scriptsize\textcolor{ForestGreen}{$\uparrow$37.1 pp}} & \cellcolor{blue!8}\textbf{0.631}\hspace{3pt}{\scriptsize\textcolor{ForestGreen}{$\uparrow$32.5 pp}} & 0.197\hspace{3pt}{\scriptsize\textcolor{BrickRed}{$\downarrow$5.4 pp}} & \cellcolor{blue!8}\textbf{0.228}\hspace{3pt}{\scriptsize\textcolor{BrickRed}{$\downarrow$7.7 pp}} \\
Bamboogle$^*$ & 0.112 & \cellcolor{orange!8}\textbf{0.376} & 0.624\hspace{3pt}{\scriptsize\textcolor{ForestGreen}{$\uparrow$51.2 pp}} & \cellcolor{orange!8}\textbf{0.688}\hspace{3pt}{\scriptsize\textcolor{ForestGreen}{$\uparrow$31.2 pp}} & 0.112\hspace{3pt}{\scriptsize\textcolor{Gray}{-- 0.0 pp}} & \cellcolor{orange!8}\textbf{0.360}\hspace{3pt}{\scriptsize\textcolor{BrickRed}{$\downarrow$1.6 pp}} & 0.440 & \cellcolor{blue!8}\textbf{0.552} & 0.688\hspace{3pt}{\scriptsize\textcolor{ForestGreen}{$\uparrow$24.8 pp}} & \cellcolor{blue!8}\textbf{0.720}\hspace{3pt}{\scriptsize\textcolor{ForestGreen}{$\uparrow$16.8 pp}} & 0.368\hspace{3pt}{\scriptsize\textcolor{BrickRed}{$\downarrow$7.2 pp}} & \cellcolor{blue!8}\textbf{0.472}\hspace{3pt}{\scriptsize\textcolor{BrickRed}{$\downarrow$8.0 pp}} \\
\midrule
\textbf{Avg.} & 0.111 & \cellcolor{orange!8}\textbf{0.313} & 0.554\hspace{3pt}{\scriptsize\textcolor{ForestGreen}{$\uparrow$44.3 pp}} & \cellcolor{orange!8}\textbf{0.588}\hspace{3pt}{\scriptsize\textcolor{ForestGreen}{$\uparrow$27.5 pp}} & 0.165\hspace{3pt}{\scriptsize\textcolor{ForestGreen}{$\uparrow$5.3 pp}} & \cellcolor{orange!8}\textbf{0.306}\hspace{3pt}{\scriptsize\textcolor{BrickRed}{$\downarrow$0.7 pp}} & 0.366 & \cellcolor{blue!8}\textbf{0.446} & 0.620\hspace{3pt}{\scriptsize\textcolor{ForestGreen}{$\uparrow$25.4 pp}} & \cellcolor{blue!8}\textbf{0.631}\hspace{3pt}{\scriptsize\textcolor{ForestGreen}{$\uparrow$18.5 pp}} & 0.343\hspace{3pt}{\scriptsize\textcolor{BrickRed}{$\downarrow$2.2 pp}} & \cellcolor{blue!8}\textbf{0.387}\hspace{3pt}{\scriptsize\textcolor{BrickRed}{$\downarrow$5.9 pp}} \\
\bottomrule
\end{tabular}
}
\vspace{-2mm}
% \caption{\textbf{Component analysis on \textsc{SteerSearch-Bench}.} First Sub-Question = only the first sub-question evaluated, no injection; Concatenated = all sub-questions presented upfront. $\dagger$~in-domain, $*$~out-of-domain. \cellcolor{orange!8}Orange and \cellcolor{blue!8}blue shading indicate the better result between \textsc{SteerSearch} and Search-R1 within each condition for Qwen2.5-3B-Instruct and Qwen3-4B-Instruct-2507 respectively.}
\caption{\textbf{Component analysis on \textsc{SteerSearch-Bench}.} First Sub-Question evaluates only the first sub-question without injection; Concatenated presents all sub-questions upfront. Arrows indicate absolute accuracy changes in percentage points (pp) relative to \textsc{SteerSearch-Bench}. $\dagger$ in-domain, $^*$ out-of-domain, highest values in \textbf{bold}.}
\label{tab:component_analysis}
\end{table*}
\subsection{Analysis}
\label{subsec:analysis}

\textbf{Attribution of Steering Gains.}
Table~\ref{tab:component_analysis} separates target reassignment from ordinary QA difficulty. When evaluated on only the first sub-question without injection, the gap between Search-R1 and \textsc{SteerSearch} narrows substantially for both models, suggesting that both can answer simpler component questions when no target switch is required. In contrast, concatenating all sub-questions upfront yields only modest changes relative to the benchmark setting and does not consistently improve Search-R1. Although the performance gap narrows, \textsc{SteerSearch} retains an average advantage for both models, indicating that access to the decomposed questions alone does not reproduce the benefits of sequential steering.

We further train a matched no-injection Search-R1 control using the same decomposed 40K PPO pool as \textsc{SteerSearch}, but with each chain concatenated upfront rather than injected during generation. \textsc{SteerSearch} retains substantial gains under this stronger control, supporting an effect from injected-trajectory training beyond multi-hop content exposure (Appendix~\ref{subsec:matched-no-injection}).

\textbf{Intermediate Context.}
Retaining stale reasoning traces causes relatively small degradation, whereas retaining completed intermediate answers is more harmful, particularly for Qwen3 (Table~\ref{tab:intermediate-answer-ablation}). This is consistent with answer-target anchoring: completed answers provide a stronger representation of a superseded objective than unfinished reasoning traces. Detailed ablations are provided in Appendix~\ref{subsec:intervention-depth}.

\textbf{Intervention Depth.}
Both methods become less accurate as the number of injected sub-questions increases, but the \textsc{SteerSearch} advantage persists across interaction depths (Figure~\ref{fig:nsubq-accuracy}). Longer chains therefore remain difficult, but the gains are not confined to a single target switch (Appendix~\ref{subsec:intervention-depth}).

\textbf{CCR and Post-Intervention Alignment.}
We test whether CCR changes the post-intervention behavior it is designed to supervise using outputs from cumulative-terminal CCR with \texttt{Qwen3-Embedding-0.6B} and the corresponding no-CCR checkpoints. For each intervention, we represent post-intervention behavior using the final reasoning block in that phase together with the first search query, when present, and compare it with the cumulative user context before and after the new turn using the same frozen encoder. A positive similarity difference denotes a \textit{pivot} toward the updated context; otherwise the event is treated as an \textit{anchor}.

\begin{figure}[!t]
\centering
\includegraphics[width=\columnwidth]{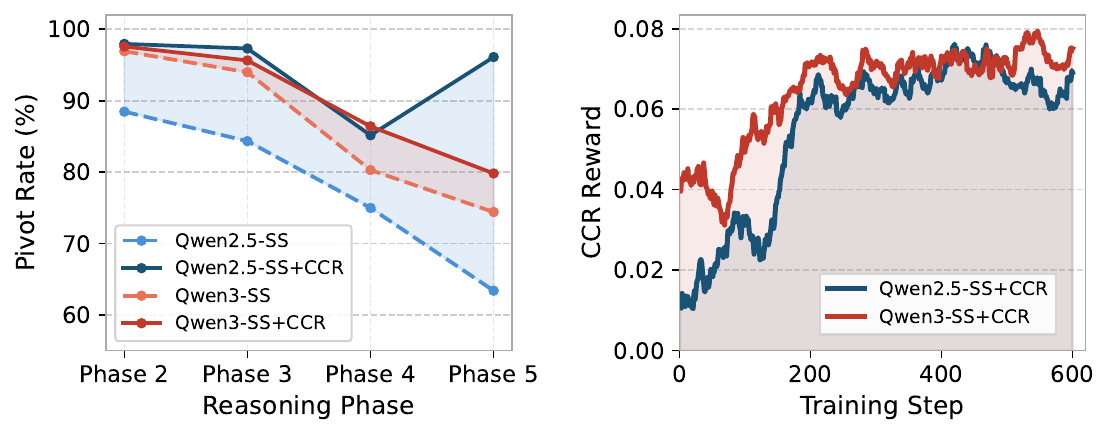}
\vspace{-8mm}
% \caption{Post-intervention pivot alignment across reasoning phases for \textsc{SteerSearch} with and without CCR. A pivot indicates that post-intervention behavior is more similar to the updated interaction context than to the context before the new user turn.}
\caption{\textbf{Pivot alignment} across reasoning phases (left) and \textbf{CCR reward} during training (right).}
\label{fig:ccr-pivot}
\vspace{-4mm}
\end{figure}

Figure~\ref{fig:ccr-pivot} shows that CCR reduces the overall anchor rate from 15.3\% to 3.6\% for Qwen2.5 and from 6.5\% to 4.8\% for Qwen3, with larger differences at later interventions. This provides behavioral evidence that CCR shifts post-intervention reasoning toward the updated user goal according to the semantic alignment signal it explicitly optimizes. The smaller change for Qwen3 is consistent with its already high pivot rate without CCR. 
% The pivot measure verifies CCR's intended alignment effect rather than explaining downstream answer accuracy directly.

\section{Conclusion}
\label{sec:conclusion}

We introduced \textsc{SteerSearch}, a reinforcement learning framework for mid-response steering in search-augmented reasoning. We showed that fixed-query search agents suffer from answer-target anchoring, where they retrieve evidence for a new user goal but still answer an earlier question. By training on interactive trajectories with injected user interventions, \textsc{SteerSearch} learns context-preserving target reassignment, substantially improving performance on \textsc{SteerSearch-Bench} while preserving standard search-reasoning ability.
\section*{Limitations}
\label{sec:limitations}

This work focuses on mid-response steering for text-based search-augmented reasoning. Our experiments use Wikipedia-style retrieval and multi-hop QA benchmarks adapted into steering trajectories, which provide controlled evaluation but do not cover every form of open-ended web research. Extending \textsc{SteerSearch} to noisier web environments, multimodal retrieval, and longer interactive sessions is a natural next step.

\textsc{SteerSearch} improves the model's ability to update its answer target after user interventions, but it does not solve all remaining reasoning errors. Some failures still require fine-grained role disambiguation, exact comparison, or synthesis across partially retrieved evidence. These limitations are orthogonal to the mid-response steering setting and can be addressed with stronger retrieval, verification, or answer validation modules.

\section*{Ethical Considerations}
\label{sec:ethics}

This work studies how search-augmented language models respond to user interventions during ongoing reasoning. The proposed training data and benchmark are derived from public QA datasets and Wikipedia-style retrieval, without collecting private user interactions. However, systems that support mid-response steering may be used in open-ended research settings where retrieved content can contain misinformation, bias, or sensitive information. Improving steering behavior therefore does not remove the need for source verification, citation, and safety filtering in deployed systems.

Because \textsc{SteerSearch} trains models to follow newly injected user goals, care should also be taken when applying it to unsafe or policy-violating requests. Future deployments should combine mid-response steering with existing safeguards for tool use, retrieval access, and refusal behavior.

% Bibliography entries for the entire Anthology, followed by custom entries
%\bibliography{custom,anthology-overleaf-1,anthology-overleaf-2}

% Custom bibliography entries only
\bibliography{custom}
\clearpage

\appendix

\section{Experimental Setup}
\label{sec:setup}

This chapter describes the models, data, training procedure, rollout environment, and evaluation benchmarks used in this work. The experimental design is built around a controlled comparison between Search-R1 and \textsc{SteerSearch}. Search-R1 trains search-augmented reasoning under a fixed user goal, while \textsc{SteerSearch} trains the same base models to handle user interventions injected during an ongoing trajectory. Unless otherwise stated, both methods use the same retriever, base model family, PPO implementation, and evaluation protocol.

\subsection{Models and Training Configuration}
\label{sec:sft}

We conduct experiments on four instruction-tuned models. The primary experiments use Qwen2.5-3B-Instruct~\cite{qwen2025qwen25technicalreport} and Qwen3-4B-Instruct-2507~\cite{yang2025qwen3technicalreport}. For brevity, we refer to Qwen3-4B-Instruct-2507 as Qwen3-4B-Instruct in the text; tables use the full model name. These two models are used for the main results, CCR analysis, and data-scaling experiments. To study model scaling, we additionally train Qwen2.5-7B-Instruct and Qwen2.5-14B-Instruct. The scaling models are evaluated separately in Section~\ref{subsec:appendix_model_scaling}.

Qwen3-4B-Instruct-2507 is used in non-thinking mode. Its built-in pre-reasoning behavior is disabled, and all reasoning is instead generated inside explicit \texttt{\treason{<reason>}} tags following the \textsc{SteerSearch} prompt format. Qwen2.5 models use the corresponding \texttt{\tthink{<think>}} format. All primary experiments use the SFT-4k/PPO-40k training budget. Smaller training budgets, SFT-2k/PPO-8k and SFT-4k/PPO-16k, are used only for the data-scaling ablation in Section~\ref{subsec:appendix_data_config}.

\subsection{SFT Initialization}

All RL runs are initialized from a supervised fine-tuned checkpoint. The SFT data contains 4,000 gold-standard trajectories generated by GPT-5.4~\cite{openai2026gpt54} over HotpotQA. The data is split equally between Search-R1-style fixed-query trajectories and \textsc{SteerSearch}-style trajectories with injected sub-questions. This mixed initialization gives both methods the same supervised starting point while exposing the model to both ordinary search reasoning and interrupted interaction formats.

The SFT objective is standard next-token prediction on assistant tokens only. Retrieved passages, user turns, and injected sub-question turns are masked from the loss because they are supplied by the environment rather than generated by the policy. For Qwen2.5-3B-Instruct and Qwen3-4B-Instruct-2507, SFT runs for up to 600 steps with batch size 32, learning rate $2 \times 10^{-6}$, cosine decay, and warmup ratio 0.025. The checkpoint with the best validation loss is used to initialize RL. For Qwen2.5-7B-Instruct and Qwen2.5-14B-Instruct, the SFT learning rate is reduced to $1 \times 10^{-6}$ and training runs for 300 steps.

\subsection{PPO Training}

Both Search-R1 and \textsc{SteerSearch} are optimized with PPO~\cite{schulman2017proximalpolicyoptimizationalgorithms} using the \texttt{verl} framework~\cite{sheng2024hybridflow}. PPO training uses a rollout batch size of 512, PPO mini-batch size of 256, and per-GPU micro-batch size of 8. The actor is optimized with AdamW at learning rate $1 \times 10^{-6}$ and warmup ratio 0.285. Rollouts are generated with vLLM using maximum response length 1,024 and temperature 1.0. Actor, critic, and reference models use FSDP2 with parameter and optimizer offloading. A KL penalty with coefficient 0.001 is applied against the frozen reference policy. Primary models are trained for 600 steps on 8 A100 GPUs and evaluated every 200 steps.

For the 7B and 14B scaling experiments, the actor learning rate is reduced to $5 \times 10^{-7}$ and RL training runs for 400 steps. These models are evaluated at step 400 because their reward curves plateau earlier than the 3B runs, indicating early convergence.

\subsection{Training Data}
\label{sec:data}

\begin{table}[t]
\centering\small
\resizebox{\columnwidth}{!}{
\begin{tabular}{lllcccc}
\toprule
\textbf{Phase} & \textbf{Method} & \textbf{Strategy} & \textbf{Samples} & \multicolumn{3}{c}{\textbf{Sub-questions}} \\
\cmidrule(lr){5-7}
 & & & & \textbf{Min} & \textbf{Mean} & \textbf{Max} \\
\midrule
\multirow{3}{*}{SFT}
  & \textsc{SteerSearch} & Gold trajectory & 2{,}000 & 2 & 2.87 & 4 \\
  & Search-R1            & Gold trajectory & 2{,}000 & \multicolumn{3}{c}{n/a} \\
\cmidrule(lr){2-7}
  & \multicolumn{2}{r}{\textit{\textbf{Total}}} & \textbf{4{,}000} & & & \\
\midrule
\multirow{6}{*}{\shortstack[l]{\textsc{SteerSearch} \\ (PPO)}}
  & \textsc{SteerSearch} & Decomposition  & 8{,}000 & 2 & 3.03 & 5 \\
  & \textsc{SteerSearch} & Hint steering  & 8{,}000 & 3 & 3.84 & 5 \\
  & \textsc{SteerSearch} & Disambiguation & 8{,}000 & 2 & 3.57 & 5 \\
  & Search-R1            & HotpotQA       & 8{,}000 & \multicolumn{3}{c}{n/a} \\
  & Search-R1            & NQ             & 8{,}000 & \multicolumn{3}{c}{n/a} \\
\cmidrule(lr){2-7}
  & \multicolumn{2}{r}{\textit{\textbf{Total}}} & \textbf{40{,}000} & & & \\
\midrule
\multirow{3}{*}{\shortstack[l]{Search-R1 \\ (PPO)}}
  & Search-R1 & HotpotQA & 20{,}000 & \multicolumn{3}{c}{n/a} \\
  & Search-R1 & NQ       & 20{,}000 & \multicolumn{3}{c}{n/a} \\
\cmidrule(lr){2-7}
  & \multicolumn{2}{r}{\textit{\textbf{Total}}} & \textbf{40{,}000} & & & \\
\bottomrule
\end{tabular}
}
\caption{Training data composition and sub-question chain length statistics. Search-R1 uses no sub-question injection (n/a).}
\label{tab:data-stats}
\end{table}

Table~\ref{tab:data-stats} summarizes the composition of the training data. The key distinction between Search-R1 and \textsc{SteerSearch} is not the base retriever or the PPO objective, but the interaction structure of the trajectories. Search-R1 receives fixed-query trajectories, while \textsc{SteerSearch} receives a mixture of fixed-query examples and multi-phase trajectories with injected user interventions.

\paragraph{\textsc{SteerSearch} sub-question generation.}
\label{sec:steersearch-data-gen}
Sub-questions for multi-hop training samples are generated from HotpotQA using GPT-5.4. We use three generation strategies. \emph{Decomposition} produces sequential sub-questions that narrow the problem through explicit constraints. \emph{Hint steering} produces softer directional interventions that redirect the search process without directly committing to an intermediate answer. \emph{Disambiguation} focuses on resolving ambiguous entities or references before factual retrieval. These strategies are designed to expose the model to different ways a user may steer an ongoing search trajectory.

Generated chains are filtered by a committee of three GPT-5.4 judge instances. A chain is discarded and regenerated if any judge rejects it. Rejection criteria include answer leakage, bare-noun keyword queries, bundled multi-part questions, malformed utterances, or missing verb structure. This filtering step is important because mid-response steering evaluation is only meaningful if the injected sub-questions guide retrieval without directly revealing the final answer.

\paragraph{SFT trajectories.}
SFT trajectories are generated by running GPT-5.4 with the \textsc{SteerSearch} prompt and the same E5-based dense retriever used during RL. A trajectory is retained only if its final answer passes exact match evaluation against the ground-truth answer. This ensures that the supervised initialization is based on successful reasoning and search behavior rather than noisy demonstrations.

\paragraph{RL data.}
The primary \textsc{SteerSearch} RL pool contains 40,000 samples. Of these, 24,000 are multi-hop HotpotQA samples with generated sub-question chains, split equally across decomposition, hint steering, and disambiguation. The remaining 16,000 are fixed-query samples split equally between HotpotQA and NQ. Mixing fixed-query and injected-trajectory examples prevents the model from becoming dependent on always receiving sub-question interventions, which is important because standard search evaluation contains no injections.

For Search-R1 training, we use 40,000 fixed-query samples drawn equally from HotpotQA and NQ, following~\citet{jin2025searchr1trainingllmsreason}. This gives Search-R1 and \textsc{SteerSearch} matched RL pool sizes while preserving the key difference in training format: fixed-goal search for Search-R1 versus mixed fixed-goal and mid-response steering trajectories for \textsc{SteerSearch}. Smaller RL pools of 8,000 and 16,000 samples are used for the data-scaling study in Section~\ref{subsec:appendix_data_config}.

\subsection{Rollout Configuration}
\label{sec:rollout-config}

The \textsc{SteerSearch} rollout loop constructs an interactive trajectory through a sequence of $K$ user-intervention phases, as described in Algorithm~\ref{alg:steersearch}. Each phase corresponds to one user goal $u_k$. The outer loop advances through the user goals $u_1,\ldots,u_K$, while the inner loop allows the model to perform a bounded number of reasoning and search turns before the next intervention is injected.

The number of phases $K$ is determined by the sub-question chain length of each example and therefore varies across samples. To avoid teaching the model a fixed injection schedule, each non-final phase receives a randomly sampled search budget. For the initial and intermediate phases, $T_k \sim \mathrm{Unif}\{1,\ldots,5\}$; for the final phase, $T_K=4$. This gives the model enough opportunity to retrieve evidence for the terminal goal while preventing unbounded exploration. The same retrieval setup is used during training and evaluation.

\subsection{Search Engine Setup}

All search calls retrieve from the 2018 Wikipedia snapshot processed following~\citet{karpukhin-etal-2020-dense}. The corpus is indexed with \texttt{intfloat/e5-base-v2}~\cite{wang2024textembeddingsweaklysupervisedcontrastive} using an exact flat index. We follow the corpus and top-$k$ configuration of~\citet{jin2025searchr1trainingllmsreason} unchanged across training and evaluation. For CCR similarity computation, the default ablation variants use the same \texttt{e5-base-v2} encoder as a frozen text encoder; the \textit{(qwenemb)} variants instead use \texttt{Qwen3-Embedding-0.6B}~\cite{zhang2025qwen3embeddingadvancingtext}.

\subsection{Per-Phase Search-Reasoning}

Within each phase, the model iterates between reasoning, retrieval, and answering. The explicit tag structure separates reasoning from tool calls, makes search actions observable to the environment, and allows retrieved passages to be inserted deterministically as part of the assistant turn rather than as model-generated text. At each inner-loop turn in phase $k$, the model first emits a reasoning block inside \texttt{\tthink{<think>}} or \texttt{\treason{<reason>}}, depending on the model format. It then either issues a search query in \texttt{\tsearch{<search>}} or produces an answer in \texttt{\tanswer{<answer>}}. A search action invokes the retriever, and the returned passages are appended as an \texttt{\tinfo{<information>}} block within the same assistant turn. The policy can therefore perform multiple search-read cycles within a phase. Phase $k$ terminates when the model emits an answer or exhausts its search budget $T_k$.

\subsection{Mid-Response User Intervention}

When a non-final phase $k<K$ terminates, the current assistant turn is closed, the next user turn $u_{k+1}$ is appended, and generation resumes from a new assistant turn. The full previous context is preserved, including reasoning, search queries, and retrieved passages. This design matches the target interaction setting: a user interrupts an ongoing search process, and the model must continue from the accumulated evidence rather than restarting from scratch.

\subsection{Intermediate Answer Removal and Final-Phase Generation}

Intermediate answer removal prevents a shortcut in the training environment. During a non-final phase, the model may emit an \texttt{\tanswer{<answer>}} before the full intervention chain is complete. If this premature answer were left in the trajectory, the policy could learn to terminate early instead of continuing to later user goals. Therefore, for any phase $k<K$, the premature \texttt{\tthink{<think>}\tanswer{<answer>}} or \texttt{\treason{<reason>}\tanswer{<answer>}} turn is removed from the policy-gradient loss by masking its token positions. Earlier reasoning and retrieval turns in the same phase are retained, preserving useful search behavior.

After deleting the premature answer, the next intervention $u_{k+1}$ is injected as if the phase had ended by budget exhaustion. This forces the rollout to continue through the complete sub-question chain. If the model exhausts the final-phase budget $T_K$ without producing an \texttt{\tanswer{<answer>}} tag, we run one additional generation pass with search disabled. This ensures that every trajectory yields a final prediction for reward computation.

\subsection{Evaluation Benchmarks}
\label{sec:benchmarks}

\begin{table}[t]
\centering\small
\resizebox{\columnwidth}{!}{%
\begin{tabular}{lccccc}
\toprule
\multirow{2}{*}{\textbf{Dataset}} & \multicolumn{2}{c}{\textbf{Samples}} & \multicolumn{3}{c}{\textbf{Sub-questions}} \\
\cmidrule(lr){2-3}\cmidrule(lr){4-6}
 & \multicolumn{1}{c}{\shortstack{\textbf{Standard}\\\textbf{Bench}}} & \multicolumn{1}{c}{\shortstack{\textbf{\textsc{SteerSearch}}\\\textbf{\textsc{-Bench}}}} & \textbf{Min} & \textbf{Mean} & \textbf{Max} \\
\midrule
HotpotQA        & 7{,}405  & 7{,}405  & 2 & 2.8 & 5 \\
2WikiMultiHopQA & 12{,}576 & 12{,}576 & 2 & 3.0 & 5 \\
MuSiQue         & 2{,}417  & 2{,}412  & 2 & 3.1 & 5 \\
Bamboogle        & 125      & 125      & 2 & 2.6 & 4 \\
\midrule
NQ              & 3{,}610  & {---}    & \multicolumn{3}{c}{---} \\
TriviaQA        & 11{,}313 & {---}    & \multicolumn{3}{c}{---} \\
PopQA           & 14{,}267 & {---}    & \multicolumn{3}{c}{---} \\
\midrule
\textbf{Total}  & \textbf{51{,}713} & \textbf{22{,}518} & & & \\
\bottomrule
\end{tabular}%
}
\caption{Dataset composition of the Standard Benchmark and \textsc{SteerSearch-Bench}.}
\label{tab:bench-stats}
\end{table}

We evaluate models on two benchmark suites. The first measures standard retrieval-augmented question answering without user intervention. The second measures mid-response steering, where sub-questions are injected during generation and the model must answer the final user goal.

\paragraph{Standard Benchmark.}
We use \textit{Standard Benchmark} as a shorthand for the held-out evaluation suite of~\citet{jin2025searchr1trainingllmsreason}, which tests non-interactive, single-query search-reasoning in the conventional setting where the user goal is fixed and no mid-response intervention occurs. It contains seven datasets spanning two types: single-hop general QA, comprising NQ~\cite{kwiatkowski-etal-2019-natural}, TriviaQA~\cite{joshi-etal-2017-triviaqa}, and PopQA~\cite{mallen2023trustlanguagemodelsinvestigating}; and multi-hop reasoning, comprising HotpotQA~\cite{yang-etal-2018-hotpotqa}, 2WikiMultiHopQA~\cite{ho-etal-2020-constructing}, MuSiQue~\cite{trivedi-etal-2022-musique}, and Bamboogle~\cite{press2023measuringnarrowingcompositionalitygap}. Models answer each item without injected sub-questions, allowing us to test whether \textsc{SteerSearch} preserves ordinary search-reasoning ability alongside its new steering capability.

\paragraph{\textsc{SteerSearch-Bench}.}
\textsc{SteerSearch-Bench} evaluates the core setting of this work: whether a model can follow user interventions injected during an ongoing search trajectory. It is constructed from the multi-hop portions of HotpotQA, 2WikiMultiHopQA, MuSiQue, and Bamboogle. NQ, TriviaQA, and PopQA are excluded because they are primarily single-hop general QA datasets and do not naturally support multi-step intervention chains.

For each example, GPT-5.4 generates a chain of two to five sub-questions. Benchmark generation uses ten stylistically distinct injection styles, including direct instructions, scope refinements, query pivots, source hunts, and distractor suppressions. This diversity is intended to prevent the benchmark from measuring only one narrow form of follow-up phrasing. The generated chains are filtered by the same GPT-5.4 judge committee used for training data. The benchmark generation prompts are provided in Appendix~\ref{sec:appendix_genprompts}.

At evaluation time, sub-questions are injected into the model context at the prescribed trajectory boundaries. The model may continue searching and reasoning after each injection, but only the final answer is scored. All results are reported using exact match against the ground-truth answer string, following the normalization convention of~\citet{jin2025searchr1trainingllmsreason}. Table~\ref{tab:bench-stats} reports the benchmark composition and sub-question chain length statistics. Table~\ref{tab:bench-comparison} in Appendix~\ref{subsec:standard_vs_steer_bench} shows representative examples from both settings side by side.

\subsection{Baselines}
\label{sec:baselines}

The primary baseline is Search-R1~\cite{jin2025searchr1trainingllmsreason}. Search-R1 trains the model to answer a fixed user question through iterative reasoning and retrieval. It uses the same base model, retriever, PPO framework, and reward format as \textsc{SteerSearch}, but does not include mid-response sub-question injection. This makes it the most direct comparison for testing whether interactive trajectory training improves steering beyond ordinary search-augmented reasoning.

We also report the Vanilla model, which denotes the original instruction-tuned checkpoint without SFT or RL training in our environment. On the Standard Benchmark, we include representative retrieval-augmented baselines reported by~\citet{jin2025searchr1trainingllmsreason}: direct inference, chain-of-thought prompting (CoT), interleaved retrieval and reasoning (IRCoT)~\cite{trivedi-etal-2023-interleaving}, Search-o1~\cite{li-etal-2025-search}, retrieval-augmented generation (RAG), supervised fine-tuning (SFT), DeepSeekR1-instruct~\cite{Guo_2025}, and rejection sampling~\cite{ahn-etal-2024-large}. These methods contextualize performance on standard QA, while Search-R1 remains the controlled baseline for the mid-response steering setting.
\begin{algorithm*}[t]
\small
\caption{\textsc{SteerSearch} Rollout with Mid-Response Sub-Question Steering}
\label{alg:steersearch}
\begin{algorithmic}[1]
\Require Question $q$, sub-question chain $\mathbf{u} = \langle u_1, \ldots, u_K \rangle$, policy $\pi_\theta$, retriever $\mathcal{R}$, phase budget ranges $[T_k^{\min}, T_k^{\max}]$ for $k=1,\ldots,K$
\Ensure Full trajectory $\tau$

\State $\tau \leftarrow \langle u_1 \rangle$ \Comment{Initialize with first sub-question}

\For{$k = 1$ \textbf{to} $K$} \Comment{Outer loop: one phase per sub-question}
    \State Sample budget $T_k \sim \mathrm{Uniform}(T_k^{\min}, T_k^{\max})$
    \State Sample injection point $t^* \sim \mathrm{Uniform}(0,\, T_k - 1)$ \Comment{Random mid-response steering point}
    \For{$t = 1$ \textbf{to} $T_k$}
        \State $y \sim \pi_\theta(\cdot \mid \tau)$ \hfill \Comment{Generate until \tsearch{</search>} or \tanswer{</answer>}}
        \State $\tau \leftarrow \tau + y$
        \If{\tsearch{<search>} detected in $y$}
            \State Extract query $q' \leftarrow \textsc{Parse}(y)$; retrieve $d \leftarrow \mathcal{R}(q')$
            \State $\tau \leftarrow \tau + \langle \tinfo{<information>}\; d\; \tinfo{</information>} \rangle$
        \ElsIf{\tanswer{<answer>} detected in $y$ \textbf{and} $k = K$}
            \State \Return $\tau$ \Comment{Final answer produced}
        \ElsIf{\tanswer{<answer>} detected in $y$ \textbf{and} $k < K$}
            \State \textbf{Delete} last \tthink{<think>}\tanswer{<answer>} turn from $\tau$ \Comment{Suppress premature answer}
            \State \textbf{break} \Comment{Proceed to injection}
        \Else
            \State $\tau \leftarrow \tau +$ ``Your action is invalid. Use \tsearch{<search>} or \tanswer{<answer>}.'' \Comment{Rethink}
        \EndIf
        \If{$t \geq t^*$ \textbf{and} $k < K$} \Comment{Steering point reached}
            \State \textbf{break}
        \EndIf
    \EndFor

    \If{$k < K$}
        \State $\tau \leftarrow \tau + \langle u_{k+1} \rangle$ \Comment{Inject next sub-question mid-response}
    \EndIf
\EndFor
\State \Comment{Budget exhausted on final phase: generate with retrieval disabled}
\State $y \sim \pi_\theta(\cdot \mid \tau,\; \texttt{no\_search})$
\State \Return $\tau + y$
\end{algorithmic}
\end{algorithm*}

\section{Training Prompt Templates}
\label{sec:appendix_training_prompts}
\label{app:prompt}

All experiments use the \textsc{SteerSearch} prompt as mentioned in Table \ref{tab:prompts}, which has two variants depending on model architecture. The \treason{<reason>} variant is used for models where \tthink{<think>} is a reserved token.

\begin{table*}[t]
\centering\footnotesize
\begin{minipage}[t]{0.48\textwidth}
\centering
\textbf{Standard (\tthink{<think>})}
\vspace{3pt}

\hrule
\vspace{4pt}
{\ttfamily\raggedright
Answer the given question. You must conduct reasoning inside
\tthink{<think>} and \tthink{</think>} first every time you get a new
question, follow-up, or new information. After reasoning, if you find
you lack some knowledge, you can call a search engine by
\tsearch{<search>} query \tsearch{</search>}, and it will return the
top searched results between \tinfo{<information>} and
\tinfo{</information>}. You can search as many times as you want. If
you find no further external knowledge needed, you can directly provide
the answer inside \tanswer{<answer>} and \tanswer{</answer>} without
detailed illustrations. For example, \tanswer{<answer>} Paris
\tanswer{</answer>}. \textbf{If another question or follow-up is given
later, treat it as the next step and continue in the same format.}
Question: \{question\}
}
\end{minipage}
\hfill
\begin{minipage}[t]{0.48\textwidth}
\centering
\textbf{Reason variant (\treason{<reason>})}
\vspace{3pt}

\hrule
\vspace{4pt}
{\ttfamily\raggedright
Answer the given question. You must conduct reasoning inside
\treason{<reason>} and \treason{</reason>} first every time you get a
new question, follow-up, or new information. After reasoning, if you
find you lack some knowledge, you can call a search engine by
\tsearch{<search>} query \tsearch{</search>}, and it will return the
top searched results between \tinfo{<information>} and
\tinfo{</information>}. You can search as many times as you want. If
you find no further external knowledge needed, you can directly provide
the answer inside \tanswer{<answer>} and \tanswer{</answer>} without
detailed illustrations. For example, \tanswer{<answer>} Paris
\tanswer{</answer>}. \textbf{If another question or follow-up is given
later, treat it as the next step and continue in the same format.}
Question: \{question\}
}
\end{minipage}
\caption{\textsc{SteerSearch} training prompt templates. The bold sentence enables mid-response user intervention. The \treason{<reason>} variant is used for models where \tthink{<think>} is a reserved token.}
\label{tab:prompts}
\end{table*}

\section{Additional Results}
\label{app:results}

\subsection{Semantic Evaluation}
\label{subsec:semantic_eval}

Table~\ref{tab:semantic_eval} reports semantic evaluation on \textsc{SteerSearch-Bench}. Exact match can underestimate valid answers when predicted and reference answers differ in surface form. We therefore additionally evaluate the Search-R1 and \textsc{SteerSearch} predictions without CCR using GPT-5.5. Each output is judged independently three times using the full trajectory, extracted final answer, and gold target, with the final label determined by majority vote. Judge accuracy is the fraction of outputs assigned correct.

\begin{table}[H]
\centering
\small
\resizebox{\columnwidth}{!}{
\begin{tabular}{llcc}
\toprule
\textbf{Model} & \textbf{Method} & \textbf{EM} & \textbf{GPT-5.5 Judge} \\
\midrule
Qwen2.5-3B & Search-R1 & 0.111 & 0.156 \\
           & \textsc{SteerSearch}
           & \textbf{0.313} {\color{green!50!black}($\uparrow$20.2 pp)}
           & \textbf{0.393} {\color{green!50!black}($\uparrow$23.7 pp)} \\
\midrule
Qwen3-4B & Search-R1 & 0.366 & 0.463 \\
         & \textsc{SteerSearch}
         & \textbf{0.446} {\color{green!50!black}($\uparrow$8.0 pp)}
         & \textbf{0.564} {\color{green!50!black}($\uparrow$10.1 pp)} \\
\bottomrule
\end{tabular}
}
\caption{Semantic evaluation on \textsc{SteerSearch-Bench}. GPT-5.5 judges each output independently three times using the full trajectory, extracted final answer, and gold target; the final label is determined by majority vote. Green values show absolute improvement over Search-R1.}
\label{tab:semantic_eval}
\end{table}

The semantic evaluation preserves the separation observed under EM. \textsc{SteerSearch} improves judge accuracy over Search-R1 for both primary models, indicating that its advantage is not explained by exact-match sensitivity to answer surface form.

\subsection{Natural Multi-Turn Evaluation on TopiOCQA}
\label{subsec:topiocqa}

\begin{table*}[!t]
\centering
\small
% \resizebox{\columnwidth}{!}{
\begin{tabular}{llccc}
\toprule
\textbf{Model} & \textbf{Method} & \textbf{EM} & \textbf{Token F1} & \textbf{GPT-5.5 Judge} \\
\midrule
Qwen2.5-3B & Search-R1 & 0.069 & 0.132 & 0.119 \\
           & \textsc{SteerSearch}
           & \textbf{0.114} {\color{green!50!black}($\uparrow$4.5 pp)}
           & \textbf{0.218} {\color{green!50!black}($\uparrow$8.6 pp)}
           & \textbf{0.259} {\color{green!50!black}($\uparrow$14.0 pp)} \\
\midrule
Qwen3-4B & Search-R1 & 0.035 & 0.067 & 0.064 \\
         & \textsc{SteerSearch}
         & \textbf{0.139} {\color{green!50!black}($\uparrow$10.4 pp)}
         & \textbf{0.233} {\color{green!50!black}($\uparrow$16.6 pp)}
         & \textbf{0.302} {\color{green!50!black}($\uparrow$23.8 pp)} \\
\bottomrule
\end{tabular}
% }
\caption{Evaluation on human-authored TopiOCQA conversations. Green values show absolute improvement over Search-R1. Models use the main checkpoints without CCR.}
\label{tab:topiocqa}
\end{table*}

Table~\ref{tab:topiocqa} reports evaluation on TopiOCQA~\cite{adlakha-etal-2022-topiocqa} validation set, an external conversational QA dataset whose conversations are written by pairs of human annotators rather than generated by our sub-question construction pipeline. We evaluate approximately 200 full conversations containing natural follow-up questions, coreference, and topic shifts over 10--16 turns, with a mean conversation length of 12.26 turns. Table~\ref{tab:topiocqa-examples} in Appendix~\ref{subsec:topiocqa_samples} shows randomly selected examples from the TopiOCQA benchmark.

We use the same Search-R1 and \textsc{SteerSearch} checkpoints reported in the main results, without CCR. To adapt TopiOCQA to the mid-response setting, the first user turn initializes the search trajectory and each subsequent human-written turn is injected while generation is ongoing. This preserves the original conversational sequence while testing whether the model can repeatedly update its active answer target under naturally occurring follow-ups.

We report exact match, token-level F1, and GPT-5.5 semantic accuracy. For semantic evaluation, each prediction is judged independently three times and the final label is determined by majority vote.

For Qwen2.5, \textsc{SteerSearch} improves over Search-R1 by 4.5 percentage points in EM, 8.6 points in token F1, and 14.0 points under semantic judging. For Qwen3, the corresponding gains are 10.4, 16.6, and 23.8 points. The larger improvements under F1 and semantic judging indicate that the result is not dependent on exact string matching. Since these conversations are human-authored and unseen during training, the result also reduces the possibility that the main steering gains arise only from the generated intervention structure of \textsc{SteerSearch-Bench}.

TopiOCQA nevertheless remains a controlled retrieval setting using clean Wikipedia evidence. This experiment therefore tests generalization to natural multi-turn intervention structure, but does not establish robustness to noisy open-web search or arbitrary browser interactions.

\subsection{Model Scaling Results}
\label{subsec:appendix_model_scaling}

Tables~\ref{tab:scaling_ss} and~\ref{tab:scaling_sr1} report model-scaling results for the Qwen2.5 Instruct family on \textsc{SteerSearch-Bench} and the Standard Benchmark, respectively. All experiments use \textsc{SteerSearch} without CCR to isolate the effect of training on interactive trajectories.

\begin{table*}[h]
\centering\small
\resizebox{\textwidth}{!}{
\begin{tabular}{llccccc}
\toprule
\textbf{Model} & \textbf{Method} & \textbf{HotpotQA$^\dagger$} & \textbf{2WikiMQA$^*$} & \textbf{MuSiQue$^*$} & \textbf{Bamboogle$^*$} & \textbf{Avg.} \\
\midrule
Qwen2.5-3B-Instruct & Vanilla & 0.101 & 0.054 & 0.039 & 0.080 & 0.069 \\
 & Search-R1 & 0.192 & 0.063 & 0.078 & 0.112 & 0.111 \\
\rowcolor{blue!8}
 & \textsc{SteerSearch} & \textbf{0.376} & \textbf{0.260} & \textbf{0.241} & \textbf{0.376} & \textbf{0.313} \\
\midrule
Qwen2.5-7B-Instruct & Vanilla & 0.240 & 0.234 & 0.133 & 0.264 & 0.218 \\
 & Search-R1 & 0.230 & 0.127 & 0.088 & 0.160 & 0.151 \\
\rowcolor{blue!8}
 & \textsc{SteerSearch} & \textbf{0.463} & \textbf{0.425} & \textbf{0.286} & \textbf{0.544} & \textbf{0.430} \\
\midrule
Qwen2.5-14B-Instruct & Vanilla & 0.367 & 0.353 & 0.199 & 0.352 & 0.318 \\
 & Search-R1 & 0.411 & 0.365 & 0.226 & 0.400 & 0.350 \\
\rowcolor{blue!8}
 & \textsc{SteerSearch} & \textbf{0.480} & \textbf{0.449} & \textbf{0.303} & \textbf{0.512} & \textbf{0.436} \\
\bottomrule
\end{tabular}
}
\caption{Exact match on \textsc{SteerSearch-Bench} across model sizes for the Qwen2.5 Instruct family.}
\label{tab:scaling_ss}
\end{table*}
\begin{table*}[!t]
\centering\small
\resizebox{\textwidth}{!}{
\begin{tabular}{llcccccccc}
\toprule
\textbf{Model} & \textbf{Method} & \multicolumn{3}{c}{\textbf{General QA}} & \multicolumn{4}{c}{\textbf{Multi-Hop QA}} & \\
\cmidrule(lr){3-5}\cmidrule(lr){6-9}
& & \textbf{NQ$^\dagger$} & \textbf{TriviaQA$^*$} & \textbf{PopQA$^*$} & \textbf{HotpotQA$^\dagger$} & \textbf{2WikiMQA$^*$} & \textbf{MuSiQue$^*$} & \textbf{Bamboogle$^*$} & \textbf{Avg.} \\
\midrule
Qwen2.5-3B-Instruct & Vanilla & 0.216 & 0.415 & 0.259 & 0.188 & 0.175 & 0.044 & 0.168 & 0.209 \\
 & Search-R1 & 0.447 & 0.602 & 0.433 & 0.429 & \textbf{0.424} & 0.175 & 0.328 & 0.405 \\
\rowcolor{blue!8}
 & \textsc{SteerSearch} & \textbf{0.450} & \textbf{0.615} & \textbf{0.442} & \textbf{0.442} & 0.409 & \textbf{0.209} & \textbf{0.384} & \textbf{0.422} \\
\midrule
Qwen2.5-7B-Instruct & Vanilla & 0.257 & 0.491 & 0.295 & 0.240 & 0.220 & 0.091 & 0.320 & 0.273 \\
 & Search-R1 & 0.465 & 0.643 & 0.453 & 0.449 & \textbf{0.432} & 0.184 & \textbf{0.456} & 0.440 \\
\rowcolor{blue!8}
 & \textsc{SteerSearch} & \textbf{0.475} & \textbf{0.645} & \textbf{0.465} & \textbf{0.458} & 0.427 & \textbf{0.200} & \textbf{0.456} & \textbf{0.446} \\
\midrule
Qwen2.5-14B-Instruct & Vanilla & 0.357 & 0.636 & 0.382 & 0.389 & 0.359 & 0.156 & 0.456 & 0.391 \\
 & Search-R1 & 0.485 & \textbf{0.684} & \textbf{0.482} & 0.482 & 0.469 & 0.228 & \textbf{0.512} & 0.477 \\
\rowcolor{blue!8}
 & \textsc{SteerSearch} & \textbf{0.488} & 0.677 & 0.480 & \textbf{0.488} & \textbf{0.486} & \textbf{0.246} & 0.488 & \textbf{0.479} \\
\bottomrule
\end{tabular}
}
\caption{Exact match on the Standard Benchmark across model sizes for the Qwen2.5 Instruct family.}
\label{tab:scaling_sr1}
\end{table*}

On \textsc{SteerSearch-Bench}, \textsc{SteerSearch} improves over Search-R1 across all three Qwen2.5 model sizes. The gain is 20.2 percentage points at 3B, 27.9 points at 7B, and 8.6 points at 14B. The smaller gap at 14B results primarily from the stronger Search-R1 baseline, which increases from 15.1\% at 7B to 35.0\% at 14B, while \textsc{SteerSearch} remains above 43\% at both scales. This suggests that larger models partially acquire target-updating behavior from scale and general instruction following, while explicit intervention training continues to provide an advantage.

The Standard Benchmark shows a different pattern. At 7B and 14B, \textsc{SteerSearch} changes average performance relative to Search-R1 by only +0.6 and +0.2 percentage points, respectively. The steering gains therefore do not arise from a general improvement in retrieval-augmented QA, but are concentrated in the interrupted interaction setting.

\subsection{Cross-Family Generalization}
\label{subsec:cross-family-generalization}

\begin{table*}[h]
\centering\small
\begin{tabular}{lccccc}
\toprule
\textbf{Method} & \textbf{HotpotQA$^\dagger$} & \textbf{2WikiMQA$^*$} & \textbf{MuSiQue$^*$} & \textbf{Bamboogle$^*$} & \textbf{Avg.} \\
\midrule
\multicolumn{6}{l}{\textit{Gemma3-4B-Instruct}} \\
Search-R1 & 0.290 & 0.210 & 0.158 & 0.264 & 0.230 \\
\rowcolor{orange!8}
\textsc{SteerSearch} & 0.356 & 0.313 & 0.216 & 0.384 & 0.317 \\
\rowcolor{blue!8}
\textsc{SteerSearch}+CCR \textit{(QwenEmb)} & \textbf{0.460} & \textbf{0.430} & \textbf{0.279} & \textbf{0.480} & \textbf{0.412} \\
\bottomrule
\end{tabular}
\caption{\textbf{Exact match on \textsc{SteerSearch-Bench} for Gemma3-4B-Instruct} with mid-response sub-question injection. $\dagger$~in-domain, $*$~out-of-domain, and \textbf{Bold} = best among fine-tuned methods. \textsc{SteerSearch} uses only the terminal outcome reward. \textsc{SteerSearch}+CCR \textit{(QwenEmb)} uses cumulative-context CCR with terminal assignment and \texttt{Qwen3-Embedding-0.6B} as the frozen CCR similarity encoder; document retrieval remains unchanged.}
\label{tab:gemma_ss_bench}
\end{table*}
\begin{table*}[!t]
\centering
\small
\resizebox{\textwidth}{!}{
\begin{tabular}{lccccccccc}
\toprule
\multirow{3}{*}{\textbf{Method}} & \multicolumn{3}{c}{\textbf{General QA}} & \multicolumn{4}{c}{\textbf{Multi-Hop QA}} & \multirow{3}{*}{\textbf{Avg.}} \\
\cmidrule(lr){2-4}\cmidrule(lr){5-8}
& \textbf{NQ$^\dagger$} & \textbf{TriviaQA$^*$} & \textbf{PopQA$^*$} & \textbf{HotpotQA$^\dagger$} & \textbf{2WikiMQA$^*$} & \textbf{MuSiQue$^*$} & \textbf{Bamboogle$^*$} & \\
\midrule
\multicolumn{9}{l}{\textit{Gemma3-4B-Instruct}} \\
Search-R1 & 0.364 & 0.576 & 0.390 & 0.341 & 0.228 & 0.113 & 0.240 & 0.322 \\
\rowcolor{orange!8}
\textsc{SteerSearch} & 0.375 & 0.587 & 0.382 & 0.371 & 0.298 & 0.129 & 0.328 & 0.353 \\
\rowcolor{blue!8}
\textsc{SteerSearch}+CCR \textit{(QwenEmb)} & \textbf{0.465} & \textbf{0.635} & \textbf{0.419} & \textbf{0.442} & \textbf{0.392} & \textbf{0.187} & \textbf{0.376} & \textbf{0.416} \\
\bottomrule
\end{tabular}
}
\caption{\textbf{Exact match on the Standard Benchmark for Gemma3-4B-Instruct.} $\dagger$~in-domain, $*$~out-of-domain, and \textbf{Bold} = best among fine-tuned methods.}
% \caption{\textbf{Exact match on the Standard Benchmark for Gemma3-4B-Instruct.} $\dagger$~in-domain, $*$~out-of-domain, and \textbf{Bold} = best among fine-tuned methods. \textsc{SteerSearch} uses only the terminal outcome reward. \textsc{SteerSearch}+CCR \textit{(QwenEmb)} uses cumulative-context CCR with terminal assignment and \texttt{Qwen3-Embedding-0.6B} as the frozen CCR similarity encoder; document retrieval remains unchanged.}
\label{tab:gemma_sr1_bench}
\end{table*}

We additionally evaluate Gemma3-4B-Instruct in the text-only setting to test whether the steering effect extends beyond the Qwen family. Tables~\ref{tab:gemma_ss_bench} and~\ref{tab:gemma_sr1_bench} compare Search-R1, \textsc{SteerSearch}, and \textsc{SteerSearch}+CCR on \textsc{SteerSearch-Bench} and the Standard Benchmark, respectively.

The steering curriculum transfers to Gemma3-4B-Instruct, with a substantially larger improvement in the interrupted setting than in standard QA. CCR further strengthens \textsc{SteerSearch} in both settings, with a larger effect than observed for either Qwen model and a clear improvement even on the Standard Benchmark. One possible explanation is that Gemma3's post-training combines distillation with multiple RL objectives and explicitly targets instruction-following and in-context attribution, which may make its policy more receptive to an auxiliary context-alignment reward. This remains a hypothesis, as the experiments do not isolate the effect of the underlying post-training recipe.

\subsection{Data Configuration Ablation}
\label{subsec:appendix_data_config}

Tables~\ref{tab:searchr1_bench_full} and~\ref{tab:steer_bench_full} report results for all training data configurations evaluated. The main results (Tables~\ref{tab:searchr1_bench} and~\ref{tab:steer_bench}) show only the SFT-4k/PPO-40k configuration for \textsc{SteerSearch}. The SFT-2k/PPO-8k configuration uses the same data budget as Search-R1, providing a matched comparison between fixed-query and mid-response steering training.

\begin{table*}[t]
\centering
\small
\resizebox{\textwidth}{!}{
\begin{tabular}{llccccccccc}
\toprule
\textbf{Method} & \textbf{Config} & \multicolumn{3}{c}{\textbf{General QA}} & \multicolumn{4}{c}{\textbf{Multi-Hop QA}} & \\
\cmidrule(lr){3-5}\cmidrule(lr){6-9}
& & \textbf{NQ$^\dagger$} & \textbf{TriviaQA$^*$} & \textbf{PopQA$^*$} & \textbf{HotpotQA$^\dagger$} & \textbf{2WikiMQA$^*$} & \textbf{MuSiQue$^*$} & \textbf{Bamboogle$^*$} & \textbf{\textbf{Avg.}} \\
\midrule
\multicolumn{10}{l}{\textit{Qwen2.5-3B-Instruct}} \\
\rowcolor{gray!8}
Vanilla & -- & 0.216 & 0.415 & 0.259 & 0.188 & 0.175 & 0.044 & 0.168 & 0.209 \\
\rowcolor{orange!8}
Search-R1 & SFT-2k / PPO-8k & 0.410 & 0.611 & 0.420 & \textbf{0.434} & 0.374 & \textbf{0.183} & \textbf{0.376} & 0.401 \\
\rowcolor{orange!8}
Search-R1 & SFT-4k / PPO-16k & 0.438 & \textbf{0.618} & \textbf{0.443} & 0.417 & 0.396 & 0.157 & 0.328 & 0.400 \\
\rowcolor{orange!8}
Search-R1 & SFT-4k / PPO-40k & \textbf{0.447} & 0.602 & 0.433 & 0.429 & \textbf{0.424} & 0.175 & 0.328 & \textbf{0.405} \\
\cdashline{1-10}
\noalign{\vskip 2pt}
\rowcolor{blue!8}
\textsc{SteerSearch} & SFT-2k / PPO-8k & 0.397 & 0.611 & 0.405 & 0.408 & 0.337 & 0.153 & 0.352 & 0.380 \\
\rowcolor{blue!8}
\textsc{SteerSearch} & SFT-4k / PPO-16k & 0.424 & 0.614 & 0.427 & \textbf{0.443} & \textbf{0.429} & 0.197 & 0.352 & 0.412 \\
\rowcolor{blue!8}
\textsc{SteerSearch} & SFT-4k / PPO-40k & \textbf{0.450} & \textbf{0.615} & \textbf{0.442} & 0.442 & 0.409 & \textbf{0.209} & \textbf{0.384} & \textbf{0.422} \\
\midrule
\multicolumn{10}{l}{\textit{Qwen3-4B-Instruct-2507}} \\
\rowcolor{gray!8}
Vanilla & -- & 0.283 & 0.497 & 0.317 & 0.244 & 0.270 & 0.060 & 0.376 & 0.292 \\
\rowcolor{orange!8}
Search-R1 & SFT-2k / PPO-8k & 0.440 & \textbf{0.668} & 0.458 & \textbf{0.500} & \textbf{0.479} & \textbf{0.236} & \textbf{0.512} & 0.470 \\
\rowcolor{orange!8}
Search-R1 & SFT-4k / PPO-16k & 0.487 & 0.664 & 0.473 & 0.484 & 0.470 & 0.211 & \textbf{0.512} & \textbf{0.472} \\
\rowcolor{orange!8}
Search-R1 & SFT-4k / PPO-40k & \textbf{0.496} & 0.660 & \textbf{0.487} & 0.481 & 0.457 & 0.233 & 0.488 & \textbf{0.472} \\
\cdashline{1-10}
\noalign{\vskip 2pt}
\rowcolor{blue!8}
\textsc{SteerSearch} & SFT-2k / PPO-8k & 0.398 & 0.644 & 0.408 & 0.457 & 0.349 & 0.186 & 0.424 & 0.409 \\
\rowcolor{blue!8}
\textsc{SteerSearch} & SFT-4k / PPO-16k & 0.452 & \textbf{0.656} & 0.445 & \textbf{0.488} & 0.476 & 0.230 & 0.456 & 0.458 \\
\rowcolor{blue!8}
\textsc{SteerSearch} & SFT-4k / PPO-40k & \textbf{0.491} & 0.652 & \textbf{0.479} & 0.475 & \textbf{0.498} & \textbf{0.239} & \textbf{0.480} & \textbf{0.473} \\
\bottomrule
\end{tabular}
}
\vspace{-2mm}
\caption{\textbf{Exact match on the Standard Benchmark.} With $\dagger$~in-domain, $*$~out-of-domain, \textbf{Bold} = best config per method per model, and Vanilla = not any training.}
\label{tab:searchr1_bench_full}
\end{table*}
\begin{table*}[t]
\centering\small
\resizebox{\textwidth}{!}{
\begin{tabular}{llcccccc}
\toprule
\textbf{Method} & \textbf{Config} & \textbf{HotpotQA$^\dagger$} & \textbf{2WikiMQA$^*$} & \textbf{MuSiQue$^*$} & \textbf{Bamboogle$^*$} & \textbf{\textbf{Avg.}} \\
\midrule
\multicolumn{7}{l}{\textit{Qwen2.5-3B-Instruct}} \\
\rowcolor{gray!8}
Vanilla & -- & 0.101 & 0.054 & 0.039 & 0.080 & 0.068 \\
\rowcolor{orange!8}
Search-R1 & SFT-2k / PPO-8k & 0.152 & \textbf{0.063} & 0.061 & 0.104 & 0.095 \\
\rowcolor{orange!8}
Search-R1 & SFT-4k / PPO-16k & 0.179 & 0.060 & 0.072 & 0.104 & 0.104 \\
\rowcolor{orange!8}
Search-R1 & SFT-4k / PPO-40k & \textbf{0.192} & \textbf{0.063} & \textbf{0.078} & \textbf{0.112} & \textbf{0.111} \\
\cdashline{1-7}
\noalign{\vskip 2pt}
\rowcolor{blue!8}
\textsc{SteerSearch} & SFT-2k / PPO-8k & 0.349 & 0.212 & 0.208 & \textbf{0.392} & 0.290 \\
\rowcolor{blue!8}
\textsc{SteerSearch} & SFT-4k / PPO-16k & 0.261 & 0.148 & 0.123 & 0.216 & 0.187 \\
\rowcolor{blue!8}
\textsc{SteerSearch} & SFT-4k / PPO-40k & \textbf{0.376} & \textbf{0.260} & \textbf{0.241} & 0.376 & \textbf{0.313} \\
\midrule
\multicolumn{7}{l}{\textit{Qwen3-4B-Instruct-2507}} \\
\rowcolor{gray!8}
Vanilla & -- & 0.171 & 0.173 & 0.068 & 0.096 & 0.127 \\
\rowcolor{orange!8}
Search-R1 & SFT-2k / PPO-8k & 0.279 & 0.160 & 0.153 & 0.312 & 0.226 \\
\rowcolor{orange!8}
Search-R1 & SFT-4k / PPO-16k & 0.301 & 0.248 & 0.127 & 0.232 & 0.227 \\
\rowcolor{orange!8}
Search-R1 & SFT-4k / PPO-40k & \textbf{0.423} & \textbf{0.349} & \textbf{0.250} & \textbf{0.440} & \textbf{0.366} \\
\cdashline{1-7}
\noalign{\vskip 2pt}
\rowcolor{blue!8}
\textsc{SteerSearch} & SFT-2k / PPO-8k & 0.433 & 0.275 & 0.284 & 0.496 & 0.372 \\
\rowcolor{blue!8}
\textsc{SteerSearch} & SFT-4k / PPO-16k & 0.464 & 0.397 & 0.278 & 0.520 & 0.415 \\
\rowcolor{blue!8}
\textsc{SteerSearch} & SFT-4k / PPO-40k & \textbf{0.472} & \textbf{0.457} & \textbf{0.305} & \textbf{0.552} & \textbf{0.446} \\
\bottomrule
\end{tabular}
}
\caption{\textbf{Exact match on \textsc{SteerSearch-Bench} with mid-response sub-question injection.} $\dagger$~in-domain, $*$~out-of-domain, \textbf{Bold} = best config per method per model, and Vanilla = not any training.}
\label{tab:steer_bench_full}
\end{table*}

\subsection{Seed Sensitivity of the Main Comparison}
\label{subsec:main-seed-sensitivity}

We evaluate seed sensitivity for the main Search-R1 versus \textsc{SteerSearch} comparison using three independent runs of Qwen2.5-3B-Instruct with the same 4K-SFT initialization and 40K-PPO training budget. Table~\ref{tab:main-seed-sensitivity} reports exact match (EM) as the mean and sample standard deviation across runs. On \textsc{SteerSearch-Bench}, \textsc{SteerSearch} outperforms Search-R1 by 21.2 percentage points on average, while performance on the Standard Benchmark remains comparable, with a difference of 0.1 percentage points. These results support the robustness of the main comparison to training-seed variation for Qwen2.5-3B-Instruct.

\begin{table}[t]
    \centering
    \small
    \setlength{\tabcolsep}{5pt}
    \begin{tabular}{lcc}
    \toprule
    Benchmark & Search-R1 & \textsc{SteerSearch} \\
    \midrule
    \textsc{SteerSearch-Bench}
    & $0.133 \pm 0.031$
    & $0.345 \pm 0.028$ \\
    Standard Benchmark
    & $0.410 \pm 0.004$
    & $0.411 \pm 0.010$ \\
    \bottomrule
    \end{tabular}
    \caption{Seed sensitivity of the main Search-R1 versus
    \textsc{SteerSearch} comparison on Qwen2.5-3B-Instruct.
    Entries report mean EM $\pm$ sample standard deviation across
    three runs.}
    \label{tab:main-seed-sensitivity}
\end{table}

\section{Additional Analysis}
\label{app:analysis}

This appendix provides additional analyses of attribution controls, training dynamics, retrieval allocation, intermediate context, and performance across intervention depths.

\subsection{Matched No-Injection Training Control}
\label{subsec:matched-no-injection}

The inference-time diagnostics in Table~\ref{tab:component_analysis} keep the original Search-R1 and \textsc{SteerSearch} training procedures fixed. To directly test whether the main gains arise from exposure to decomposed multi-hop content rather than mid-response target reassignment, we evaluate additional controls on \textsc{SteerSearch-Bench}. For the concat-chain training control, we retrain Search-R1 using the same SFT initialization and 40K PPO pool as \textsc{SteerSearch}, but concatenate the decomposed sub-question chains upfront without mid-response injection. The base model, retriever, training budget, and final-answer reward are otherwise matched, while \textsc{SteerSearch} retains its standard injected-trajectory training. The remaining controls evaluate upfront concatenation and the first sub-question without additional training.

% \begin{table*}[!h]
% \centering
% \small
% \begin{tabular}{lcc}
% \toprule
% \textbf{Control} & \textbf{Qwen2.5-3B-Instruct} & \textbf{Qwen3-4B-Instruct-2507} \\
% \midrule
% Concat-chain training &
% 0.173 / \textbf{0.313} {\color{green!50!black}($\uparrow$14.0 pp)} &
% 0.343 / \textbf{0.446} {\color{green!50!black}($\uparrow$10.3 pp)} \\
% Upfront-concat evaluation &
% 0.165 / \textbf{0.306} {\color{green!50!black}($\uparrow$14.1 pp)} &
% 0.343 / \textbf{0.387} {\color{green!50!black}($\uparrow$4.4 pp)} \\
% First sub-question &
% 0.554 / \textbf{0.588} {\color{green!50!black}($\uparrow$3.5 pp)} &
% 0.620 / \textbf{0.631} {\color{green!50!black}($\uparrow$1.1 pp)} \\
% \bottomrule
% \end{tabular}
% \caption{Controls separating mid-response target reassignment from decomposed-content exposure. Cells report the relevant Search-R1 control / \textsc{SteerSearch} EM. For concat-chain training, only Search-R1 is retrained on the matched decomposed pool with chains concatenated upfront; \textsc{SteerSearch} uses its standard injected-trajectory training.}
% \label{tab:matched-no-injection}
% \end{table*}

\begin{table*}[!t]
\centering
\small
\begin{tabular}{lcccccc}
\toprule
& \multicolumn{3}{c}{\textbf{Qwen2.5-3B-Instruct}}
& \multicolumn{3}{c}{\textbf{Qwen3-4B-Instruct-2507}} \\
\cmidrule(lr){2-4} \cmidrule(lr){5-7}
\textbf{Control}
& \textbf{Search-R1} & \textbf{\textsc{SteerSearch}} & \textbf{$\Delta$ (pp)}
& \textbf{Search-R1} & \textbf{\textsc{SteerSearch}} & \textbf{$\Delta$ (pp)} \\
\midrule
Concat-chain training
& 0.173 & \textbf{0.313} & {\color{green!50!black}+14.0}
& 0.343 & \textbf{0.446} & {\color{green!50!black}+10.3} \\

Upfront-concat evaluation
& 0.165 & \textbf{0.306} & {\color{green!50!black}+14.1}
& 0.343 & \textbf{0.387} & {\color{green!50!black}+4.4} \\

First sub-question
& 0.554 & \textbf{0.588} & {\color{green!50!black}+3.5}
& 0.620 & \textbf{0.631} & {\color{green!50!black}+1.1} \\
\bottomrule
\end{tabular}
\caption{
Controls separating mid-response target reassignment from decomposed-content exposure on \textsc{SteerSearch-Bench}. The table reports average exact-match (EM) scores for Search-R1 and \textsc{SteerSearch}.
}
\label{tab:matched-no-injection}
\end{table*}

As shown in Table~\ref{tab:matched-no-injection}, matching Search-R1 to the decomposed training content substantially strengthens the baseline but does not close the gap on \textsc{SteerSearch-Bench}. \textsc{SteerSearch} retains advantages of 14.0 percentage points on Qwen2.5-3B-Instruct and 10.3 percentage points on Qwen3-4B-Instruct-2507. The residual difference therefore cannot be explained by decomposed multi-hop exposure alone and supports the contribution of presenting target changes as interventions within an active trajectory.

\subsection{Training Dynamics and Search Allocation}
\label{subsec:training-dynamics}

\begin{figure}[t]
\centering
\includegraphics[width=\columnwidth]{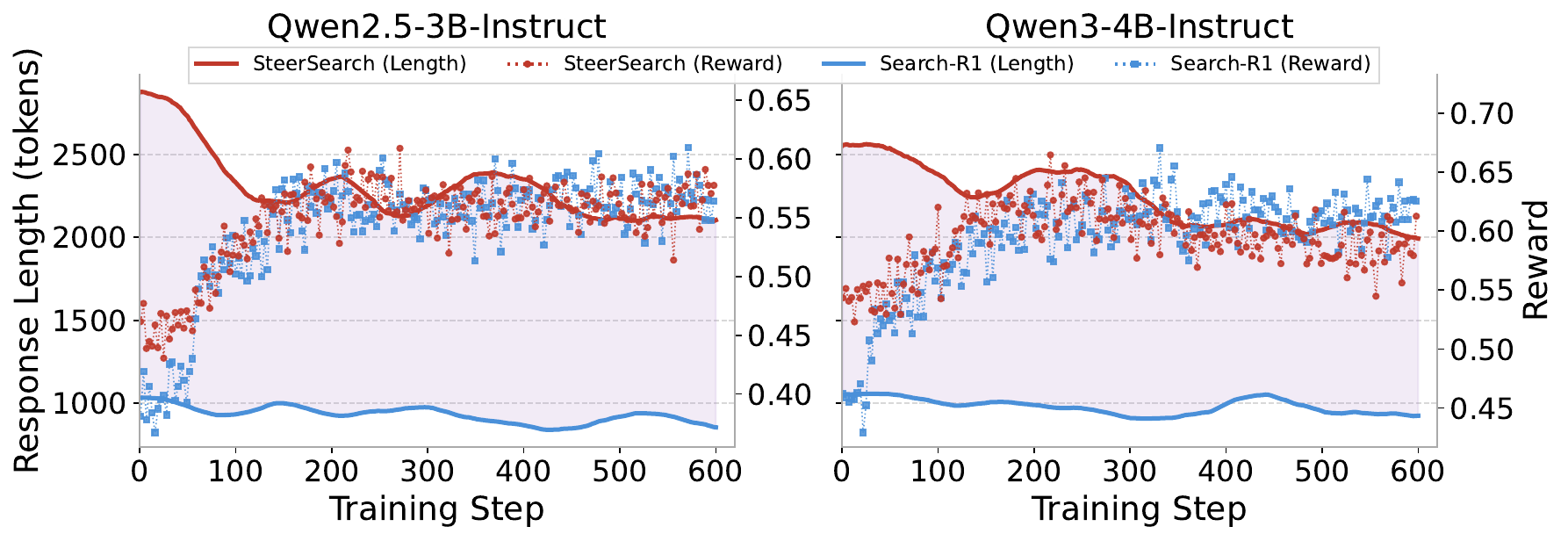}
\vspace{-6mm}
\caption{Response length (solid lines, left axis) and training reward (dotted lines, right axis) for \textsc{SteerSearch} and Search-R1 on Qwen2.5-3B-Instruct and Qwen3-4B-Instruct.}
\label{fig:response-length}
\end{figure}

Figure~\ref{fig:response-length} compares PPO reward and response length for Search-R1 and \textsc{SteerSearch}. Both methods improve over training, although \textsc{SteerSearch} exhibits greater variability because each rollout may contain multiple intervention, reasoning, and retrieval phases before a single terminal answer is evaluated.

A notable difference is that \textsc{SteerSearch} reward improves while average response length decreases. Its trajectories begin substantially longer because they contain multiple interaction phases, but become progressively shorter during training without a corresponding decline in reward. The improvement therefore does not arise from generating increasingly long reasoning trajectories; instead, the policy learns to complete the multi-phase interaction more compactly.

\begin{figure}[t]
\centering
\includegraphics[width=\columnwidth]{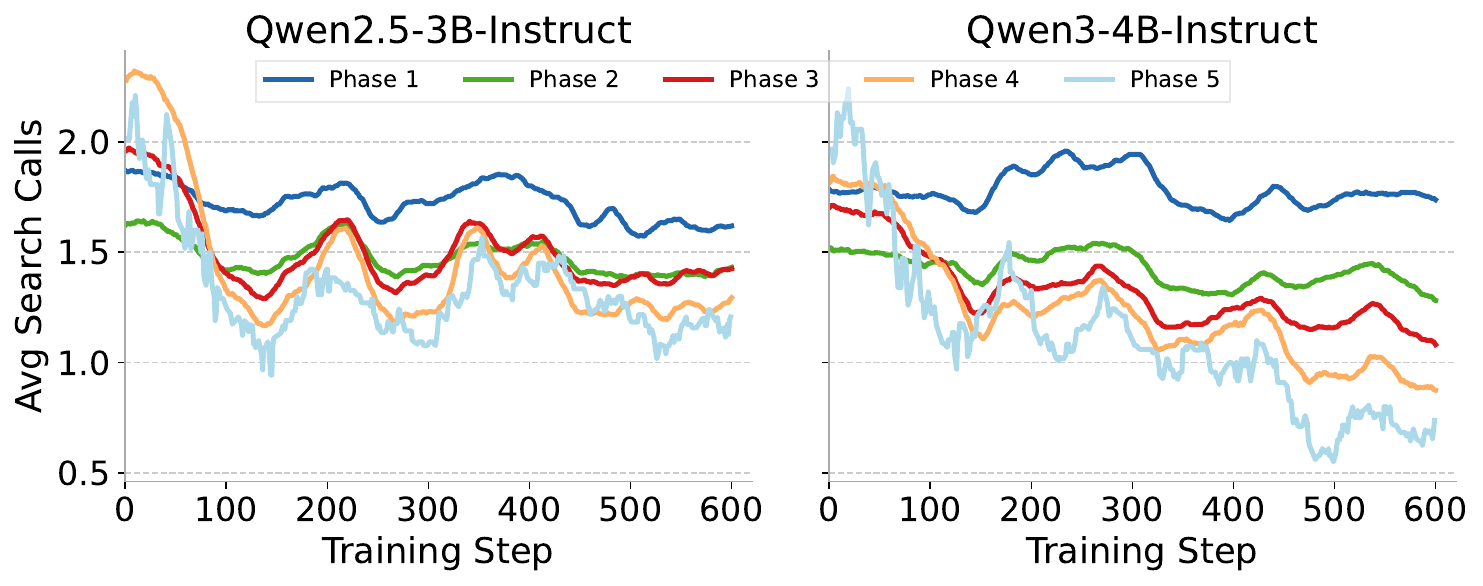}
\vspace{-6mm}
\caption{Average search calls per injection phase during \textsc{SteerSearch} training.}
\label{fig:search-per-phase}
\end{figure}

Figure~\ref{fig:search-per-phase} further separates retrieval behavior by intervention phase. Search usage in the initial phase remains comparatively stable, whereas later phases require progressively fewer retrieval calls during training. This pattern is inconsistent with a uniform reduction in search behavior: the model continues to retrieve when little evidence has been accumulated but becomes more selective after earlier phases have already established useful context.

This behavior is consistent with context-preserving target reassignment. Later interventions often depend on entities or relations established earlier in the trajectory, so updating the active answer target does not necessarily require restarting retrieval. The policy increasingly reuses accumulated evidence and reserves additional search for information required by the new goal.

\subsection{Intermediate Context and Intervention Depth}
\label{subsec:intervention-depth}

We further examine whether the content retained from earlier phases affects target reassignment. Table~\ref{tab:intermediate-answer-ablation} compares the full \textsc{SteerSearch} rollout with variants that restore intermediate reasoning and answers that are removed in the main configuration.

\begin{table*}[t]
\centering
\small
\begin{tabular}{lcccc}
\toprule
& \multicolumn{2}{c}{\textbf{Qwen2.5-3B-Instruct}} &
\multicolumn{2}{c}{\textbf{Qwen3-4B-Instruct-2507}} \\
\cmidrule(lr){2-3}\cmidrule(lr){4-5}
\textbf{Configuration} &
\textbf{\textsc{SteerSearch-Bench}} & \textbf{Standard} &
\textbf{\textsc{SteerSearch-Bench}} & \textbf{Standard} \\
\midrule

Full \textsc{SteerSearch}
& \textbf{0.313}
& \textbf{0.422}
& \textbf{0.446}
& \textbf{0.473} \\

Keep reasoning, remove answers
& 0.303 {\color{red!60!black}($\downarrow$1.0 pp)}
& 0.414 {\color{red!60!black}($\downarrow$0.8 pp)}
& 0.426 {\color{red!60!black}($\downarrow$2.0 pp)}
& 0.466 {\color{red!60!black}($\downarrow$0.7 pp)} \\

Keep reasoning and answers
& 0.285 {\color{red!60!black}($\downarrow$2.8 pp)}
& 0.400 {\color{red!60!black}($\downarrow$2.2 pp)}
& 0.371 {\color{red!60!black}($\downarrow$7.5 pp)}
& 0.456 {\color{red!60!black}($\downarrow$1.7 pp)} \\

\bottomrule
\end{tabular}
\caption{Effect of retaining intermediate context from earlier phases. Red values show absolute degradation relative to the full \textsc{SteerSearch} configuration. Retaining stale intermediate answers causes a larger reduction than retaining reasoning traces alone, particularly on \textsc{SteerSearch-Bench}.}
\label{tab:intermediate-answer-ablation}
\end{table*}

Retaining reasoning while continuing to remove intermediate answers reduces \textsc{SteerSearch-Bench} performance by 1.0 percentage point on Qwen2.5 and 2.0 points on Qwen3. Retaining both reasoning and completed answers produces larger drops of 2.8 and 7.5 points, respectively. The stronger effect of stale answers is consistent with answer-target anchoring rather than sequence length alone: a completed answer explicitly represents an earlier objective and can continue to influence generation after that objective has been superseded. This remains a partial component ablation because randomized phase budgets are not independently isolated.

\begin{figure}[t]
\centering
\includegraphics[width=\columnwidth]{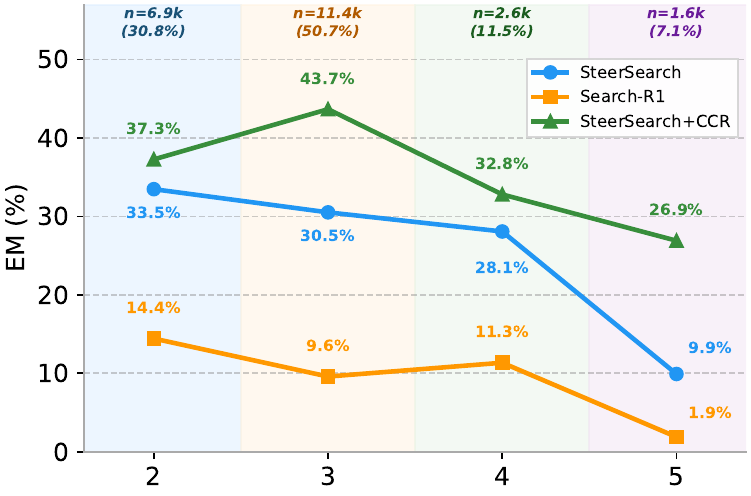}
\vspace{-6mm}
\caption{Exact match by number of injected sub-questions per sample for Qwen2.5-3B-Instruct. Shaded bands mark each chain-length group, with sample counts shown above each band.}
\label{fig:nsubq-accuracy}
\end{figure}

Figure~\ref{fig:nsubq-accuracy} examines performance as the number of injected sub-questions increases. Both methods become less accurate on deeper interaction chains, but \textsc{SteerSearch} remains stronger than Search-R1 across intervention depths. Longer chains simultaneously increase context length and the number of superseded goals represented in the trajectory, making later target updates increasingly difficult.

CCR is particularly helpful on deeper chains. For Qwen2.5, the five-sub-question score increases from 9.9\% without CCR to 26.9\% with CCR. This is consistent with the pivot analysis in Section~\ref{subsec:analysis}: as more previous goals accumulate in context, direct supervision of post-intervention alignment becomes increasingly useful.

\section{CCR Analysis}
\label{app:appendix_ccr}

This appendix provides the CCR configurations and full ablation results omitted from the main paper. We vary the construction of the user context, the reasoning span used to compute the contrastive signal, the reward-assignment strategy, and the frozen similarity encoder. We additionally compare CCR with IGPO and report a three-seed robustness experiment for one CCR configuration. All CCR configurations use $w_{\mathrm{ccr}}=0.1$ and the SFT-4k/PPO-40k training budget. The configuration sweep consists of single-run comparisons and is used primarily to study reward design rather than establish variance-controlled differences between every variant.

\subsection{CCR Configuration Details}
\label{subsec:ccr-config}

CCR measures whether reasoning after an intervention becomes more aligned with the updated user goal than with the preceding context. The evaluated configurations differ in how this contrast is constructed and where the resulting auxiliary reward is applied.

\textbf{Context Variants.}
The \textit{individual} variant compares consecutive user goals,
\[
p_{k+1}=u_{k+1},
\qquad
n_k=u_k,
\]
while the \textit{cumulative} variant compares the user-goal history before and after the intervention,
\[
p_{k+1}=u_1\oplus\cdots\oplus u_{k+1},
\qquad
n_k=u_1\oplus\cdots\oplus u_k,
\]
where $\oplus$ denotes concatenation. Individual CCR therefore provides a local contrast between adjacent goals, whereas cumulative CCR retains earlier user context that may be required to interpret later interventions.

\textbf{Reasoning Scope and Reward Assignment.}
The \textit{Last-think} configuration computes CCR from the final reasoning block generated after each intervention. \textit{All-thinks} instead scores every reasoning block within that phase, providing denser post-intervention supervision. In the \textit{Terminal} configuration, CCR is computed across the reasoning blocks associated with each intervention phase and the accumulated phase rewards are assigned to the final answer token. Tables~\ref{tab:ccr_ablation_ss} and~\ref{tab:ccr_ablation_sr1} use this shorthand for the evaluated configurations; these experiments are not intended as a complete Cartesian sweep over all possible scoring and assignment choices.

\textbf{Embedding Model.}
Unless otherwise specified, CCR similarity is computed using the frozen \texttt{intfloat/e5-base-v2} encoder~\cite{wang2024textembeddingsweaklysupervisedcontrastive}. We additionally evaluate \texttt{Qwen3-Embedding-0.6B}~\cite{zhang2025qwen3embeddingadvancingtext}, a text embedding model built on Qwen3 foundation models and trained for diverse embedding and ranking tasks. Since our policy models generate the reasoning traces being evaluated, we hypothesize that representations from a Qwen-based encoder may provide a useful alternative for measuring alignment between these traces and the evolving user context. This is a motivation for the encoder comparison, rather than an assumption that shared model-family membership guarantees better alignment. E5 is also a general-purpose embedding model trained on heterogeneous text pairs, including retrieval and natural-language inference data~\cite{wang2024textembeddingsweaklysupervisedcontrastive}. The retriever remains unchanged in all experiments; Qwen3-Embedding is used only to compute the CCR similarity signal.

% We separately repeat Individual Last-think CCR with E5 across three Qwen2.5 runs. On \textsc{SteerSearch-Bench}, the CCR runs obtain $0.390$, $0.374$, and $0.384$, yielding $0.383 \pm 0.008$, compared with $0.345 \pm 0.028$ for the no-CCR runs. On the Standard Benchmark, the corresponding CCR scores are $0.401$, $0.395$, and $0.419$, yielding $0.405 \pm 0.012$, compared with $0.411 \pm 0.010$ without CCR. The CCR and no-CCR groups are not matched seed by seed, so the difference may include ordinary training variation in addition to the reward effect. These repeats also do not establish the variance of denser CCR assignments or the Qwen3 configurations.

\subsection{Seed Sensitivity of CCR}
\label{subsec:ccr-seed-sensitivity}

We additionally evaluate one CCR configuration across three independent Qwen2.5-3B-Instruct runs: Individual context with Last-think assignment and the E5 encoder. Table~\ref{tab:ccr-seed-sensitivity} reports mean EM and sample standard deviation across runs.  On \textsc{SteerSearch-Bench}, this CCR configuration achieves a mean EM of $0.383$, compared with $0.345$ for the no-CCR baseline, a difference of 3.8 percentage points. On the Standard Benchmark, the corresponding means are $0.405$ and $0.411$, respectively. Since the CCR and no-CCR runs were not matched seed by seed, these differences are descriptive and may also reflect ordinary training variation.  This experiment provides a robustness check for the specific CCR configuration evaluated; it does not establish seed robustness for other CCR variants or for the cumulative-terminal CCR configuration with Qwen3-Embedding reported in the main tables. Due to computational constraints, the remaining CCR ablations are based on single runs.

\begin{table}[h]
    \centering
    \small
    \setlength{\tabcolsep}{5pt}
    \begin{tabular}{lcc}
    \toprule
    Benchmark & CCR & No-CCR \\
    \midrule
    \textsc{SteerSearch-Bench}
    & $0.383 \pm 0.008$
    & $0.345 \pm 0.028$ \\
    Standard Benchmark
    & $0.405 \pm 0.012$
    & $0.411 \pm 0.010$ \\
    \bottomrule
    \end{tabular}
    \caption{Seed sensitivity for Individual-context, Last-think CCR
    with the E5 encoder on Qwen2.5-3B-Instruct. Entries report mean
    EM $\pm$ sample standard deviation across three runs.}
    \label{tab:ccr-seed-sensitivity}
\end{table}

\textbf{Comparison with IGPO.}
We compare CCR against IGPO~\cite{wang2026informationgainbasedpolicyoptimization}, an alternative dense reward based on marginal gains in ground-truth confidence. IGPO is originally optimized with Group Relative Policy Optimization (GRPO)~\cite{shao2024deepseekmathpushinglimitsmathematical}; here, we apply the same reward signal within the PPO training loop to isolate the reward design from the optimizer. Unlike CCR, IGPO assumes that the answer target remains fixed throughout the trajectory. Under mid-response steering, however, the operative user goal may change after each intervention, so progress toward a previous target need not correspond to progress toward the final answer.

\begin{table*}[!t]
\centering\small
\resizebox{\textwidth}{!}{
\begin{tabular}{llccccc}
\toprule
\textbf{Method} & \textbf{Reward} & \textbf{HotpotQA$^\dagger$} & \textbf{2WikiMQA$^*$} & \textbf{MuSiQue$^*$} & \textbf{Bamboogle$^*$} & \textbf{\textbf{Avg.}} \\
\midrule
\multicolumn{6}{l}{\textit{Qwen2.5-3B-Instruct}} \\
\rowcolor{gray!8}
Search-R1 & Outcome + Format & 0.192 & 0.063 & 0.078 & 0.112 & 0.111 \\
\rowcolor{gray!8}
\textsc{SteerSearch} & Outcome + Format & \textbf{0.376} & 0.260 & \textbf{0.241} & 0.376 & \textbf{0.313} \\
\rowcolor{gray!8}
\textsc{SteerSearch} & IGPO & 0.358 & \textbf{0.263} & 0.220 & \textbf{0.384} & 0.306 \\
\cdashline{1-7}
\noalign{\vskip 2pt}
\rowcolor{orange!8}
\textsc{SteerSearch} (Individual) & CCR (Last-think) & 0.429 & 0.361 & \textbf{0.268} & 0.480 & 0.384 \\
\rowcolor{orange!8}
 & CCR (All-thinks) & \textbf{0.437} & \textbf{0.378} & 0.255 & \textbf{0.488} & \textbf{0.389} \\
\rowcolor{orange!8}
 & CCR (Terminal) & 0.411 & 0.337 & 0.252 & 0.448 & 0.362 \\
\cdashline{1-7}
\noalign{\vskip 2pt}
\rowcolor{blue!8}
\textsc{SteerSearch} (Cumulative) & CCR (Last-think) & 0.409 & 0.328 & \textbf{0.266} & 0.464 & 0.367 \\
\rowcolor{blue!8}
 & CCR (Terminal) & 0.413 & 0.344 & 0.260 & \textbf{0.504} & 0.380 \\
\rowcolor{blue!8}
 & CCR (Terminal) \textit{(qwenemb)} & \textbf{0.435} & \textbf{0.391} & 0.264 & 0.496 & \textbf{0.396} \\
\midrule
\multicolumn{6}{l}{\textit{Qwen3-4B-Instruct-2507}} \\
\rowcolor{gray!8}
Search-R1 & Outcome + Format & 0.423 & 0.349 & 0.250 & 0.440 & 0.366 \\
\rowcolor{gray!8}
\textsc{SteerSearch} & Outcome + Format & \textbf{0.472} & \textbf{0.457} & \textbf{0.305} & \textbf{0.552} & \textbf{0.446} \\
\cdashline{1-7}
\noalign{\vskip 2pt}
\rowcolor{orange!8}
\textsc{SteerSearch} (Individual) & CCR (Last-think) & \textbf{0.471} & \textbf{0.446} & \textbf{0.300} & 0.480 & \textbf{0.424} \\
\rowcolor{orange!8}
 & CCR (All-thinks) & 0.456 & 0.389 & 0.275 & 0.480 & 0.400 \\
\rowcolor{orange!8}
 & CCR (Terminal) & 0.470 & 0.421 & 0.287 & \textbf{0.512} & 0.422 \\
\cdashline{1-7}
\noalign{\vskip 2pt}
\rowcolor{blue!8}
\textsc{SteerSearch} (Cumulative) & CCR (Last-think) & 0.473 & 0.427 & 0.304 & 0.520 & 0.431 \\
\rowcolor{blue!8}
 & CCR (Terminal) & \textbf{0.477} & 0.472 & 0.306 & 0.520 & 0.444 \\
\rowcolor{blue!8}
 & CCR (Terminal) \textit{(qwenemb)} & \textbf{0.477} & \textbf{0.474} & \textbf{0.308} & \textbf{0.544} & \textbf{0.451} \\
\bottomrule
\end{tabular}
}
\caption{\textbf{CCR ablation on \textsc{SteerSearch-Bench}}. $\dagger$~in-domain, $*$~out-of-domain. All CCR variants use $w_{\mathrm{ccr}}=0.1$ and SFT-4k/PPO-40k. Rows without \textit{(qwenemb)} use the E5 encoder for CCR similarity; \textit{(qwenemb)} rows use the Qwen-embedding encoder. \cellcolor{gray!8}Gray = baselines, \cellcolor{blue!8}blue = cumulative CCR, \cellcolor{orange!8}orange = individual CCR. \textbf{Bold} = best per column within each group.}
\label{tab:ccr_ablation_ss}
\end{table*}
\begin{table*}[h]
\centering\small
\resizebox{\textwidth}{!}{
\begin{tabular}{llcccccccc}
\toprule
\textbf{Method} & \textbf{Reward} & \multicolumn{3}{c}{\textbf{General QA}} & \multicolumn{4}{c}{\textbf{Multi-Hop QA}} \\
\cmidrule(lr){3-5}\cmidrule(lr){6-9}
& & \textbf{NQ$^\dagger$} & \textbf{TriviaQA$^*$} & \textbf{PopQA$^*$} & \textbf{HotpotQA$^\dagger$} & \textbf{2WikiMQA$^*$} & \textbf{MuSiQue$^*$} & \textbf{Bamboogle$^*$} & \textbf{\textbf{Avg.}} \\
\midrule
\multicolumn{9}{l}{\textit{Qwen2.5-3B-Instruct}} \\
\rowcolor{gray!8}
Search-R1 & Outcome + Format & 0.447 & 0.602 & 0.433 & 0.429 & 0.424 & 0.175 & 0.328 & 0.405 \\
\rowcolor{gray!8}
\textsc{SteerSearch} & Outcome + Format & \textbf{0.450} & \textbf{0.615} & \textbf{0.442} & 0.442 & 0.409 & \textbf{0.209} & \textbf{0.384} & \textbf{0.422} \\
\rowcolor{gray!8}
\textsc{SteerSearch} & IGPO & 0.418 & 0.608 & 0.419 & \textbf{0.448} & \textbf{0.428} & 0.188 & 0.376 & 0.412 \\
\cdashline{1-10}
\noalign{\vskip 2pt}
\rowcolor{orange!8}
\textsc{SteerSearch} (Individual) & CCR (Last-think) & 0.456 & \textbf{0.617} & 0.444 & \textbf{0.426} & \textbf{0.413} & 0.163 & \textbf{0.416} & \textbf{0.419} \\
\rowcolor{orange!8}
 & CCR (All-thinks) & 0.448 & 0.605 & 0.425 & 0.410 & 0.388 & 0.147 & 0.336 & 0.394 \\
\rowcolor{orange!8}
 & CCR (Terminal) & \textbf{0.458} & \textbf{0.617} & \textbf{0.455} & 0.422 & 0.394 & \textbf{0.175} & 0.344 & 0.409 \\
\cdashline{1-10}
\noalign{\vskip 2pt}
\rowcolor{blue!8}
\textsc{SteerSearch} (Cumulative) & CCR (Last-think) & 0.450 & 0.618 & 0.438 & 0.431 & \textbf{0.426} & \textbf{0.181} & \textbf{0.376} & \textbf{0.417} \\
\rowcolor{blue!8}
 & CCR (Terminal) & 0.446 & \textbf{0.621} & \textbf{0.460} & 0.431 & 0.423 & 0.179 & 0.352 & 0.416 \\
\rowcolor{blue!8}
 & CCR (Terminal) \textit{(qwenemb)} & \textbf{0.456} & \textbf{0.621} & 0.447 & \textbf{0.437} & \textbf{0.426} & 0.165 & 0.328 & 0.411 \\
\midrule
\multicolumn{9}{l}{\textit{Qwen3-4B-Instruct-2507}} \\
\rowcolor{gray!8}
Search-R1 & Outcome + Format & \textbf{0.496} & \textbf{0.660} & \textbf{0.487} & \textbf{0.481} & 0.457 & 0.233 & \textbf{0.488} & 0.472 \\
\rowcolor{gray!8}
\textsc{SteerSearch} & Outcome + Format & 0.491 & 0.652 & 0.479 & 0.475 & \textbf{0.498} & \textbf{0.239} & 0.480 & \textbf{0.473} \\
\cdashline{1-10}
\noalign{\vskip 2pt}
\rowcolor{orange!8}
\textsc{SteerSearch} (Individual) & CCR (Last-think) & 0.480 & 0.649 & 0.469 & \textbf{0.490} & \textbf{0.488} & \textbf{0.226} & 0.472 & 0.468 \\
\rowcolor{orange!8}
 & CCR (All-thinks) & \textbf{0.485} & 0.646 & 0.472 & 0.476 & 0.455 & 0.208 & 0.416 & 0.451 \\
\rowcolor{orange!8}
 & CCR (Terminal) & \textbf{0.485} & \textbf{0.661} & \textbf{0.473} & 0.477 & 0.473 & 0.210 & \textbf{0.512} & \textbf{0.470} \\
\cdashline{1-10}
\noalign{\vskip 2pt}
\rowcolor{blue!8}
\textsc{SteerSearch} (Cumulative) & CCR (Last-think) & \textbf{0.488} & 0.654 & 0.475 & \textbf{0.482} & 0.457 & 0.209 & 0.472 & 0.462 \\
\rowcolor{blue!8}
 & CCR (Terminal) & 0.484 & 0.655 & 0.468 & 0.470 & 0.472 & \textbf{0.223} & 0.488 & 0.466 \\
\rowcolor{blue!8}
 & CCR (Terminal) \textit{(qwenemb)} & 0.487 & \textbf{0.664} & \textbf{0.480} & 0.480 & \textbf{0.498} & 0.220 & \textbf{0.512} & \textbf{0.477} \\
\bottomrule
\end{tabular}
}
\caption{\textbf{CCR ablation on the Standard Benchmark.} $\dagger$~in-domain, $*$~out-of-domain. All CCR variants use $w_{\mathrm{ccr}}=0.1$ and SFT-4k/PPO-40k. Rows without \textit{(qwenemb)} use the E5 encoder for CCR similarity; \textit{(qwenemb)} rows use the Qwen-embedding encoder. \cellcolor{gray!8}Gray = baselines, \cellcolor{blue!8}blue = cumulative CCR, \cellcolor{orange!8}orange = individual CCR. \textbf{Bold} = best per column within each group.}
\label{tab:ccr_ablation_sr1}
\end{table*}

\subsection{CCR Ablation Results}
\label{subsec:ccr-ablation}

\begin{figure}[h]
\centering
\includegraphics[width=\columnwidth]{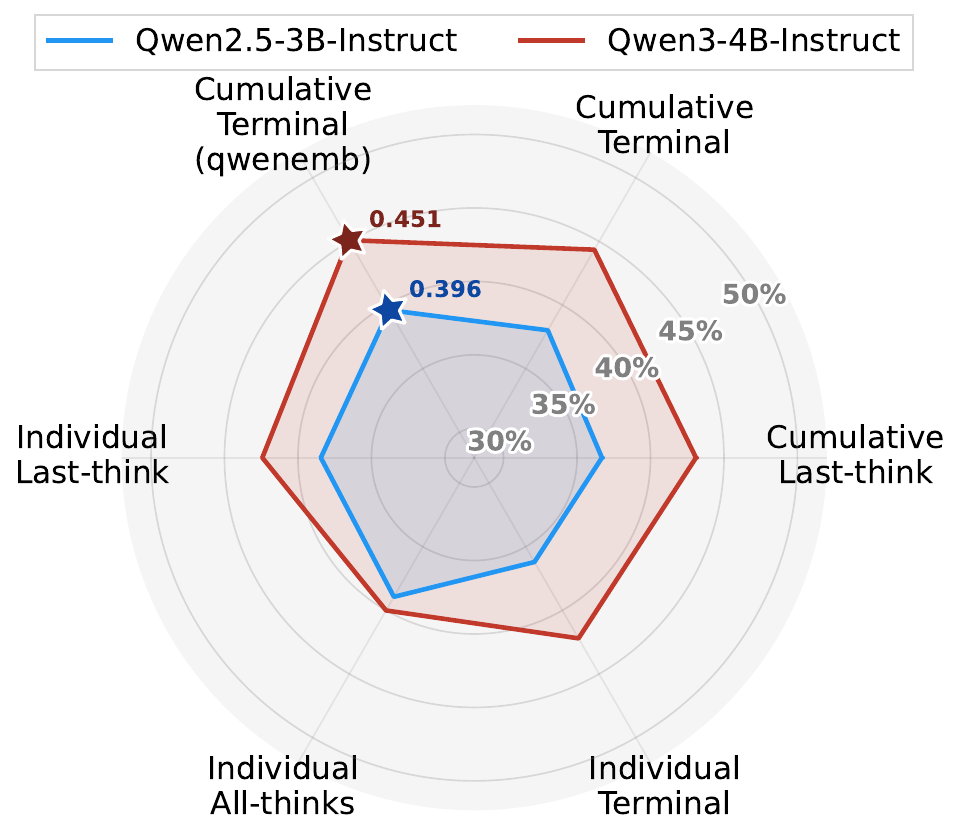}
\caption{Average EM on \textsc{SteerSearch-Bench} across CCR configurations. Stars mark the best-performing configuration for each model.}
\label{fig:ccr-radar}
\end{figure}

Table~\ref{tab:ccr_ablation_ss} and Figure~\ref{fig:ccr-radar} report the CCR configuration sweep on \textsc{SteerSearch-Bench}. Among the E5-based variants, cumulative-terminal CCR provides the strongest aggregate performance across the two evaluation settings and the strongest average performance on \textsc{SteerSearch-Bench}. We select this configuration for a targeted comparison of frozen similarity encoders. Replacing E5 with \texttt{Qwen3-Embedding-0.6B} improves steering accuracy for both models in the evaluated runs, motivating its use for the final CCR results. This selection is based on observed performance within this configuration rather than a claim of general encoder superiority.

The effect of CCR depends on both the base model and the reward design. Qwen2.5 benefits more consistently from the auxiliary signal, whereas Qwen3 shows smaller and less consistent changes across configurations. Denser supervision is also not uniformly beneficial: All-thinks does not consistently outperform more selective or terminal assignment, suggesting that every intermediate reasoning block is not equally informative about whether the model has updated its answer target.

Within cumulative-terminal CCR, replacing E5 with Qwen3-Embedding improves \textsc{SteerSearch-Bench} EM from 38.0\% to 39.6\% for Qwen2.5 and from 44.4\% to 45.1\% for Qwen3. On the Standard Benchmark, the effect is mixed. These results motivate our choice of Qwen3-Embedding for the final CCR configuration, although the comparison is limited to single runs and one CCR configuration.

Table~\ref{tab:ccr_ablation_sr1} reports the corresponding configurations on the Standard Benchmark. In contrast to the steering setting, CCR does not produce consistent gains when the user goal remains fixed. Most configurations remain close to the no-CCR \textsc{SteerSearch} baseline, and changes vary in direction across models and reward designs. This supports the narrower interpretation of CCR as an auxiliary signal for target reassignment rather than a general mechanism for improving retrieval-augmented question answering. 

Taken together, the ablations indicate that injected-trajectory training provides the primary steering improvement, while CCR can further encourage post-intervention alignment, its contribution is configuration-dependent. The improvement also extends to Gemma3-4B-Instruct in our cross-family evaluation~\ref{subsec:cross-family-generalization}, suggesting that the observed benefit is not restricted to Qwen policy models. However, these results do not isolate whether the gains arise from the encoder's training, its underlying model family, or other differences between the representations.

\subsection{Computational Cost of Contrastive Context Rewards}
\label{subsec:ccr-computational-cost}

\begin{figure}[h]
    \centering
    \includegraphics[width=\columnwidth]{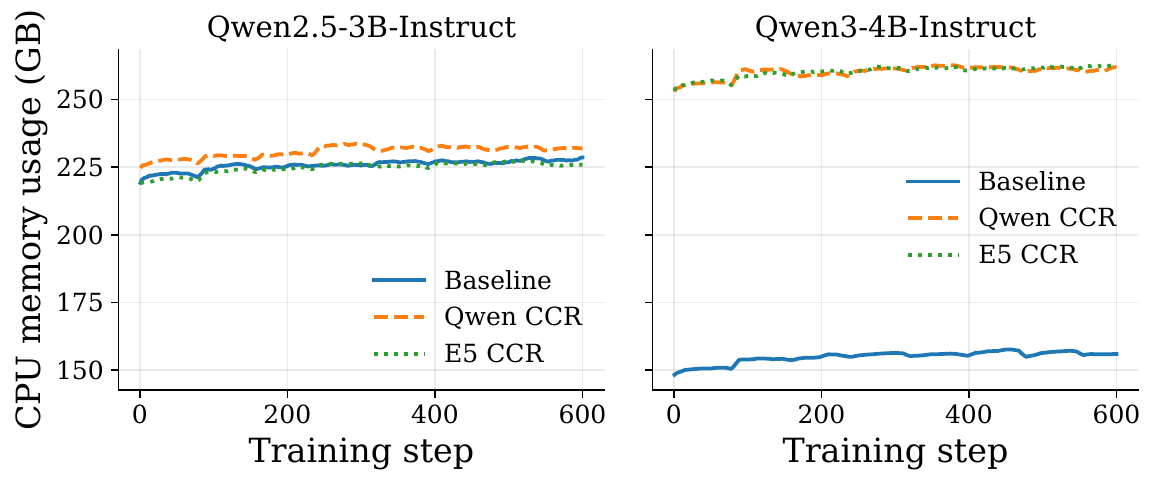}
    \caption{
        System RAM usage during training for the baseline, Qwen-based CCR, and E5-based CCR. Curves are smoothed using a 10-step rolling mean.
    }
    \label{fig:ccr-cpu-memory}
\end{figure}

\begin{figure}[h]
    \centering
    \includegraphics[width=\columnwidth]{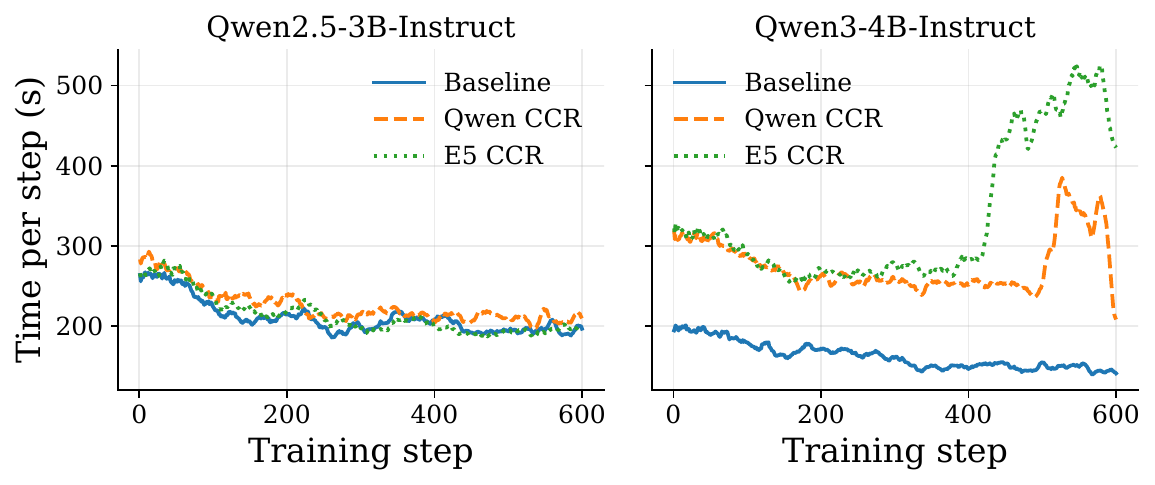}
    \caption{
        Per-step training time for the baseline, Qwen-based CCR, and E5-based CCR. Curves are smoothed using a 10-step rolling mean.
    }
    \label{fig:ccr-step-time}
\end{figure}

We analyze the computational overhead of Contrastive Context Reward (CCR) across six training runs, comparing the baseline with Qwen-based and E5-based reward models for Qwen2.5-3B and Qwen3-4B. Each run comprises 600 training steps across eight GPUs. Figures~\ref{fig:ccr-cpu-memory} and~\ref{fig:ccr-step-time} show system RAM usage and per-step training time, respectively.

For Qwen2.5-3B, total training time remains comparable across configurations: 40.5 hours for the baseline, 40.4 hours with Qwen-based CCR, and 38.3 hours with E5-based CCR. Mean step time increases by 7.1\% and 1.4\%, respectively. For Qwen3-4B, training time increases from 29.4 hours for the baseline to 48.9 hours with Qwen-based CCR and 60.0 hours with E5-based CCR, representing overheads of 66.2\% and 104.0\%. Mean step time increases by 69.6\% and 106.4\%, respectively.

Peak GPU memory usage remains broadly comparable across configurations, ranging from 90--97 GB reserved and 88--94 GB allocated.

\onecolumn
\clearpage
\section{Examples from the Benchmarks}

\subsection{Standard Benchmark vs.\ \textsc{SteerSearch-Bench}}
\label{subsec:standard_vs_steer_bench}

Table~\ref{tab:bench-comparison} shows one representative question per source dataset, illustrating the structural difference between the Standard Benchmark and \textsc{SteerSearch-Bench} evaluation settings. \textsc{SteerSearch-Bench} covers the four multi-hop datasets; NQ, TriviaQA, and PopQA are omitted as they provide no multi-hop structure from which to derive injection chains. Dataset composition statistics are in Table~\ref{tab:bench-stats} (Section~\ref{sec:benchmarks}).

\begin{table}[H]
\resizebox{\textwidth}{!}{
\small
\begin{tabular}{p{0.09\textwidth} p{0.43\textwidth} p{0.43\textwidth}}
\toprule
\textbf{Dataset} & \textbf{Standard Benchmark} & \textbf{\textsc{SteerSearch-Bench}} \\
\midrule
\multirow{2}{*}{\textbf{HotpotQA}} & \begin{minipage}[t]{0.43\textwidth}\raggedright \textbf{Q:} Are Local H and For Against both from the United States? \newline \textbf{GT:} \textit{yes}\end{minipage} & \begin{minipage}[t]{0.43\textwidth}\raggedright \textbf{Q:} Are Local H and For Against both from the United States? \newline \tuser{[inj\,1] }Local H band origin country \newline \tuser{[inj\,2] }For Against band origin country \newline \tuser{[inj\,3] }Are Local H and For Against both from the United States? \newline \textbf{GT:} \textit{yes}\end{minipage} \\
\addlinespace[1pt]\cline{2-3}\addlinespace[1pt]
 & \begin{minipage}[t]{0.43\textwidth}\raggedright \textbf{Q:} When was Poison's album "Shut Up, Make Love" released? \newline \textbf{GT:} \textit{2000}\end{minipage} & \begin{minipage}[t]{0.43\textwidth}\raggedright \textbf{Q:} When was Poison's album "Shut Up, Make Love" released? \newline \tuser{[inj\,1] }Poison discography albums \newline \tuser{[inj\,2] }Poison album Shut Up, Make Love \newline \tuser{[inj\,3] }Shut Up, Make Love release date \newline \textbf{GT:} \textit{2000}\end{minipage} \\
\midrule
\multirow{2}{*}{\textbf{2WikiMQA}} & \begin{minipage}[t]{0.43\textwidth}\raggedright \textbf{Q:} Which film has the director born first, The Raven'S Dance or Keïta! L'Héritage Du Griot? \newline \textbf{GT:} \textit{The Raven's Dance}\end{minipage} & \begin{minipage}[t]{0.43\textwidth}\raggedright \textbf{Q:} Which film has the director born first, The Raven'S Dance or Keïta! L'Héritage Du Griot? \newline \tuser{[inj\,1] }Who directed Keïta! L'Héritage Du Griot? \newline \tuser{[inj\,2] }Who directed the film The Raven'S Dance? \newline \tuser{[inj\,3] }Which film has the earlier-born director, The Raven'S Dance or Keïta! L'Héritage Du Griot? \newline \textbf{GT:} \textit{The Raven's Dance}\end{minipage} \\
\addlinespace[1pt]\cline{2-3}\addlinespace[1pt]
 & \begin{minipage}[t]{0.43\textwidth}\raggedright \textbf{Q:} Who died first, Madame Pasca or James A. Donohoe? \newline \textbf{GT:} \textit{Madame Pasca}\end{minipage} & \begin{minipage}[t]{0.43\textwidth}\raggedright \textbf{Q:} Who died first, Madame Pasca or James A. Donohoe? \newline \tuser{[inj\,1] }When did James A. Donohoe die? \newline \tuser{[inj\,2] }When did the performer known as Madame Pasca die? \newline \tuser{[inj\,3] }Who died first, the performer known as Madame Pasca or James A. Donohoe? \newline \textbf{GT:} \textit{Madame Pasca}\end{minipage} \\
\midrule
\multirow{2}{*}{\textbf{MuSiQue}} & \begin{minipage}[t]{0.43\textwidth}\raggedright \textbf{Q:} What company succeeded the owner of Empire Sports Network? \newline \textbf{GT:} \textit{Time Warner Cable}\end{minipage} & \begin{minipage}[t]{0.43\textwidth}\raggedright \textbf{Q:} What company succeeded the owner of Empire Sports Network? \newline \tuser{[inj\,1] }Empire Sports Network owner \newline \tuser{[inj\,2] }company that succeeded Adelphia Communications Corporation \newline \textbf{GT:} \textit{Time Warner Cable}\end{minipage} \\
\addlinespace[1pt]\cline{2-3}\addlinespace[1pt]
 & \begin{minipage}[t]{0.43\textwidth}\raggedright \textbf{Q:} What is the capital of the county that Pine Springs is located in? \newline \textbf{GT:} \textit{Van Horn}\end{minipage} & \begin{minipage}[t]{0.43\textwidth}\raggedright \textbf{Q:} What is the capital of the county that Pine Springs is located in? \newline \tuser{[inj\,1] }Pine Springs Texas located in which county \newline \tuser{[inj\,2] }What is the county seat of Culberson County, Texas \newline \textbf{GT:} \textit{Van Horn}\end{minipage} \\
\midrule
\multirow{2}{*}{\textbf{Bamboogle}} & \begin{minipage}[t]{0.43\textwidth}\raggedright \textbf{Q:} When did the author of Annabel Lee enlist in the army? \newline \textbf{GT:} \textit{1827}\end{minipage} & \begin{minipage}[t]{0.43\textwidth}\raggedright \textbf{Q:} When did the author of Annabel Lee enlist in the army? \newline \tuser{[inj\,1] }Who wrote "Annabel Lee"? \newline \tuser{[inj\,2] }What year did Edgar Allan Poe enlist in the army? \newline \textbf{GT:} \textit{1827}\end{minipage} \\
\addlinespace[1pt]\cline{2-3}\addlinespace[1pt]
 & \begin{minipage}[t]{0.43\textwidth}\raggedright \textbf{Q:} The material out of which the Great Sphinx of Giza is made of is mainly composed of what mineral? \newline \textbf{GT:} \textit{calcite}\end{minipage} & \begin{minipage}[t]{0.43\textwidth}\raggedright \textbf{Q:} The material out of which the Great Sphinx of Giza is made of is mainly composed of what mineral? \newline \tuser{[inj\,1] }What type of rock is the Great Sphinx of Giza primarily carved from? \newline \tuser{[inj\,2] }Main mineral composition of limestone \newline \textbf{GT:} \textit{calcite}\end{minipage} \\
\bottomrule
\end{tabular}
}
\caption{Representative questions from the Standard Benchmark (multi-hop, no interventions) and \textsc{SteerSearch-Bench} (same question with injected sub-question chains).}
\label{tab:bench-comparison}
\end{table}

\subsection{TopiOCQA}
\label{subsec:topiocqa_samples}

Table~\ref{tab:topiocqa-examples} presents randomly selected conversations from TopiOCQA (validation set), illustrating naturally occurring follow-up questions, coreference, and topic shifts. Unlike the constructed intervention chains in \textsc{SteerSearch-Bench}, these conversations preserve the original sequence of human-authored turns. They illustrate how subsequent questions can depend on earlier conversational context or introduce a new topic, requiring the model to update its active answer target during generation.

\begin{table*}[h]
\centering
\small
\begin{tabular}{p{0.72\textwidth} p{0.22\textwidth}}
\toprule
\textbf{Question and Injected Subquestions} & \textbf{GT} \\
\midrule
\begin{minipage}[t]{0.70\textwidth}\raggedright \textbf{Q:} when will the new dunkirk film be released on dvd \newline \tuser{[inj\,1] }what is this film about? \newline \tuser{[inj\,2] }can you mention a few members of the cast? \newline \tuser{[inj\,3] }where was it shot? \newline \tuser{[inj\,4] }can you tell me anything about the latter place's past? \newline \tuser{[inj\,5] }what does the economy of this place depend upon? \newline \tuser{[inj\,6] }is there any waterbody nearby ? \newline \tuser{[inj\,7] }what kind of marine organisms are found in this sea? \newline \tuser{[inj\,8] }what kind of minerals are found in this sea? \newline \tuser{[inj\,9] }does this place experiences any natural hazards? \newline \tuser{[inj\,10] }are whales captured here? \newline \tuser{[inj\,11] }any particular whale species that was hunted? \newline \tuser{[inj\,12] }where are they found? \newline \tuser{[inj\,13] }what do they feed on?\end{minipage} & \begin{minipage}[t]{0.22\textwidth}\raggedright Benthic crustaceans\end{minipage} \\
\addlinespace[5pt]
\midrule
\begin{minipage}[t]{0.70\textwidth}\raggedright \textbf{Q:} the primary high energy molecule in human metabolism is \newline \tuser{[inj\,1] }what is catabolism? \newline \tuser{[inj\,2] }which hormones are related to it? \newline \tuser{[inj\,3] }what is the emotional response due to the third one? \newline \tuser{[inj\,4] }how does the first hormone related to metabolism of glucose? \newline \tuser{[inj\,5] }is process resisted by insulin? \newline \tuser{[inj\,6] }when was this substance which resists the process discovered? \newline \tuser{[inj\,7] }can it be extracted? \newline \tuser{[inj\,8] }from where? \newline \tuser{[inj\,9] }can we find in plants the last hormone in the list of hormones mentioned above? \newline \tuser{[inj\,10] }is it responsible for any memory related disease? \newline \tuser{[inj\,11] }how does it affect immune system? \newline \tuser{[inj\,12] }is it also found in animals? \newline \tuser{[inj\,13] }what is its function in animals? \newline \tuser{[inj\,14] }can it be given through supplements?\end{minipage} & \begin{minipage}[t]{0.22\textwidth}\raggedright Melatonin is categorized by the US Food and Drug Administration (FDA) as a dietary supplement.\end{minipage} \\
\addlinespace[5pt]
\midrule
\begin{minipage}[t]{0.70\textwidth}\raggedright \textbf{Q:} what does the song hotel california by the eagles mean \newline \tuser{[inj\,1] }what is hotel california? \newline \tuser{[inj\,2] }who is the singer of this song? \newline \tuser{[inj\,3] }when did this song release? \newline \tuser{[inj\,4] }where was it recorded? \newline \tuser{[inj\,5] }when was the first one established? \newline \tuser{[inj\,6] }does it also have branches in other places? \newline \tuser{[inj\,7] }did they also have a fan club? \newline \tuser{[inj\,8] }how did the song perform according to the chart? \newline \tuser{[inj\,9] }can you tell me one interpretation that was made about this song? \newline \tuser{[inj\,10] }did the song belong to a band or a single singer? \newline \tuser{[inj\,11] }did the song win any award? \newline \tuser{[inj\,12] }were there any cover versions of this song?\end{minipage} & \begin{minipage}[t]{0.22\textwidth}\raggedright Yes, flamenco version, The Orb version, etc\end{minipage} \\
\bottomrule
\end{tabular}
\caption{Randomly sampled TopiOCQA examples with injected subquestions and ground-truth answers.}
\label{tab:topiocqa-examples}
\end{table*}

\twocolumn
\section{Case Study: Search-R1 vs.\ \textsc{SteerSearch}}
\label{sec:appendix_casestudy}

Trajectories are shown verbatim from the evaluation outputs of Qwen2.5-3B-Instruct. Tags are colored: \tthink{$\langle$think$\rangle$} reasoning, \tsearch{$\langle$search$\rangle$} query, \tinfo{$\langle$information$\rangle$} retrieved passage (truncated for space), \tanswer{$\langle$answer$\rangle$} final answer.

\subsection{Search-R1 Fails, \textsc{SteerSearch} Succeeds}

Tables~\ref{tab:case1}--\ref{tab:case4} show trajectories where Search-R1 fails to follow the injected sub-question and produces an incorrect final answer, while \textsc{SteerSearch} correctly pivots its answer target upon receiving the injection.

% Example 1
\begin{table*}[htbp]
\small
\setlength{\tabcolsep}{6pt}
\begin{tabular}{p{0.48\textwidth} p{0.48\textwidth}}
\toprule
\textbf{Search-R1} \textcolor{red}{\ding{55}} & \textbf{\textsc{SteerSearch}} \textcolor{ForestGreen}{\ding{51}} \\
\midrule
\multicolumn{2}{p{\textwidth}}{
\textbf{Question:} Which singer released a studio album titled "Unapologetic" (album) and what is their stage name?
\newline \textbf{Injection 1:} What Eminem songs or albums include guest vocals from the singer who released the album "Unapologetic"?
\newline \textbf{Injection 2:} Which Eminem album includes vocals from that singer?
} \\
\midrule
\multicolumn{2}{p{\textwidth}}{
\textbf{Ground truth:} "The Marshall Mathers LP 2"
} \\
\midrule
\begin{minipage}[t]{0.48\textwidth}\raggedright
\tthink{<think>}I need to identify the singer who released the studio album “Unapologetic” and provide their stage name.\tthink{</think>} \newline \tsearch{<search>}studio album "Unapologetic" singer stage name\tsearch{</search>} \newline \tinfo{<information>}Doc 1(Title: Unapologetic) Unapologetic Unapologetic is the seventh studio album by Barbadian singer Rihanna. It was released on November 19, 2012, by Def Jam Recordings and SRP Records. It was recorded between June and November 2012, during promotion of her sixth album, "Talk That Talk" (2011)....\tinfo{</information>} \newline \tuser{[user] }What Eminem songs or albums include guest vocals from the singer who released the album "Unapologetic"? \newline \tthink{<think>}I need to identify the singer who released “Unapologetic” and then find which Eminem songs or albums include guest vocals from that singer.\tthink{</think>} \newline \tsearch{<search>}singer who released "Unapologetic" Eminem guest vocals\tsearch{</search>} \newline \tinfo{<information>}Doc 1(Title: Numb (Rihanna song)) Songs chart. Credits adapted from the liner notes of "Unapologetic", Def Jam Recordings, SRP Records. Numb (Rihanna song) "Numb" is a song by Barbadian singer Rihanna from her seventh studio album "Unapologetic" (2012)....\tinfo{</information>} \newline \tuser{[user] }Which Eminem album includes vocals from that singer? \newline \tthink{<think>}The search shows “Unapologetic” is by Barbadian singer Rihanna. I need to identify an Eminem album that includes guest vocals by Rihanna.\tthink{</think>} \newline \tsearch{<search>}Eminem album guest vocals Rihanna\tsearch{</search>} \newline \tinfo{<information>}Doc 1(Title: Eminem) in the Game by MTV and Emcee of the Year by the online magazine HipHopDX. He and Rihanna again collaborated on "Love the Way You Lie (Part II)", the sequel of their hit single. Unlike the original, Rihanna is the lead vocalist and it is sung from the female perspective....\tinfo{</information>} \newline \tthink{<think>}The search results show “Unapologetic” is by Barbadian singer Rihanna. The search for Eminem albums with guest vocals by Rihanna shows he collaborated on “Love the Way You Lie (Part II)” with Rihanna.\tthink{</think>} \newline \tanswer{<answer>}Rihanna\tanswer{</answer>}
\end{minipage}
 & 
\begin{minipage}[t]{0.48\textwidth}\raggedright
\tthink{<think>}I need to identify the singer who released a studio album titled "Unapologetic" and confirm their stage name.\tthink{</think>} \newline \tsearch{<search>}studio album "Unapologetic" singer stage name\tsearch{</search>} \newline \tinfo{<information>}Doc 1(Title: Unapologetic) Unapologetic Unapologetic is the seventh studio album by Barbadian singer Rihanna. It was released on November 19, 2012, by Def Jam Recordings and SRP Records. It was recorded between June and November 2012, during promotion of her sixth album, "Talk That Talk" (2011)....\tinfo{</information>} \newline \tuser{[user] }What Eminem songs or albums include guest vocals from the singer who released the album "Unapologetic"? \newline \tthink{<think>}I need to identify the singer who released "Unapologetic" (Rihanna) and then determine which of her songs or albums include guest vocals from her.\tthink{</think>} \newline \tsearch{<search>}Rihanna guest vocals on Eminem songs or albums\tsearch{</search>} \newline \tinfo{<information>}Doc 1(Title: Numb (Rihanna song)) Songs chart. Credits adapted from the liner notes of "Unapologetic", Def Jam Recordings, SRP Records. Numb (Rihanna song) "Numb" is a song by Barbadian singer Rihanna from her seventh studio album "Unapologetic" (2012)....\tinfo{</information>} \newline \tuser{[user] }Which Eminem album includes vocals from that singer? \newline \tthink{<think>}I need to identify the singer who released "Unapologetic" (Rihanna) and then determine which Eminem album includes guest vocals from her.\tthink{</think>} \newline \tsearch{<search>}Eminem album with vocals from Rihanna\tsearch{</search>} \newline \tinfo{<information>}Doc 1(Title: Rihanna discography) the Music", "Take a Bow" and "Disturbia". Rihanna made guest appearances on numerous singles, including "Live Your Life" with rapper T.I., and "Run This Town" with Jay-Z and Kanye West....\tinfo{</information>} \newline \tthink{<think>}The singer who released "Unapologetic" is Rihanna. The song "Love the Way You Lie" features guest vocals by Eminem, and this song appears on Eminem's album "The Marshall Mathers LP 2."\tthink{</think>} \newline \tanswer{<answer>}The Marshall Mathers LP 2\tanswer{</answer>}
\end{minipage} \\
\bottomrule
\end{tabular}
\caption{Search-R1 locks onto the answer to the base question (an artist name) and cannot redirect toward a different entity type (an album title) even after two explicit injections. \textsc{SteerSearch} follows the full chain and produces the correct album.}
\label{tab:case1}
\end{table*}

% Example 2
\begin{table*}[htbp]
\small
\setlength{\tabcolsep}{6pt}
\begin{tabular}{p{0.48\textwidth} p{0.48\textwidth}}
\toprule
\textbf{Search-R1} \textcolor{red}{\ding{55}} & \textbf{\textsc{SteerSearch}} \textcolor{ForestGreen}{\ding{51}} \\
\midrule
\multicolumn{2}{p{\textwidth}}{
\textbf{Question:} Conrad Anker located the body of which mountaineer?
\newline \textbf{Injection 1:} Which mountaineer is described as having taken part in the first three British expeditions to a major Himalayan peak?
\newline \textbf{Injection 2:} What were the first three British expeditions that this mountaineer participated in, and what mountain were they attempting?
} \\
\midrule
\multicolumn{2}{p{\textwidth}}{
\textbf{Ground truth:} Mount Everest
} \\
\midrule
\begin{minipage}[t]{0.48\textwidth}\raggedright
\tthink{<think>}I need to identify the mountaineer whose body was located by Conrad Anker.\tthink{</think>} \newline \tsearch{<search>}Conrad Anker located the body of which mountaineer\tsearch{</search>} \newline \tinfo{<information>}Doc 1(Title: 2007 Altitude Everest expedition) 2007 Altitude Everest expedition The 2007 Altitude Everest expedition, led by the American climber Conrad Anker, arrived at Base Camp below the north face of Everest in May 2007 and retraced the last journey of British climber George Mallory....\tinfo{</information>} \newline \tuser{[user] }Which mountaineer is described as having taken part in the first three British expeditions to a major Himalayan peak? \newline \tthink{<think>}I need to identify the mountaineer who was part of the first three British expeditions to a major Himalayan peak.\tthink{</think>} \newline \tsearch{<search>}first three British expeditions to a major Himalayan peak mountaineer\tsearch{</search>} \newline \tinfo{<information>}Doc 1(Title: Timeline of Mount Everest expeditions) Timeline of Mount Everest expeditions Mount Everest is the world's highest mountain, with a peak at 8,848 metres (29,029 ft) above sea level. It is situated in the Himalayan range....\tinfo{</information>} \newline \tuser{[user] }What were the first three British expeditions that this mountaineer participated in, and what mountain were they attempting? \newline \tthink{<think>}The search results show the first British expedition to Mount Everest included George Mallory. I need to answer which mountaineer’s body Conrad Anker located.\tthink{</think>} \newline \tanswer{<answer>}George Mallory\tanswer{</answer>}
\end{minipage}
 & 
\begin{minipage}[t]{0.48\textwidth}\raggedright
\tthink{<think>}I need to identify the specific mountaineer whose body was located by Conrad Anker.\tthink{</think>} \newline \tsearch{<search>}Conrad Anker located the body of which mountaineer\tsearch{</search>} \newline \tinfo{<information>}Doc 1(Title: 2007 Altitude Everest expedition) 2007 Altitude Everest expedition The 2007 Altitude Everest expedition, led by the American climber Conrad Anker, arrived at Base Camp below the north face of Everest in May 2007 and retraced the last journey of British climber George Mallory....\tinfo{</information>} \newline \tuser{[user] }Which mountaineer is described as having taken part in the first three British expeditions to a major Himalayan peak? \newline \tthink{<think>}I need to identify the mountaineer who participated in the first three British expeditions to a major Himalayan peak.\tthink{</think>} \newline \tsearch{<search>}first three British expeditions to a major Himalayan peak mountaineer\tsearch{</search>} \newline \tinfo{<information>}Doc 1(Title: Timeline of Mount Everest expeditions) Timeline of Mount Everest expeditions Mount Everest is the world's highest mountain, with a peak at 8,848 metres (29,029 ft) above sea level. It is situated in the Himalayan range....\tinfo{</information>} \newline \tthink{<think>}I need to identify the mountaineer associated with the first three British expeditions to a major Himalayan peak, which are the 1913–1914, 1921, and 1924 British Mount Everest Expedition (involving Ma\tthink{</think>} \newline \tsearch{<search>}first three British expeditions to a major Himalayan peak mountaineer George Mallory\tsearch{</search>} \newline \tinfo{<information>}Doc 1(Title: George Mallory) George Mallory George Herbert Leigh Mallory (18 June 1886 – 8 or 9 June 1924) was an English mountaineer who took part in the first three British expeditions to Mount Everest, in the early 1920s....\tinfo{</information>} \newline \tuser{[user] }What were the first three British expeditions that this mountaineer participated in, and what mountain were they attempting? \newline \tthink{<think>}I need to identify the first three British expeditions for Mount Everest and confirm that the mountaineer was George Mallory.\tthink{</think>} \newline \tsearch{<search>}first three British expeditions to Mount Everest George Mallory\tsearch{</search>} \newline \tinfo{<information>}Doc 1(Title: George Mallory) George Mallory George Herbert Leigh Mallory (18 June 1886 – 8 or 9 June 1924) was an English mountaineer who took part in the first three British expeditions to Mount Everest, in the early 1920s....\tinfo{</information>} \newline \tthink{<think>}The search results state that the mountaineer was George Mallory, who took part in the first three British expeditions to Mount Everest.\tthink{</think>} \newline \tanswer{<answer>}Mount Everest\tanswer{</answer>}
\end{minipage} \\
\bottomrule
\end{tabular}
\caption{Search-R1 correctly retrieves an intermediate fact (the mountaineer) and answers with it, ignoring two injections that redirect toward a different entity (the mountain). \textsc{SteerSearch} follows the injections to the final answer.}
\label{tab:case2}
\end{table*}

% Example 3
\begin{table*}[htbp]
\small
\setlength{\tabcolsep}{6pt}
\begin{tabular}{p{0.48\textwidth} p{0.48\textwidth}}
\toprule
\textbf{Search-R1} \textcolor{red}{\ding{55}} & \textbf{\textsc{SteerSearch}} \textcolor{ForestGreen}{\ding{51}} \\
\midrule
\multicolumn{2}{p{\textwidth}}{
\textbf{Question:} My Chick Bad official remix track listing and credits featured artists
\newline \textbf{Injection 1:} Brittany Nicole Carpentero My Chick Bad official remix how is she credited
\newline \textbf{Injection 2:} My Chick Bad official remix female guest verses which artists are featured
\newline \textbf{Injection 3:} Official remix to My Chick Bad featuring Ludacris and Nicki Minaj which other female rappers are on the remix
} \\
\midrule
\multicolumn{2}{p{\textwidth}}{
\textbf{Ground truth:} Eve, and Trina
} \\
\midrule
\begin{minipage}[t]{0.48\textwidth}\raggedright
\tthink{<think>}I need to identify the official remix track listing and featured artists for “My Chick Bad”.\tthink{</think>} \newline \tsearch{<search>}My Chick Bad official remix track listing and credits featured artists\tsearch{</search>} \newline \tinfo{<information>}Doc 1(Title: My Chick Bad) My Chick Bad "My Chick Bad" is a song by American rapper Ludacris. The song features Trinidadian recording artist Nicki Minaj, who co-wrote the song with Ludacris, Samuel Lindley, and Derrelle Davidson. The track was produced by The Legendary Traxster....\tinfo{</information>} \newline \tuser{[user] }Brittany Nicole Carpentero My Chick Bad official remix how is she credited \newline \tthink{<think>}I need to confirm Brittany Nicole Carpentero’s credit for the official remix of “My Chick Bad”.\tthink{</think>} \newline \tsearch{<search>}Brittany Nicole Carpentero credited on My Chick Bad official remix\tsearch{</search>} \newline \tinfo{<information>}Doc 1(Title: My Chick Bad) My Chick Bad "My Chick Bad" is a song by American rapper Ludacris. The song features Trinidadian recording artist Nicki Minaj, who co-wrote the song with Ludacris, Samuel Lindley, and Derrelle Davidson. The track was produced by The Legendary Traxster....\tinfo{</information>} \newline \tuser{[user] }My Chick Bad official remix female guest verses which artists are featured \newline \tuser{[user] }Official remix to My Chick Bad featuring Ludacris and Nicki Minaj which other female rappers are on the remix \newline \tthink{<think>}The search results show the official remix features female rappers Diamond, Eve, and Trina.\tthink{</think>} \newline \tanswer{<answer>}Diamond, Eve, Trina\tanswer{</answer>}
\end{minipage}
 & 
\begin{minipage}[t]{0.48\textwidth}\raggedright
\tthink{<think>}I need to identify the specific remix of "My Chick Bad" and then find its official track listing and featured artists from reliable music or remix-related sources.\tthink{</think>} \newline \tsearch{<search>}My Chick Bad official remix track listing and credits featured artists\tsearch{</search>} \newline \tinfo{<information>}Doc 1(Title: My Chick Bad) My Chick Bad "My Chick Bad" is a song by American rapper Ludacris. The song features Trinidadian recording artist Nicki Minaj, who co-wrote the song with Ludacris, Samuel Lindley, and Derrelle Davidson. The track was produced by The Legendary Traxster....\tinfo{</information>} \newline \tuser{[user] }Brittany Nicole Carpentero My Chick Bad official remix how is she credited \newline \tthink{<think>}I need to identify who Brittany Nicole Carpentero is in the context of the "My Chick Bad" official remix and determine her specific credit (e.g., featured artist, producer, etc.).\tthink{</think>} \newline \tsearch{<search>}Brittany Nicole Carpentero My Chick Bad official remix credits\tsearch{</search>} \newline \tinfo{<information>}Doc 1(Title: My Chick Bad) Krueger character, complete with the trademark glove made of razor blades. She is found lying in a couch, held by restraints, before the video ends with Ludacris and Nicki performing in the lair....\tinfo{</information>} \newline \tuser{[user] }My Chick Bad official remix female guest verses which artists are featured \newline \tthink{<think>}I need to identify the official remix of "My Chick Bad" and then list the featured artists from reliable sources.\tthink{</think>} \newline \tsearch{<search>}My Chick Bad official remix female guest verses featured artists Eve Diamond Trina\tsearch{</search>} \newline \tinfo{<information>}Doc 1(Title: My Chick Bad) the interview, he said, "That’s one of the records that we gonna put out there." The official remix features Eve, Trina, and Diamond, while Ludacris only has background vocals....\tinfo{</information>} \newline \tuser{[user] }Official remix to My Chick Bad featuring Ludacris and Nicki Minaj which other female rappers are on the remix \newline \tthink{<think>}The search results state that the official remix to "My Chick Bad" features the female rappers Diamond, Eve, and Trina.\tthink{</think>} \newline \tanswer{<answer>}Eve and Trina\tanswer{</answer>}
\end{minipage} \\
\bottomrule
\end{tabular}
\caption{Search-R1 retrieves a list of featured artists and cannot refine it across three injections that narrow the question to a specific subset. \textsc{SteerSearch} progressively narrows to the correct answer.}
\label{tab:case3}
\end{table*}

% Example 4
\begin{table*}[htbp]
\small
\setlength{\tabcolsep}{6pt}
\begin{tabular}{p{0.48\textwidth} p{0.48\textwidth}}
\toprule
\textbf{Search-R1} \textcolor{red}{\ding{55}} & \textbf{\textsc{SteerSearch}} \textcolor{ForestGreen}{\ding{51}} \\
\midrule
\multicolumn{2}{p{\textwidth}}{
\textbf{Question:} "Live at Radio City" live album
\newline \textbf{Injection 1:} Discogs "Live at Radio City" album
\newline \textbf{Injection 2:} AllMusic "Live at Radio City" live album details
\newline \textbf{Injection 3:} Where was the album "Live at Radio City" recorded (city)?
} \\
\midrule
\multicolumn{2}{p{\textwidth}}{
\textbf{Ground truth:} New York City
} \\
\midrule
\begin{minipage}[t]{0.48\textwidth}\raggedright
\tthink{<think>}I need to identify the title of the live album “Live at Radio City.\tthink{</think>} \newline \tsearch{<search>}Live at Radio City live album\tsearch{</search>} \newline \tinfo{<information>}Doc 1(Title: Live at Radio City) various bonus features, such as a behind-the-scenes documentary, a 5.1 audio mix, and a photo gallery. The DVD was directed by Sam Erickson of 44 Pictures and Fenton Williams of Filament Productions. The album debuted at number three on the U.S....\tinfo{</information>} \newline \tuser{[user] }Discogs "Live at Radio City" album \newline \tuser{[user] }AllMusic "Live at Radio City" live album details \newline \tuser{[user] }Where was the album "Live at Radio City" recorded (city)? \newline \tthink{<think>}The search results show “Live at Radio City” is a live album recorded at Radio City Music Hall.\tthink{</think>} \newline \tanswer{<answer>}Radio City Music Hall\tanswer{</answer>}
\end{minipage}
 & 
\begin{minipage}[t]{0.48\textwidth}\raggedright
\tthink{<think>}I need to identify the specific live album titled "Live at Radio City" and confirm its artist and release details from reliable music sources.\tthink{</think>} \newline \tsearch{<search>}Live at Radio City live album\tsearch{</search>} \newline \tinfo{<information>}Doc 1(Title: Live at Radio City) various bonus features, such as a behind-the-scenes documentary, a 5.1 audio mix, and a photo gallery. The DVD was directed by Sam Erickson of 44 Pictures and Fenton Williams of Filament Productions. The album debuted at number three on the U.S....\tinfo{</information>} \newline \tuser{[user] }Discogs "Live at Radio City" album \newline \tuser{[user] }AllMusic "Live at Radio City" live album details \newline \tuser{[user] }Where was the album "Live at Radio City" recorded (city)? \newline \tthink{<think>}I need to identify the specific city where the album "Live at Radio City" was recorded, which is described as a live acoustic rock concert at Radio City Music Hall in New York City.\tthink{</think>} \newline \tsearch{<search>}Live at Radio City recorded at Radio City Music Hall in New York City\tsearch{</search>} \newline \tinfo{<information>}Doc 1(Title: Live at Radio City) various bonus features, such as a behind-the-scenes documentary, a 5.1 audio mix, and a photo gallery. The DVD was directed by Sam Erickson of 44 Pictures and Fenton Williams of Filament Productions. The album debuted at number three on the U.S....\tinfo{</information>} \newline \tthink{<think>}The search results state that the album "Live at Radio City" was recorded at a live acoustic rock concert at Radio City Music Hall in New York City.\tthink{</think>} \newline \tanswer{<answer>}New York City\tanswer{</answer>}
\end{minipage} \\
\bottomrule
\end{tabular}
\caption{Search-R1 exhausts its search budget without answering, receives all three injections stacked, and still answers with the venue name (Radio City Music Hall) even though the final injection explicitly asks for the recording city. \textsc{SteerSearch} correctly extracts the city (New York City).}
\label{tab:case4}
\end{table*}

\subsection{Both Methods Succeed}

Tables~\ref{tab:case5}--\ref{tab:case10} show trajectories where both Search-R1 and \textsc{SteerSearch} produce the correct final answer. These cases illustrate that when the injection aligns with evidence the model has already retrieved, even fixed-goal search training can follow the new goal. The contrast with the previous subsection highlights that Search-R1's failure is not universal but depends on the degree of context shift required.

% Example 5
\begin{table*}[htbp]
\small
\setlength{\tabcolsep}{6pt}
\begin{tabular}{p{0.48\textwidth} p{0.48\textwidth}}
\toprule
\textbf{Search-R1} \textcolor{ForestGreen}{\ding{51}} & \textbf{\textsc{SteerSearch}} \textcolor{ForestGreen}{\ding{51}} \\
\midrule
\multicolumn{2}{p{\textwidth}}{
\textbf{Question:} Wigan Athletic F.C. 2017–18 season overview
\newline \textbf{Injection 1:} English Football League Cup sponsorship name in the 2017–18 season
} \\
\midrule
\multicolumn{2}{p{\textwidth}}{
\textbf{Ground truth:} Carabao Cup
} \\
\midrule
\begin{minipage}[t]{0.48\textwidth}\raggedright
\tthink{<think>}I need to describe the 2017–18 season overview for Wigan Athletic F.C.\tthink{</think>} \newline \tsearch{<search>}Wigan Athletic F.C. 2017–18 season overview\tsearch{</search>} \newline \tinfo{<information>}Doc 1(Title: 2017–18 Wigan Athletic F.C. season) 2017–18 Wigan Athletic F.C. season The 2017–18 season is Wigan Athletic's 86th year in existence and their first back in League One, after being relegated the previous season....\tinfo{</information>} \newline \tuser{[user] }English Football League Cup sponsorship name in the 2017–18 season \newline \tthink{<think>}I need to identify the English Football League Cup sponsorship name for the 2017–18 season.\tthink{</think>} \newline \tsearch{<search>}2017–18 English Football League Cup sponsorship name\tsearch{</search>} \newline \tinfo{<information>}Doc 1(Title: 2017–18 EFL Cup) 2017–18 EFL Cup The 2017–18 EFL Cup was the 58th season of the EFL Cup. The competition was open to all 92 clubs participating in the Premier League and the English Football League....\tinfo{</information>} \newline \tthink{<think>}The search results show the 2017–18 EFL Cup was known as the Carabao Cup.\tthink{</think>} \newline \tanswer{<answer>}Carabao Cup\tanswer{</answer>}
\end{minipage}
 & 
\begin{minipage}[t]{0.48\textwidth}\raggedright
\tthink{<think>}I need to identify the comprehensive overview or summary of the 2017–18 season for Wigan Athletic Football Club, including key statistics, results, and any notable events or performances.\tthink{</think>} \newline \tsearch{<search>}Wigan Athletic F.C. 2017–18 season overview\tsearch{</search>} \newline \tinfo{<information>}Doc 1(Title: 2017–18 Wigan Athletic F.C. season) 2017–18 Wigan Athletic F.C. season The 2017–18 season is Wigan Athletic's 86th year in existence and their first back in League One, after being relegated the previous season....\tinfo{</information>} \newline \tuser{[user] }English Football League Cup sponsorship name in the 2017–18 season \newline \tthink{<think>}I need to identify the sponsorship name of the English Football League Cup for the 2017–18 season.\tthink{</think>} \newline \tsearch{<search>}2017–18 English Football League Cup sponsorship name\tsearch{</search>} \newline \tinfo{<information>}Doc 1(Title: 2017–18 EFL Cup) 2017–18 EFL Cup The 2017–18 EFL Cup was the 58th season of the EFL Cup. The competition was open to all 92 clubs participating in the Premier League and the English Football League....\tinfo{</information>} \newline \tthink{<think>}The search results state that the 2017–18 EFL Cup was known as the Carabao Cup due to a sponsorship deal with Carabao Energy Drinks.\tthink{</think>} \newline \tanswer{<answer>}Carabao Cup\tanswer{</answer>}
\end{minipage} \\
\bottomrule
\end{tabular}
\caption{Both models correctly identify the sponsored name of the competition after the injection. Search-R1 retrieves the right document and produces the correct answer; \textsc{SteerSearch} pivots its search and also retrieves the correct answer.}
\label{tab:case5}
\end{table*}

% Example 6
\begin{table*}[htbp]
\small
\setlength{\tabcolsep}{6pt}
\begin{tabular}{p{0.48\textwidth} p{0.48\textwidth}}
\toprule
\textbf{Search-R1} \textcolor{ForestGreen}{\ding{51}} & \textbf{\textsc{SteerSearch}} \textcolor{ForestGreen}{\ding{51}} \\
\midrule
\multicolumn{2}{p{\textwidth}}{
\textbf{Question:} Bordan Tkachuk CEO company
\newline \textbf{Injection 1:} What products and services does that company provide?
} \\
\midrule
\multicolumn{2}{p{\textwidth}}{
\textbf{Ground truth:} IT products and services
} \\
\midrule
\begin{minipage}[t]{0.48\textwidth}\raggedright
\tthink{<think>}I need to identify the company that Bordan Tkachuk is the CEO of.\tthink{</think>} \newline \tsearch{<search>}Bordan Tkachuk CEO company\tsearch{</search>} \newline \tinfo{<information>}Doc 1(Title: Bordan Tkachuk) Bordan Tkachuk Bordan Tkachuk ( ) is a British business executive, the former CEO of Viglen, also known from his appearances on the BBC-produced British version of "The Apprentice," interviewing for his boss Lord Sugar....\tinfo{</information>} \newline \tuser{[user] }What products and services does that company provide? \newline \tthink{<think>}The search results show Bordan Tkachuk is the former CEO of Viglen. I need to identify what Viglen provides.\tthink{</think>} \newline \tsearch{<search>}Viglen company provides what products and services\tsearch{</search>} \newline \tinfo{<information>}Doc 1(Title: Viglen) Viglen Viglen Ltd provides IT products and services, including storage systems, servers, workstations and data/voice communications equipment and services. The British company was formed in 1975 by Vigen Boyadjian and was acquired by Alan Sugar (later Lord Sugar)'s company Amstr...\tinfo{</information>} \newline \tthink{<think>}The search results show Bordan Tkachuk is the former CEO of Viglen, and Viglen provides IT products and services.\tthink{</think>} \newline \tanswer{<answer>}IT products and services\tanswer{</answer>}
\end{minipage}
 & 
\begin{minipage}[t]{0.48\textwidth}\raggedright
\tthink{<think>}I need to identify Bordan Tkachuk and determine the company he currently serves as CEO.\tthink{</think>} \newline \tsearch{<search>}Bordan Tkachuk CEO company\tsearch{</search>} \newline \tinfo{<information>}Doc 1(Title: Bordan Tkachuk) Bordan Tkachuk Bordan Tkachuk ( ) is a British business executive, the former CEO of Viglen, also known from his appearances on the BBC-produced British version of "The Apprentice," interviewing for his boss Lord Sugar....\tinfo{</information>} \newline \tuser{[user] }What products and services does that company provide? \newline \tthink{<think>}I need to identify the company Bordan Tkachuk was CEO of (Viglen) and then find what products and services Viglen provides.\tthink{</think>} \newline \tsearch{<search>}Viglen company products and services\tsearch{</search>} \newline \tinfo{<information>}Doc 1(Title: Viglen) Viglen Viglen Ltd provides IT products and services, including storage systems, servers, workstations and data/voice communications equipment and services. The British company was formed in 1975 by Vigen Boyadjian and was acquired by Alan Sugar (later Lord Sugar)'s company Amstr...\tinfo{</information>} \newline \tthink{<think>}The search results state that Bordan Tkachuk was the former CEO of Viglen, and that Viglen provides IT products and services, including storage systems, servers, workstations, and data/voice communica\tthink{</think>} \newline \tanswer{<answer>}IT products and services\tanswer{</answer>}
\end{minipage} \\
\bottomrule
\end{tabular}
\caption{Both models correctly follow the injection and extract the products and services provided by the company. The retrieved document contains the answer verbatim, and both models identify and surface it.}
\label{tab:case6}
\end{table*}

% Example 7
\begin{table*}[htbp]
\small
\setlength{\tabcolsep}{6pt}
\begin{tabular}{p{0.48\textwidth} p{0.48\textwidth}}
\toprule
\textbf{Search-R1} \textcolor{ForestGreen}{\ding{51}} & \textbf{\textsc{SteerSearch}} \textcolor{ForestGreen}{\ding{51}} \\
\midrule
\multicolumn{2}{p{\textwidth}}{
\textbf{Question:} Lead single from an album titled "Confessions" song title and artist list
\newline \textbf{Injection 1:} Which song was number one for eight consecutive weeks and kept a popular song stuck at number two behind it for eight weeks
\newline \textbf{Injection 2:} Who is the singer of the song that was the lead single from the album "Confessions" and spent eight consecutive weeks at number one?
} \\
\midrule
\multicolumn{2}{p{\textwidth}}{
\textbf{Ground truth:} Usher
} \\
\midrule
\begin{minipage}[t]{0.48\textwidth}\raggedright
\tthink{<think>}I need to identify the lead single from the album “Confessions” and list its title and artist.\tthink{</think>} \newline \tsearch{<search>}lead single from album "Confessions" song title and artist\tsearch{</search>} \newline \tinfo{<information>}Doc 1(Title: Confessions Part II) Confessions Part II "Confessions Part II" is a song by R\&B singer Usher, produced by Jermaine Dupri and Bryan-Michael Cox for Usher's fourth album "Confessions"....\tinfo{</information>} \newline \tthink{<think>}The search results show “Confessions” is Usher’s fourth album. I need the lead single from this album.\tthink{</think>} \newline \tsearch{<search>}lead single from Usher album "Confessions"\tsearch{</search>} \newline \tinfo{<information>}Doc 1(Title: Confessions Part II) Confessions Part II "Confessions Part II" is a song by R\&B singer Usher, produced by Jermaine Dupri and Bryan-Michael Cox for Usher's fourth album "Confessions"....\tinfo{</information>} \newline \tuser{[user] }Which song was number one for eight consecutive weeks and kept a popular song stuck at number two behind it for eight weeks \newline \tthink{<think>}The search results show “Confessions” is Usher’s fourth album. I need the lead single from this album and its title and artist.\tthink{</think>} \newline \tsearch{<search>}lead single from Usher album "Confessions" title and artist\tsearch{</search>} \newline \tinfo{<information>}Doc 1(Title: Confessions Part II) Confessions Part II "Confessions Part II" is a song by R\&B singer Usher, produced by Jermaine Dupri and Bryan-Michael Cox for Usher's fourth album "Confessions"....\tinfo{</information>} \newline \tuser{[user] }Who is the singer of the song that was the lead single from the album "Confessions" and spent eight consecutive weeks at number one? \newline \tthink{<think>}The search results show the lead single from Usher’s album “Confessions” is “Burn.” I need to answer the question about the singer of that song.\tthink{</think>} \newline \tanswer{<answer>}Usher\tanswer{</answer>}
\end{minipage}
 & 
\begin{minipage}[t]{0.48\textwidth}\raggedright
\tthink{<think>}I need to identify the album titled "Confessions" and find its lead single, then list the song title and the artist(s) associated with that single.\tthink{</think>} \newline \tsearch{<search>}album "Confessions" lead single song title artist\tsearch{</search>} \newline \tinfo{<information>}Doc 1(Title: Confessions Part II) Note Award for co-writing the song, alongside his many achievements and contributions to American popular music. British record company EMI was recognized as Publisher of the Year for the song, including "Burn" and songs by Alicia Keys, Beyoncé Knowles,...\tinfo{</information>} \newline \tthink{<think>}I need to identify the lead single from the album "Confessions" and then find the song title and the artist(s) associated with that single.\tthink{</think>} \newline \tsearch{<search>}lead single from the album "Confessions" Usher song title\tsearch{</search>} \newline \tinfo{<information>}Doc 1(Title: Confessions Part II) Confessions Part II "Confessions Part II" is a song by R\&B singer Usher, produced by Jermaine Dupri and Bryan-Michael Cox for Usher's fourth album "Confessions"....\tinfo{</information>} \newline \tuser{[user] }Which song was number one for eight consecutive weeks and kept a popular song stuck at number two behind it for eight weeks \newline \tthink{<think>}I need to identify the song that held the number-one position for eight consecutive weeks and was behind another popular song at number two for eight weeks.\tthink{</think>} \newline \tsearch{<search>}number-one for eight consecutive weeks kept a popular song stuck at number two behind it\tsearch{</search>} \newline \tinfo{<information>}Doc 1(Title: You're Still the One) most successful single on the "Billboard" Hot 100. It debuted at number 51 on February 14, 1998. It spent 42 weeks on the chart and peaked at number two for nine non-consecutive weeks starting May 2, 1998, making it one of the longest runs at No....\tinfo{</information>} \newline \tuser{[user] }Who is the singer of the song that was the lead single from the album "Confessions" and spent eight consecutive weeks at number one? \newline \tthink{<think>}I need to identify the lead single from the album "Confessions" and then determine the singer of that single.\tthink{</think>} \newline \tsearch{<search>}lead single from the album "Confessions" Usher Confessions lead single\tsearch{</search>} \newline \tinfo{<information>}Doc 1(Title: Confessions Part II) Confessions Part II "Confessions Part II" is a song by R\&B singer Usher, produced by Jermaine Dupri and Bryan-Michael Cox for Usher's fourth album "Confessions"....\tinfo{</information>} \newline \tthink{<think>}The lead single from the album "Confessions" is "Confessions Part II" by Usher.\tthink{</think>} \newline \tanswer{<answer>}Usher\tanswer{</answer>}
\end{minipage} \\
\bottomrule
\end{tabular}
\caption{Both models correctly identify the artist across two injections that progressively narrow from the song title toward the singer. The injection chain provides sufficient guidance for both methods to pivot and produce the correct answer.}
\label{tab:case7}
\end{table*}

% Example 8
\begin{table*}[htbp]
\small
\setlength{\tabcolsep}{6pt}
\begin{tabular}{p{0.48\textwidth} p{0.48\textwidth}}
\toprule
\textbf{Search-R1} \textcolor{ForestGreen}{\ding{51}} & \textbf{\textsc{SteerSearch}} \textcolor{ForestGreen}{\ding{51}} \\
\midrule
\multicolumn{2}{p{\textwidth}}{
\textbf{Question:} 1974 NCAA Division I college football teams list Raiders
\newline \textbf{Injection 1:} 1974 Texas Tech Raiders football team season summary represented university
\newline \textbf{Injection 2:} Texas Tech Raiders football team represented what university in Lubbock Texas
} \\
\midrule
\multicolumn{2}{p{\textwidth}}{
\textbf{Ground truth:} Texas Tech University
} \\
\midrule
\begin{minipage}[t]{0.48\textwidth}\raggedright
\tthink{<think>}I need to identify the 1974 NCAA Division I college football teams listed as “Raiders”.\tthink{</think>} \newline \tsearch{<search>}1974 NCAA Division I college football teams list Raiders\tsearch{</search>} \newline \tinfo{<information>}Doc 1(Title: 1974 Oakland Raiders season) 1974 Oakland Raiders season The 1974 Oakland Raiders season was the team's 15th season in Oakland and fifth in the National Football League. The team would post a superb 12–2 record; the campaign's two losses would be by a total of four points....\tinfo{</information>} \newline \tuser{[user] }1974 Texas Tech Raiders football team season summary represented university \newline \tthink{<think>}The search results show the 1974 NCAA Division I football season list. I need to identify the team listed as “Raiders”.\tthink{</think>} \newline \tsearch{<search>}1974 Texas Tech Raiders football team\tsearch{</search>} \newline \tinfo{<information>}Doc 1(Title: 1974 Texas Tech Red Raiders football team) 1974 Texas Tech Red Raiders football team The 1974 Texas Tech Red Raiders football team represented Texas Tech University in the Southwest Conference during the 1974 NCAA Division I football season....\tinfo{</information>} \newline \tuser{[user] }Texas Tech Raiders football team represented what university in Lubbock Texas \newline \tthink{<think>}The search results show the 1974 Texas Tech Red Raiders football team represented Texas Tech University.\tthink{</think>} \newline \tanswer{<answer>}Texas Tech University\tanswer{</answer>}
\end{minipage}
 & 
\begin{minipage}[t]{0.48\textwidth}\raggedright
\tthink{<think>}I need to identify the 1974 NCAA Division I college football teams that are referred to as the "Raiders," likely referring to a specific university team such as the Oakland Raiders or another college \tthink{</think>} \newline \tsearch{<search>}1974 NCAA Division I college football teams list Raiders\tsearch{</search>} \newline \tinfo{<information>}Doc 1(Title: 1974 Oakland Raiders season) 1974 Oakland Raiders season The 1974 Oakland Raiders season was the team's 15th season in Oakland and fifth in the National Football League. The team would post a superb 12–2 record; the campaign's two losses would be by a total of four points....\tinfo{</information>} \newline \tthink{<think>}The search results show the 1974 NCAA Division I football season lists teams such as Oklahoma, Notre Dame, Alabama, Ohio State, and USC. I need to identify which of these are referred to as "Raiders" \tthink{</think>} \newline \tsearch{<search>}1974 NCAA Division I football teams referred to as "Raiders"\tsearch{</search>} \newline \tinfo{<information>}Doc 1(Title: 1974 Oakland Raiders season) 1974 Oakland Raiders season The 1974 Oakland Raiders season was the team's 15th season in Oakland and fifth in the National Football League. The team would post a superb 12–2 record; the campaign's two losses would be by a total of four points....\tinfo{</information>} \newline \tuser{[user] }1974 Texas Tech Raiders football team season summary represented university \newline \tuser{[user] }Texas Tech Raiders football team represented what university in Lubbock Texas \newline \tthink{<think>}The search results state that the 1974 Texas Tech Red Raiders football team represented Texas Tech University.\tthink{</think>} \newline \tanswer{<answer>}Texas Tech University\tanswer{</answer>}
\end{minipage} \\
\bottomrule
\end{tabular}
\caption{Both models resolve the disambiguation between a college team and a professional franchise after injections explicitly name the right team and university. Both produce the correct answer despite the ambiguity in the base question.}
\label{tab:case8}
\end{table*}

% Example 9
\begin{table*}[htbp]
\small
\setlength{\tabcolsep}{6pt}
\begin{tabular}{p{0.48\textwidth} p{0.48\textwidth}}
\toprule
\textbf{Search-R1} \textcolor{ForestGreen}{\ding{51}} & \textbf{\textsc{SteerSearch}} \textcolor{ForestGreen}{\ding{51}} \\
\midrule
\multicolumn{2}{p{\textwidth}}{
\textbf{Question:} Buck-Tick Japanese rock band profile biography overview
\newline \textbf{Injection 1:} Buck-Tick official website band biography origin information
\newline \textbf{Injection 2:} Major music databases (AllMusic, Discogs, MusicBrainz) Buck-Tick artist page
\newline \textbf{Injection 3:} Where does Buck-Tick hail from?
} \\
\midrule
\multicolumn{2}{p{\textwidth}}{
\textbf{Ground truth:} Fujioka, Gunma
} \\
\midrule
\begin{minipage}[t]{0.48\textwidth}\raggedright
\tthink{<think>}I need to provide a profile biography overview of the Japanese rock band Buck-Tick.\tthink{</think>} \newline \tsearch{<search>}Buck-Tick Japanese rock band profile biography\tsearch{</search>} \newline \tinfo{<information>}Doc 1(Title: Buck-Tick) Buck-Tick Buck-Tick (stylized as BUCK-TICK) is a Japanese rock band, formed in Fujioka, Gunma in 1983. The group has consisted of lead vocalist Atsushi Sakurai, lead guitarist Hisashi Imai, rhythm guitarist Hidehiko Hoshino,...\tinfo{</information>} \newline \tuser{[user] }Buck-Tick official website band biography origin information \newline \tuser{[user] }Major music databases (AllMusic, Discogs, MusicBrainz) Buck-Tick artist page \newline \tuser{[user] }Where does Buck-Tick hail from? \newline \tthink{<think>}The search results show Buck-Tick are a Japanese rock band formed in Fujioka, Gunma.\tthink{</think>} \newline \tanswer{<answer>}Fujioka, Gunma\tanswer{</answer>}
\end{minipage}
 & 
\begin{minipage}[t]{0.48\textwidth}\raggedright
\tthink{<think>}I need to gather a comprehensive profile and biography overview of the Japanese rock band Buck-Tick, including their history, members, musical style, and notable achievements.\tthink{</think>} \newline \tsearch{<search>}Buck-Tick Japanese rock band profile biography overview\tsearch{</search>} \newline \tinfo{<information>}Doc 1(Title: Buck-Tick) Buck-Tick Buck-Tick (stylized as BUCK-TICK) is a Japanese rock band, formed in Fujioka, Gunma in 1983. The group has consisted of lead vocalist Atsushi Sakurai, lead guitarist Hisashi Imai, rhythm guitarist Hidehiko Hoshino,...\tinfo{</information>} \newline \tuser{[user] }Buck-Tick official website band biography origin information \newline \tthink{<think>}I need to find the official Buck-Tick website or the band’s official biography section that provides detailed information on their origin and formation.\tthink{</think>} \newline \tsearch{<search>}Buck-Tick official website band biography origin information\tsearch{</search>} \newline \tinfo{<information>}Doc 1(Title: Buck-Tick) Buck-Tick Buck-Tick (stylized as BUCK-TICK) is a Japanese rock band, formed in Fujioka, Gunma in 1983. The group has consisted of lead vocalist Atsushi Sakurai, lead guitarist Hisashi Imai, rhythm guitarist Hidehiko Hoshino,...\tinfo{</information>} \newline \tuser{[user] }Major music databases (AllMusic, Discogs, MusicBrainz) Buck-Tick artist page \newline \tthink{<think>}I need to identify the specific Buck-Tick artist pages on major music databases like AllMusic, Discogs, and MusicBrainz, which should provide a detailed biography and origin information.\tthink{</think>} \newline \tsearch{<search>}Buck-Tick AllMusic Discogs MusicBrainz artist page\tsearch{</search>} \newline \tinfo{<information>}Doc 1(Title: Buck-Tick) Buck-Tick Buck-Tick (stylized as BUCK-TICK) is a Japanese rock band, formed in Fujioka, Gunma in 1983. The group has consisted of lead vocalist Atsushi Sakurai, lead guitarist Hisashi Imai, rhythm guitarist Hidehiko Hoshino,...\tinfo{</information>} \newline \tuser{[user] }Where does Buck-Tick hail from? \newline \tthink{<think>}The search results state that Buck-Tick is a Japanese rock band formed in Fujioka, Gunma in 1983.\tthink{</think>} \newline \tanswer{<answer>}Fujioka, Gunma\tanswer{</answer>}
\end{minipage} \\
\bottomrule
\end{tabular}
\caption{Both models follow the final injection and produce the specific city rather than the country. Both retrieve the relevant information and correctly redirect their answer target from the coarse base-question framing to the precise injected goal.}
\label{tab:case9}
\end{table*}

% Example 10
\begin{table*}[htbp]
\small
\setlength{\tabcolsep}{6pt}
\begin{tabular}{p{0.48\textwidth} p{0.48\textwidth}}
\toprule
\textbf{Search-R1} \textcolor{ForestGreen}{\ding{51}} & \textbf{\textsc{SteerSearch}} \textcolor{ForestGreen}{\ding{51}} \\
\midrule
\multicolumn{2}{p{\textwidth}}{
\textbf{Question:} 2008 film screenplay by Andrew Davies adapted from an 1811 novel
\newline \textbf{Injection 1:} Which 1811 novel is the source material for the 2008 film with a screenplay by Andrew Davies?
\newline \textbf{Injection 2:} Who wrote the 1811 novel that the 2008 Andrew Davies–screenplay film is based on?
\newline \textbf{Injection 3:} How many major novels did the author of that 1811 novel write?
} \\
\midrule
\multicolumn{2}{p{\textwidth}}{
\textbf{Ground truth:} six
} \\
\midrule
\begin{minipage}[t]{0.48\textwidth}\raggedright
\tthink{<think>}I need to identify the 2008 film that was adapted from an 1811 novel by Andrew Davies.\tthink{</think>} \newline \tsearch{<search>}2008 film screenplay by Andrew Davies adapted from 1811 novel\tsearch{</search>} \newline \tinfo{<information>}Doc 1(Title: Sense and Sensibility (2008 TV series)) Phipps earned nominations for a BAFTA Television Craft Award for Best Original Television Music and a Creative Arts Emmy Award for Outstanding Music Composition for a Miniseries, Movie or Special....\tinfo{</information>} \newline \tuser{[user] }Which 1811 novel is the source material for the 2008 film with a screenplay by Andrew Davies? \newline \tuser{[user] }Who wrote the 1811 novel that the 2008 Andrew Davies–screenplay film is based on? \newline \tuser{[user] }How many major novels did the author of that 1811 novel write? \newline \tthink{<think>}The search results show the 2008 film is “Sense and Sensibility,” adapted from Jane Austen’s 1811 novel. I need to identify how many major novels Jane Austen wrote.\tthink{</think>} \newline \tsearch{<search>}Jane Austen wrote how many major novels\tsearch{</search>} \newline \tinfo{<information>}Doc 1(Title: Jane Austen) Jane Austen Jane Austen (; 16 December 1775 – 18 July 1817) was an English novelist known primarily for her six major novels, which interpret, critique and comment upon the British landed gentry at the end of the 18th century....\tinfo{</information>} \newline \tthink{<think>}The search results show the 2008 film is “Sense and Sensibility,” adapted from Jane Austen’s 1811 novel “Sense and Sensibility.” The results also state Jane Austen wrote six major novels.\tthink{</think>} \newline \tanswer{<answer>}six\tanswer{</answer>}
\end{minipage}
 & 
\begin{minipage}[t]{0.48\textwidth}\raggedright
\tthink{<think>}I need to identify the 2008 film that was adapted from an 1811 novel and had a screenplay written by Andrew Davies.\tthink{</think>} \newline \tsearch{<search>}2008 film adapted from an 1811 novel screenplay by Andrew Davies\tsearch{</search>} \newline \tinfo{<information>}Doc 1(Title: Sense and Sensibility (2008 TV series)) Phipps earned nominations for a BAFTA Television Craft Award for Best Original Television Music and a Creative Arts Emmy Award for Outstanding Music Composition for a Miniseries, Movie or Special....\tinfo{</information>} \newline \tuser{[user] }Which 1811 novel is the source material for the 2008 film with a screenplay by Andrew Davies? \newline \tthink{<think>}I need to identify the 2008 film adapted from an 1811 novel and then confirm that its screenplay was written by Andrew Davies.\tthink{</think>} \newline \tsearch{<search>}2008 film adapted from an 1811 novel screenplay by Andrew Davies\tsearch{</search>} \newline \tinfo{<information>}Doc 1(Title: Sense and Sensibility (2008 TV series)) Phipps earned nominations for a BAFTA Television Craft Award for Best Original Television Music and a Creative Arts Emmy Award for Outstanding Music Composition for a Miniseries, Movie or Special....\tinfo{</information>} \newline \tuser{[user] }Who wrote the 1811 novel that the 2008 Andrew Davies–screenplay film is based on? \newline \tthink{<think>}I need to identify the 1811 novel that the 2008 film adapted from, which is Jane Austen's "Sense and Sensibility," and then find the author of that novel.\tthink{</think>} \newline \tsearch{<search>}1811 novel Sense and Sensibility author Jane Austen\tsearch{</search>} \newline \tinfo{<information>}Doc 1(Title: Sense and Sensibility) Sense and Sensibility Sense and Sensibility is a novel by Jane Austen, published in 1811. It was published anonymously; "By A Lady" appears on the title page where the author's name might have been....\tinfo{</information>} \newline \tuser{[user] }How many major novels did the author of that 1811 novel write? \newline \tthink{<think>}I need to identify the author of "Sense and Sensibility" (Jane Austen) and then determine how many major novels she wrote.\tthink{</think>} \newline \tsearch{<search>}how many major novels did jane austen write\tsearch{</search>} \newline \tinfo{<information>}Doc 1(Title: Jane Austen) Jane Austen Jane Austen (; 16 December 1775 – 18 July 1817) was an English novelist known primarily for her six major novels, which interpret, critique and comment upon the British landed gentry at the end of the 18th century....\tinfo{</information>} \newline \tthink{<think>}The 2008 film is adapted from Jane Austen's 1811 novel "Sense and Sensibility," and the author of that novel is Jane Austen, who wrote six major novels.\tthink{</think>} \newline \tanswer{<answer>}six\tanswer{</answer>}
\end{minipage} \\
\bottomrule
\end{tabular}
\caption{Both models follow all three hops in the chain, starting from the novel title, redirecting toward the author, and finally producing the correct count of novels. The progressive injection structure guides both methods to the correct answer.}
\label{tab:case10}
\end{table*}

\subsection{Both Methods Fail}

Tables~\ref{tab:case11}--\ref{tab:case15} show 5 trajectories where neither method produces the correct answer. These involve multi-hop chains that require precise role disambiguation, compound answers, or abstract conceptual reasoning that the retrieved passages do not surface directly.

% Example 11
\begin{table*}[htbp]
\small
\setlength{\tabcolsep}{6pt}
\begin{tabular}{p{0.48\textwidth} p{0.48\textwidth}}
\toprule
\textbf{Search-R1} \textcolor{red}{\ding{55}} & \textbf{\textsc{SteerSearch}} \textcolor{red}{\ding{55}} \\
\midrule
\multicolumn{2}{p{\textwidth}}{
\textbf{Question:} Teide National Park location country and island
\newline \textbf{Injection 1:} Garajonay National Park location country and island
\newline \textbf{Injection 2:} Spain national parks in Atlantic islands autonomous community
} \\
\midrule
\multicolumn{2}{p{\textwidth}}{
\textbf{Ground truth:} Canary Islands, Spain
} \\
\midrule
\begin{minipage}[t]{0.48\textwidth}\raggedright
\tthink{<think>}I need to identify the country and island where Teide National Park is located.\tthink{</think>} \newline \tsearch{<search>}Teide National Park location country and island\tsearch{</search>} \newline \tinfo{<information>}Doc 1(Title: Teide National Park) Teide National Park Teide National Park (, ) is a national park located in Tenerife (Canary Islands, Spain). The national park is centered on Mount Teide, the highest mountain of Spain (3,718 meters high) in addition,...\tinfo{</information>} \newline \tuser{[user] }Garajonay National Park location country and island \newline \tuser{[user] }Spain national parks in Atlantic islands autonomous community \newline \tthink{<think>}The search results show Teide National Park is located in Tenerife, Canary Islands, Spain.\tthink{</think>} \newline \tanswer{<answer>}Spain\tanswer{</answer>}
\end{minipage}
 & 
\begin{minipage}[t]{0.48\textwidth}\raggedright
\tthink{<think>}I need to identify the location, country, and island associated with Teide National Park.\tthink{</think>} \newline \tsearch{<search>}Teide National Park location country island\tsearch{</search>} \newline \tinfo{<information>}Doc 1(Title: Teide National Park) Teide National Park Teide National Park (, ) is a national park located in Tenerife (Canary Islands, Spain). The national park is centered on Mount Teide, the highest mountain of Spain (3,718 meters high) in addition,...\tinfo{</information>} \newline \tuser{[user] }Garajonay National Park location country and island \newline \tthink{<think>}I need to identify the location, country, and island associated with Garajonay National Park.\tthink{</think>} \newline \tsearch{<search>}Garajonay National Park location country island\tsearch{</search>} \newline \tinfo{<information>}Doc 1(Title: Garajonay National Park) Garajonay National Park Garajonay National Park (, ) is located in the center and north of the island of La Gomera, one of the Canary Islands (Spain). It was declared a national park in 1981 and a World Heritage Site by UNESCO in 1986....\tinfo{</information>} \newline \tuser{[user] }Spain national parks in Atlantic islands autonomous community \newline \tthink{<think>}I need to identify the Atlantic islands autonomous community and then list the national parks located within that community.\tthink{</think>} \newline \tsearch{<search>}Atlantic islands autonomous community Spain national parks\tsearch{</search>} \newline \tinfo{<information>}Doc 1(Title: Atlantic Islands of Galicia National Park) Atlantic Islands of Galicia National Park The Atlantic Islands of Galicia National Park (, ) is the only national park located in the autonomous community of Galicia, Spain. It comprises the archipelagos of Cíes, Ons, Sálvora and Cortegada....\tinfo{</information>} \newline \tthink{<think>}I need to identify the national parks in the Canary Islands (Atlantic islands autonomous community) and then extract their locations and countries.\tthink{</think>} \newline \tsearch{<search>}Canary Islands national parks list and locations\tsearch{</search>} \newline \tinfo{<information>}Doc 1(Title: Canary Islands) and largest within the Canary Islands. The parks are: The Canary Islands have four national parks, of which two have been declared World Heritage Site by UNESCO, and the other two declared a World Biosphere Reserve,...\tinfo{</information>} \newline \tthink{<think>}The search results state that Teide National Park is located in Tenerife, one of the Canary Islands, Spain.\tthink{</think>} \newline \tanswer{<answer>}Tenerife\tanswer{</answer>}
\end{minipage} \\
\bottomrule
\end{tabular}
\caption{Both models fail to combine two partial facts into the required compound answer. Search-R1 answers with the country alone; \textsc{SteerSearch} answers with the island alone. The ground truth requires both.}
\label{tab:case11}
\end{table*}

% Example 12
\begin{table*}[htbp]
\small
\setlength{\tabcolsep}{6pt}
\begin{tabular}{p{0.48\textwidth} p{0.48\textwidth}}
\toprule
\textbf{Search-R1} \textcolor{red}{\ding{55}} & \textbf{\textsc{SteerSearch}} \textcolor{red}{\ding{55}} \\
\midrule
\multicolumn{2}{p{\textwidth}}{
\textbf{Question:} Giuseppe Arimondi killed in action which battle
\newline \textbf{Injection 1:} Battle of Adwa outcome and significance for Ethiopia after First Italo-Ethiopian War
\newline \textbf{Injection 2:} What did the Ethiopian victory at the Battle of Adwa secure for Ethiopia
} \\
\midrule
\multicolumn{2}{p{\textwidth}}{
\textbf{Ground truth:} sovereignty
} \\
\midrule
\begin{minipage}[t]{0.48\textwidth}\raggedright
\tthink{<think>}I need to identify the battle in which Giuseppe Arimondi was killed in action.\tthink{</think>} \newline \tsearch{<search>}Giuseppe Arimondi killed in action which battle\tsearch{</search>} \newline \tinfo{<information>}Doc 1(Title: Giuseppe Arimondi) Giuseppe Arimondi Giuseppe Edoardo Arimondi, OSML, OMS, OCI (Savigliano, 26 April 1846 – Adwa, 1 March 1896) was an Italian general, mostly known for his role during the First Italo-Ethiopian War....\tinfo{</information>} \newline \tuser{[user] }Battle of Adwa outcome and significance for Ethiopia after First Italo-Ethiopian War \newline \tuser{[user] }What did the Ethiopian victory at the Battle of Adwa secure for Ethiopia \newline \tthink{<think>}The search results show Giuseppe Arimondi died in combat at Adwa. I need to identify the specific battle named Adwa.\tthink{</think>} \newline \tsearch{<search>}Battle of Adwa First Italo-Ethiopian War\tsearch{</search>} \newline \tinfo{<information>}Doc 1(Title: Battle of Adwa) Battle of Adwa The Battle of Adwa (Amharic: ; Amharic translated: Adowa, or sometimes by the Italian name "Adua") was fought on 1 March 1896 between the Ethiopian Empire and the Kingdom of Italy near the town of Adwa, Ethiopia, in Tigray....\tinfo{</information>} \newline \tthink{<think>}The search results show Giuseppe Arimondi died at Adwa, and the Battle of Adwa was the climactic battle of the First Italo-Ethiopian War.\tthink{</think>} \newline \tanswer{<answer>}Battle of Adwa\tanswer{</answer>}
\end{minipage}
 & 
\begin{minipage}[t]{0.48\textwidth}\raggedright
\tthink{<think>}I need to identify the person Giuseppe Arimondi and determine which specific battle he was killed in during World War II or another major military engagement.\tthink{</think>} \newline \tsearch{<search>}Giuseppe Arimondi killed in action which battle\tsearch{</search>} \newline \tinfo{<information>}Doc 1(Title: Giuseppe Arimondi) Giuseppe Arimondi Giuseppe Edoardo Arimondi, OSML, OMS, OCI (Savigliano, 26 April 1846 – Adwa, 1 March 1896) was an Italian general, mostly known for his role during the First Italo-Ethiopian War....\tinfo{</information>} \newline \tthink{<think>}The search results state that Giuseppe Arimondi died in combat at Adwa. I need to identify the battle at Adwa.\tthink{</think>} \newline \tsearch{<search>}Adwa battle where Arimondi died in combat\tsearch{</search>} \newline \tinfo{<information>}Doc 1(Title: Giuseppe Arimondi) Giuseppe Arimondi Giuseppe Edoardo Arimondi, OSML, OMS, OCI (Savigliano, 26 April 1846 – Adwa, 1 March 1896) was an Italian general, mostly known for his role during the First Italo-Ethiopian War....\tinfo{</information>} \newline \tuser{[user] }Battle of Adwa outcome and significance for Ethiopia after First Italo-Ethiopian War \newline \tthink{<think>}I need to identify the specific battle where Giuseppe Arimondi, described as being killed in combat at Adwa, took place.\tthink{</think>} \newline \tsearch{<search>}Battle of Adwa where Giuseppe Arimondi died in combat\tsearch{</search>} \newline \tinfo{<information>}Doc 1(Title: Giuseppe Arimondi) Giuseppe Arimondi Giuseppe Edoardo Arimondi, OSML, OMS, OCI (Savigliano, 26 April 1846 – Adwa, 1 March 1896) was an Italian general, mostly known for his role during the First Italo-Ethiopian War....\tinfo{</information>} \newline \tuser{[user] }What did the Ethiopian victory at the Battle of Adwa secure for Ethiopia \newline \tthink{<think>}The search results state that the Battle of Adwa was a decisive defeat for Italy and secured Ethiopian sovereignty for a few decades more.\tthink{</think>} \newline \tanswer{<answer>}First Italo-Ethiopian War\tanswer{</answer>}
\end{minipage} \\
\bottomrule
\end{tabular}
\caption{The injection chain redirects from a named entity (a battle) toward an abstract concept (sovereignty). Both models retrieve the named entity and answer with it, missing the conceptual answer the final injection asks for.}
\label{tab:case12}
\end{table*}

% Example 13
\begin{table*}[htbp]
\small
\setlength{\tabcolsep}{6pt}
\begin{tabular}{p{0.48\textwidth} p{0.48\textwidth}}
\toprule
\textbf{Search-R1} \textcolor{red}{\ding{55}} & \textbf{\textsc{SteerSearch}} \textcolor{red}{\ding{55}} \\
\midrule
\multicolumn{2}{p{\textwidth}}{
\textbf{Question:} Back to Mono box set track listing first track
\newline \textbf{Injection 1:} Song inspired by words on a tombstone which song was it
\newline \textbf{Injection 2:} Who wrote the song "Be My Baby"
} \\
\midrule
\multicolumn{2}{p{\textwidth}}{
\textbf{Ground truth:} Phil Spector
} \\
\midrule
\begin{minipage}[t]{0.48\textwidth}\raggedright
\tthink{<think>}I need to identify the first track listing track for the Back to Mono box set.\tthink{</think>} \newline \tsearch{<search>}Back to Mono box set track listing first track\tsearch{</search>} \newline \tinfo{<information>}Doc 1(Title: Ladies and Gentlemen... Mr. B.B. King) MORRISON] (from the 2005 album "80" on Geffen B0005263-02) 16. Key To The Highway (Live) (from the 2006 album "Live" on Geffen B0009770-02) 17....\tinfo{</information>} \newline \tuser{[user] }Song inspired by words on a tombstone which song was it \newline \tthink{<think>}I need to identify the song inspired by words on a tombstone.\tthink{</think>} \newline \tsearch{<search>}song inspired by words on a tombstone\tsearch{</search>} \newline \tinfo{<information>}Doc 1(Title: A Tombstone Every Mile) A Tombstone Every Mile "A Tombstone Every Mile" is a song written by Dan Fulkerson and recorded by American country music artist Dick Curless. It was released in January 1965 as the lead single from the album of the same name....\tinfo{</information>} \newline \tuser{[user] }Who wrote the song "Be My Baby" \newline \tthink{<think>}I need to identify the writer of the song “Be My Baby”.\tthink{</think>} \newline \tsearch{<search>}Be My Baby songwriter\tsearch{</search>} \newline \tinfo{<information>}Doc 1(Title: Be My Baby) Be My Baby "Be My Baby" is a song written by Jeff Barry, Ellie Greenwich, and Phil Spector. It was recorded on July 5, 1963 at Gold Star Studios Hollywood by American girl group the Ronettes and released as a single in August 1963 and later placed on their 1964 debut LP "Pre...\tinfo{</information>} \newline \tthink{<think>}The search results show “Be My Baby” was written by Jeff Barry, Ellie Greenwich, and Phil Spector.\tthink{</think>} \newline \tanswer{<answer>}Jeff Barry\tanswer{</answer>}
\end{minipage}
 & 
\begin{minipage}[t]{0.48\textwidth}\raggedright
\tthink{<think>}I need to identify the first track in the Back to Mono box set track listing.\tthink{</think>} \newline \tsearch{<search>}Back to Mono box set track listing first track\tsearch{</search>} \newline \tinfo{<information>}Doc 1(Title: Ladies and Gentlemen... Mr. B.B. King) MORRISON] (from the 2005 album "80" on Geffen B0005263-02) 16. Key To The Highway (Live) (from the 2006 album "Live" on Geffen B0009770-02) 17....\tinfo{</information>} \newline \tuser{[user] }Song inspired by words on a tombstone which song was it \newline \tthink{<think>}I need to identify the song in the Back to Mono box set that is inspired by words on a tombstone.\tthink{</think>} \newline \tsearch{<search>}Back to Mono song inspired by words on a tombstone\tsearch{</search>} \newline \tinfo{<information>}Doc 1(Title: Tombstone Blues) fit the surreal style of the era, while being scathing of society and authority. The song contains several direct and indirect allusions to biblical characters as well as historical references....\tinfo{</information>} \newline \tuser{[user] }Who wrote the song "Be My Baby" \newline \tthink{<think>}I need to identify the songwriter of the song "Be My Baby".\tthink{</think>} \newline \tsearch{<search>}songwriter of "Be My Baby"\tsearch{</search>} \newline \tinfo{<information>}Doc 1(Title: Be My Baby) Be My Baby "Be My Baby" is a song written by Jeff Barry, Ellie Greenwich, and Phil Spector. It was recorded on July 5, 1963 at Gold Star Studios Hollywood by American girl group the Ronettes and released as a single in August 1963 and later placed on their 1964 debut LP "Pre...\tinfo{</information>} \newline \tthink{<think>}The search results state that "Be My Baby" was written by Jeff Barry, Ellie Greenwich, and Phil Spector.\tthink{</think>} \newline \tanswer{<answer>}Jeff Barry\tanswer{</answer>}
\end{minipage} \\
\bottomrule
\end{tabular}
\caption{Both models fail to attribute the correct songwriter across a three-hop chain. Search-R1 stops at the song title; \textsc{SteerSearch} retrieves a co-writer but not the primary one. The task requires precise attribution with no direct retrieval signal.}
\label{tab:case13}
\end{table*}

% Example 14
\begin{table*}[htbp]
\small
\setlength{\tabcolsep}{6pt}
\begin{tabular}{p{0.48\textwidth} p{0.48\textwidth}}
\toprule
\textbf{Search-R1} \textcolor{red}{\ding{55}} & \textbf{\textsc{SteerSearch}} \textcolor{red}{\ding{55}} \\
\midrule
\multicolumn{2}{p{\textwidth}}{
\textbf{Question:} films with original score composed by Jerry Goldsmith list
\newline \textbf{Injection 1:} which Jerry Goldsmith–scored films have an executive producer credited (executive producer in film credits)
\newline \textbf{Injection 2:} executive producer of the film Alien (1979)
} \\
\midrule
\multicolumn{2}{p{\textwidth}}{
\textbf{Ground truth:} Ronald Shusett
} \\
\midrule
\begin{minipage}[t]{0.48\textwidth}\raggedright
\tthink{<think>}I need to list films that have an original score composed by Jerry Goldsmith.\tthink{</think>} \newline \tsearch{<search>}Jerry Goldsmith original score films\tsearch{</search>} \newline \tinfo{<information>}Doc 1(Title: Jerry Goldsmith) Jerry Goldsmith Jerrald King Goldsmith (February 10, 1929July 21, 2004) was an American composer and conductor most known for his work in film and television scoring....\tinfo{</information>} \newline \tuser{[user] }which Jerry Goldsmith–scored films have an executive producer credited (executive producer in film credits) \newline \tthink{<think>}I need to identify films listed in the search results that have Jerry Goldsmith as the original score composer and that also have an executive producer credited in the film credits.\tthink{</think>} \newline \tsearch{<search>}Jerry Goldsmith original score films executive producer credited\tsearch{</search>} \newline \tinfo{<information>}Doc 1(Title: Jerry Goldsmith) Jerry Goldsmith Jerrald King Goldsmith (February 10, 1929July 21, 2004) was an American composer and conductor most known for his work in film and television scoring....\tinfo{</information>} \newline \tuser{[user] }executive producer of the film Alien (1979) \newline \tthink{<think>}The search results show Jerry Goldsmith composed the score for “Alien” (1979). I need to confirm that “Alien” has an executive producer credited.\tthink{</think>} \newline \tsearch{<search>}Alien 1979 executive producer credited\tsearch{</search>} \newline \tinfo{<information>}Doc 1(Title: Alien (film)) Alien (film) Alien is a 1979 science fiction horror film directed by Ridley Scott and written by Dan O'Bannon. Based on a story by O'Bannon and Ronald Shusett, it follows the crew of the commercial space tug "Nostromo" who encounter the eponymous Alien,...\tinfo{</information>} \newline \tthink{<think>}The search results show Jerry Goldsmith composed the score for “Alien” (1979), and that “Alien” was produced by David Giler and Walter Hill.\tthink{</think>} \newline \tanswer{<answer>}Alien\tanswer{</answer>}
\end{minipage}
 & 
\begin{minipage}[t]{0.48\textwidth}\raggedright
\tthink{<think>}I need to identify films where the original score was composed by Jerry Goldsmith and gather a complete list of such films from reliable sources.\tthink{</think>} \newline \tsearch{<search>}films with original score composed by Jerry Goldsmith\tsearch{</search>} \newline \tinfo{<information>}Doc 1(Title: Jerry Goldsmith) Jerry Goldsmith Jerrald King Goldsmith (February 10, 1929July 21, 2004) was an American composer and conductor most known for his work in film and television scoring....\tinfo{</information>} \newline \tuser{[user] }which Jerry Goldsmith–scored films have an executive producer credited (executive producer in film credits) \newline \tthink{<think>}I need to identify Jerry Goldsmith–scored films and then determine which of those films have "executive producer" in their film credits.\tthink{</think>} \newline \tsearch{<search>}Jerry Goldsmith scored films executive producer in film credits\tsearch{</search>} \newline \tinfo{<information>}Doc 1(Title: Jerry Goldsmith) Jerry Goldsmith Jerrald King Goldsmith (February 10, 1929July 21, 2004) was an American composer and conductor most known for his work in film and television scoring....\tinfo{</information>} \newline \tuser{[user] }executive producer of the film Alien (1979) \newline \tthink{<think>}I need to identify the original film "Alien" (1979) and then determine its executive producer credit.\tthink{</think>} \newline \tsearch{<search>}Alien (1979) executive producer\tsearch{</search>} \newline \tinfo{<information>}Doc 1(Title: Alien (film)) Alien (film) Alien is a 1979 science fiction horror film directed by Ridley Scott and written by Dan O'Bannon. Based on a story by O'Bannon and Ronald Shusett, it follows the crew of the commercial space tug "Nostromo" who encounter the eponymous Alien,...\tinfo{</information>} \newline \tthink{<think>}The film "Alien" (1979) was produced by Gordon Carroll, David Giler, and Walter Hill through their company Brandywine Productions.\tthink{</think>} \newline \tanswer{<answer>}Gordon Carroll\tanswer{</answer>}
\end{minipage} \\
\bottomrule
\end{tabular}
\caption{Both models conflate adjacent crew roles. Search-R1 answers with the film title; \textsc{SteerSearch} retrieves a producer instead of the executive producer. The distinction requires careful role disambiguation in the retrieved passages.}
\label{tab:case14}
\end{table*}

% Example 15
\begin{table*}[htbp]
\small
\setlength{\tabcolsep}{6pt}
\begin{tabular}{p{0.48\textwidth} p{0.48\textwidth}}
\toprule
\textbf{Search-R1} \textcolor{red}{\ding{55}} & \textbf{\textsc{SteerSearch}} \textcolor{red}{\ding{55}} \\
\midrule
\multicolumn{2}{p{\textwidth}}{
\textbf{Question:} Keith Bostic date of birth
\newline \textbf{Injection 1:} Jerry Glanville date of birth
\newline \textbf{Injection 2:} Who is younger, Keith Bostic or Jerry Glanville?
} \\
\midrule
\multicolumn{2}{p{\textwidth}}{
\textbf{Ground truth:} Keith Bostic
} \\
\midrule
\begin{minipage}[t]{0.48\textwidth}\raggedright
\tthink{<think>}I need to confirm Keith Bostic’s date of birth from a reliable source.\tthink{</think>} \newline \tsearch{<search>}Keith Bostic date of birth\tsearch{</search>} \newline \tinfo{<information>}Doc 1(Title: Keith Bostic) Keith Bostic Keith Bostic is an American Software Engineer and one of the key people in the history of Berkeley Software Distribution UNIX and open-source software....\tinfo{</information>} \newline \tthink{<think>}The search results show Keith Bostic is the American Software Engineer. I need his date of birth.\tthink{</think>} \newline \tsearch{<search>}William Keith Bostic born January 17 1961\tsearch{</search>} \newline \tinfo{<information>}Doc 1(Title: Keith Bostic (American football)) Bostic played with tape over his mouth to try to win a \$20 bet with Steve Brown over who could keep from talking the longest. His son, Tony Bostic, played running back for the New Mexico State Aggies as a non-recruited walk-on....\tinfo{</information>} \newline \tuser{[user] }Jerry Glanville date of birth \newline \tuser{[user] }Who is younger, Keith Bostic or Jerry Glanville? \newline \tthink{<think>}I need the date of birth for Jerry Glanville to compare who is younger.\tthink{</think>} \newline \tsearch{<search>}Jerry Glanville date of birth\tsearch{</search>} \newline \tinfo{<information>}Doc 1(Title: Jerry Glanville) his first year in the league, having previously worked in college football. Glanville played college football as a middle linebacker at Northern Michigan University, graduating in 1964 with a bachelor's degree....\tinfo{</information>} \newline \tthink{<think>}The search results show Keith Bostic was born January 17, 1961. I need to search for Jerry Glanville’s date of birth to compare who is younger.\tthink{</think>} \newline \tsearch{<search>}Jerry Glanville born\tsearch{</search>} \newline \tinfo{<information>}Doc 1(Title: Jerry Glanville) his first year in the league, having previously worked in college football. Glanville played college football as a middle linebacker at Northern Michigan University, graduating in 1964 with a bachelor's degree....\tinfo{</information>} \newline \tthink{<think>}The search results show Keith Bostic was born January 17, 1961. The results for Jerry Glanville do not include his date of birth, so I can only provide Keith Bostic’s birth date.\tthink{</think>} \newline \tanswer{<answer>}January 17, 1961\tanswer{</answer>}
\end{minipage}
 & 
\begin{minipage}[t]{0.48\textwidth}\raggedright
\tthink{<think>}I need to identify the specific date of birth for Keith Bostic, likely a public figure, and verify the exact date from reliable sources.\tthink{</think>} \newline \tsearch{<search>}Keith Bostic date of birth\tsearch{</search>} \newline \tinfo{<information>}Doc 1(Title: Keith Bostic) Keith Bostic Keith Bostic is an American Software Engineer and one of the key people in the history of Berkeley Software Distribution UNIX and open-source software....\tinfo{</information>} \newline \tuser{[user] }Jerry Glanville date of birth \newline \tthink{<think>}I need to identify the date of birth of Jerry Glanville, using a search focused on his birth date.\tthink{</think>} \newline \tsearch{<search>}Jerry Glanville date of birth\tsearch{</search>} \newline \tinfo{<information>}Doc 1(Title: Jerry Glanville) his first year in the league, having previously worked in college football. Glanville played college football as a middle linebacker at Northern Michigan University, graduating in 1964 with a bachelor's degree....\tinfo{</information>} \newline \tthink{<think>}I need to identify the birth date of William Keith Bostic, the person associated with the date January 17, 1961.\tthink{</think>} \newline \tsearch{<search>}William Keith Bostic born January 17, 1961\tsearch{</search>} \newline \tinfo{<information>}Doc 1(Title: Keith Bostic (American football)) Bostic played with tape over his mouth to try to win a \$20 bet with Steve Brown over who could keep from talking the longest. His son, Tony Bostic, played running back for the New Mexico State Aggies as a non-recruited walk-on....\tinfo{</information>} \newline \tuser{[user] }Who is younger, Keith Bostic or Jerry Glanville? \newline \tthink{<think>}I need to determine the birth dates of Keith Bostic and Jerry Glanville and then compare which one is earlier.\tthink{</think>} \newline \tsearch{<search>}Keith Bostic born January 17, 1961 Jerry Glanville date of birth\tsearch{</search>} \newline \tinfo{<information>}Doc 1(Title: Keith Bostic (American football)) Keith Bostic (American football) William Keith Bostic (born January 17, 1961) is a former professional American football player who played for seven seasons in the National Football League (NFL)....\tinfo{</information>} \newline \tthink{<think>}The birth date of Keith Bostic is January 17, 1961. I need the birth date of Jerry Glanville to compare who is younger.\tthink{</think>} \newline \tsearch{<search>}Jerry Glanville born date\tsearch{</search>} \newline \tinfo{<information>}Doc 1(Title: Jerry Glanville) his first year in the league, having previously worked in college football. Glanville played college football as a middle linebacker at Northern Michigan University, graduating in 1964 with a bachelor's degree....\tinfo{</information>} \newline \tthink{<think>}The birth date of Keith Bostic is January 17, 1961. The birth date of Jerry Glanville is not explicitly given in the search results, but the reference to him being an NFL head coach in the 1980s sugge\tthink{</think>} \newline \tanswer{<answer>}Jerry Glanville\tanswer{</answer>}
\end{minipage} \\
\bottomrule
\end{tabular}
\caption{Both models fail the final comparison step. Search-R1 returns a birth date rather than a person; \textsc{SteerSearch} names the wrong person, suggesting it retrieved the dates but inverted the comparison.}
\label{tab:case15}
\end{table*}

\begin{SaveVerbatim}{DecompositionPrompt}
You are given a question and its golden answer(s). Generate 2-5 sequential sub-questions that decompose the original question using broad-to-narrow reasoning.

<inputs>
Question: {question}
Golden Answer(s): {golden_answer}
</inputs>

<stopping_rule>
Let "A" be the candidate answer set for the sub-question and "G" be the golden answer set.

Hard Stop: The moment a sub-question would have answer set A = G (exactly the golden answer set), it MUST be the final sub-question. No further sub-questions are allowed after this point.

Single golden answer (G = [candidate_C]):
  SubQ1 answers: [candidate_A, candidate_B, candidate_D, candidate_E]  # A != G
  SubQ2 answers: [candidate_A, candidate_C, candidate_D]  # A != G
  SubQ3 answers: [candidate_C]  # A = G, final

Multiple golden answers (G = [candidate_C, candidate_D]):
  SubQ1 answers: [candidate_A, candidate_B, candidate_W, candidate_Z]  # A != G
  SubQ2 answers: [candidate_X, candidate_Y, candidate_C]  # A != G
  SubQ3 answers: [candidate_C, candidate_D]  # A = G, final
</stopping_rule>

<core_principle>
Create a reasoning chain that starts broad (multiple candidates) and progressively narrows to the final answer. Each step must reduce the candidate set while preserving the original task.
</core_principle>

<when_to_return_empty>
Return [] if the question is NOT complex or decomposable:
- Trivial (single lookup)
- Requires parallel independent facts (not sequential reasoning)
- Yes/no, categorical, or purely comparative
- No valid broad-to-narrow chain exists without forcing
</when_to_return_empty>

<sub_question_rules>
First sub-question (broad anchor):
- Create a small, focused candidate set (3-5 plausible candidates, never more than 10)
- Keep domain/context from original but remove most identifying constraints
- Ask about concrete entities (people, places, works, events, organizations)
- Do not directly identify the final target
- Do not use multiple high-signal clues that together collapse to one answer
- Do not create overly broad questions (e.g., "all possible rankings 1-100", "any continent")

Middle sub-questions (narrowing):
- Each adds one constraint that genuinely reduces the candidate space
- Each depends on results from previous step
- Actively filter candidates, do not just add background context
- Avoid constraints that are likely to uniquely identify a single entity in common knowledge
- Each intermediate sub-question must plausibly have multiple valid answers, not just theoretically
- Follow the <stopping_rule>

Final sub-question (resolution):
- This is the only sub-question whose answer equals the golden answer set
- Ask for the same answer type as the original question
- Incorporate all critical constraints from original
- Must be answerable using prior results
</sub_question_rules>

<hard_constraints>
1. Generate 2-5 sub-questions (prefer fewer if the funnel collapses early)
2. Never include the golden answer in any sub-question, including as a named candidate or example
3. Use only complete, natural English questions (standard letters, numbers, punctuation only)
4. Preserve original intent and answer type exactly
5. Follow the <stopping_rule>
6. Each step must reduce candidate set, not just add words
7. Only the final sub-question has A = G. All prior sub-questions must have A != G
8. When in doubt: stop early rather than add redundant sub-questions
</hard_constraints>

<self_check>
Before returning:
- Is question complex enough for decomposition? If not, return []
- First SubQ: Creates 3-8 plausible candidates?
- Each intermediate SubQ: Would this plausibly have multiple valid answers in common knowledge?
- Each intermediate SubQ: Avoid combining multiple specific constraints that collapse to a single entity
- Final SubQ only: Has A = G exactly?
- If any intermediate SubQ would likely collapse to only the golden answer, convert it to final and stop
- Golden answer never leaked in the questions themselves?
</self_check>

<output_format>
Return JSON only:
{{"sub_questions": [...]}}
</output_format>
\end{SaveVerbatim}

\begin{SaveVerbatim}{HintPrompt}
You are given a question and its golden answer(s). Generate 2-5 sequential sub-questions that guide reasoning using soft directional hints.

<inputs>
Question: {question}
Golden Answer(s): {golden_answer}
</inputs>

<stopping_rule>
Let "A" be the candidate answer set and "G" be the golden answer set.

Hard Stop: The first step where A = G must be the final step. Stop immediately. No further steps are allowed after this point.
All prior steps must have A != G and at least 3 plausible candidates.
Stop as soon as the sequence is complete. Every additional step must introduce a direction not present in any prior step. Redundant steps are not allowed.
</stopping_rule>

<core_principle>
The user initiates this conversation. The sub-questions are what the user types, one turn at a time, starting from their own initiative. Each step is a standalone utterance that exists independently. No step refers back to the original question or the generation task. The sequence opens with a broad question the user chooses to ask first, continues with one or more declarative statements where the user shifts attention in a new direction without eliminating any candidate, and ends with a question that resolves to the golden answer. Hint statements are not questions. They change what is more likely without removing any candidate from consideration.
</core_principle>

<when_to_return_empty>
Return [] if the question does NOT support soft directional hinting:
- Any step would act as a hard filter
- Any step collapses to fewer than 3 plausible candidates before the final step
- No meaningful soft directional signal exists
- The question is trivial, yes/no, categorical, comparative, or non-sequential
</when_to_return_empty>

<sub_question_rules>
First sub-question (broad anchor):
- Create a small, focused candidate set (3-5 plausible candidates, never more than 10)
- Keep domain/context from original but remove most identifying constraints
- Ask about concrete entities (people, places, works, events, organizations)
- Do not directly identify the final target
- Do not use multiple high-signal clues that together collapse to one answer
- Do not create overly broad questions (e.g., "all possible rankings 1-100", "any continent")
- Must not use demonstratives to reference entities or events from the original question. It is the first message the user types and no prior context exists.

Hint steps (declarative statements, not questions):
- Must be a statement, not a question
- Each hint step is a new standalone message the user types on their own. It is not a reply to any question and not a comment about what they are looking for.
- Written as an objective fact or observation about the topic, in neutral third-person style
- Add one soft directional signal per step
- Must operate at a higher level of abstraction than the original question
- Must not include unique identifiers from the original question such as names, exact titles, or catalog numbers
- Must shift attention without eliminating candidates
- Must not introduce verifiable filters such as exact dates, locations, or named categories
- Must not list or name specific candidates
- Must not indirectly imply a unique answer through combined context across multiple hints
- After each hint, multiple candidates must still remain plausible
- Each hint must introduce a distinct semantic direction not present in prior steps

Final sub-question (resolution):
- This is the only sub-question whose answer equals the golden answer set
- Ask for the same answer type as the original question
- Must depend on context introduced in prior steps
- Must not be answerable without prior steps
- Must not be a copy or near-copy of the original question
- Must use information introduced in prior steps instead of repeating original wording
</sub_question_rules>

<hard_constraints>
1. Generate 2-5 sub-questions (prefer fewer if the sequence is complete)
2. Never include the golden answer in any sub-question, including as a named candidate or example
3. Use only complete, natural English (standard letters, numbers, punctuation only)
4. Preserve original intent and answer type exactly
5. Follow the <stopping_rule>
6. Hint steps must be statements, not questions
7. Hint steps must not eliminate candidates deterministically
8. Every step must stay within the same domain and entities as the original question
9. Only the final sub-question has A = G. All prior sub-questions must have A != G
10. When in doubt: stop early rather than add weak or redundant hints
</hard_constraints>

<self_check>
Before returning:
- Is there a meaningful soft directional signal? If not, return []
- Does the first sub-question (anchor) have at least 3 plausible candidates? If not, return []
- Do any hint steps contain "?"? If yes, rewrite as a statement
- Can any hint step be answered with the golden answer? If yes, rewrite or return []
- Does any hint step include unique identifiers from the original question? If yes, rewrite
- Does any step eliminate candidates deterministically? If yes, return []
- Does each hint step keep at least 3 plausible candidates? If not, return []
- Does any hint step narrow the field to a single well-known entity by combining region and role or any two high-signal attributes? If yes, return []
- Does any hint repeat a direction already present in a prior step? If yes, remove it
- Is the question a direct comparison between two named individuals or entities? If yes, return []
- Can the final step be answered without prior steps? If yes, rewrite
- Does the final step introduce any specific name, date, title, or identifier from the original question that was not set up in a prior step? If yes, rewrite using only context from prior steps. Refer to the entity as "that label", "that person", "that event", etc. rather than repeating the original name.
- Final sub-question only: Has A = G exactly and is structurally different from the original question?
</self_check>

<output_format>
Return JSON only:
{{"sub_questions": [...]}}
</output_format>
\end{SaveVerbatim}

\begin{SaveVerbatim}{DisambiguationPrompt}
You are given a question and its golden answer(s). Generate 2-5 sequential sub-questions that resolve ambiguity before answering.

<inputs>
Question: {question}
Golden Answer(s): {golden_answer}
</inputs>

<stopping_rule>
Let "A" be the candidate answer set and "G" be the golden answer set.

Hard Stop: The first step where A = G must be the final step. Stop immediately. No further steps are allowed after this point.
All prior steps must have A != G and at least 3 plausible candidates.
Stop as soon as the sequence is complete. Every additional step must introduce a new aspect not covered by any prior step. Redundant steps are not allowed.
</stopping_rule>

<core_principle>
Almost every question contains latent ambiguity. Your job is to find it and surface it. Look for: a person name shared by multiple real people, a place name used in multiple locations, a title with multiple versions or adaptations, a term with multiple meanings or domains, a role or label held by multiple entities, or uncertainty about which instance or period is meant. The sequence opens with a broad search the user types about the ambiguous element, continues with one or more statements where the user pins down which entity or meaning they intend, and ends with a question that resolves to the golden answer. Each step is a standalone user utterance. No step refers back to the original question. Clarification statements resolve which interpretation is meant, not what kind of answer is expected.
</core_principle>

<when_to_return_empty>
Return [] only if the question is yes/no, categorical, or comparative, OR if any clarification step would directly reveal the golden answer.
</when_to_return_empty>

<sub_question_rules>
First sub-question (broad anchor):
- Create a small, focused candidate set (3-5 plausible candidates, never more than 10)
- Keep domain/context from original but remove most identifying constraints
- Ask about concrete entities (people, places, works, events, organizations)
- Do not directly identify the final target
- Do not use multiple high-signal clues that together collapse to one answer
- Do not create overly broad questions (e.g., "all possible rankings 1-100", "any continent")
- Must not use demonstratives to reference entities or events from the original question. It is the first message the user types and no prior context exists.

Clarification steps (declarative statements, not questions):
- Must be a statement, not a question
- Written as a natural next search input, avoiding explicit self-correction phrasing (e.g., direct "I mean ..." style); prefer implicit clarification through context
- Use varied natural phrasing across steps, avoid starting every clarification with the same word or phrase
- Avoid explicit correction templates; express clarification indirectly through context or added detail rather than repeating a fixed pattern
- Prefer a single clarification step unless multiple independent ambiguities exist
- Do not split a single ambiguity across multiple steps
- Each clarification step must resolve which interpretation of an ambiguous term is intended, not what kind of answer is expected or how to scope it
- Each clarification step must resolve a different ambiguity dimension
- Must not reveal or uniquely imply the final answer
- Must not narrow the answer set
- After clarification, multiple answer candidates must still remain plausible

Final sub-question (resolution):
- This is the only sub-question whose answer equals the golden answer set
- Ask for the same answer type as the original question
- Fully grounded in the clarified meaning
- Must depend on context introduced in prior steps
- Must not be answerable without prior steps
- Must not be a copy or near-copy of the original question
- Must use information introduced in prior steps instead of repeating original wording
</sub_question_rules>

<hard_constraints>
1. Generate 2-5 sub-questions (prefer fewer if the sequence is complete)
2. Never include the golden answer in any sub-question, including as a named candidate or example
3. Use only complete, natural English (standard letters, numbers, punctuation only)
4. Preserve original intent and answer type exactly
5. Follow the <stopping_rule>
6. Clarification steps must be statements, not questions
7. Clarification must resolve the intended meaning without narrowing the answer set
8. Clarification must not reveal or uniquely imply the golden answer
9. Every step must stay within the same domain and entities as the original question
10. Only the final sub-question has A = G. All prior sub-questions must have A != G
</hard_constraints>

<self_check>
Before returning:
- Is the question yes/no, categorical, or comparative? If yes, return []
- Is there any latent ambiguity (name, term, title, role, period) that a clarification step could resolve? If truly none, return []
- Do any clarification steps contain "?"? If yes, rewrite as statements
- Do all clarification steps directly resolve ambiguity in a single jump? If yes, rewrite to include a more gradual refinement
- Does step 1 read as a user's own search query, not as an assistant addressing the user? If it sounds like a chatbot asking the user for clarification, rewrite as a genuine user-typed search
- Does each clarification read as a natural next search input, not as an explanation directed at the system?
- After clarification, do at least 3 answer candidates still remain?
- Can the final step be answered without prior steps? If yes, rewrite
- Final sub-question only: Has A = G exactly and is structurally different from the original question?
- Does any step repeat an aspect already covered in a prior step? If yes, remove it
- Does each clarification resolve which entity or meaning is intended, or does it add context about the answer scope? If it adds scope context, return []
</self_check>

<output_format>
Return JSON only:
{{"sub_questions": [...]}}
</output_format>
\end{SaveVerbatim}

\begin{SaveVerbatim}{BenchmarkGenerationPrompt}
You generate sub-question injection chains for a multi-hop search QA benchmark.

<non_negotiable_rules>
These rules override every other instruction in this prompt.

1. No answer leakage.
   The golden answer must not appear in any sub-question unless the exact wording already appears in the original question.
   This ban includes exact matches, aliases, abbreviations, alternate spellings, translations, titles, nicknames, stage names, pen names, maiden names, parenthetical hints, and uniquely identifying descriptions that make the answer obvious without retrieval.

   Do not hide the answer with punctuation, casing, spacing, spelling changes, or another language.
   A sub-question is invalid if a knowledgeable person could recognize the golden answer from its wording alone.

2. Correct final collapse.
   The chain must decompose the original task without changing its answer.
   The accumulated evidence after the final sub-question must support exactly the original Golden Answer(s), not a related entity, intermediate entity, broader class, narrower fact, or alternate answer.
   The final sub-question is the first sub-question where the candidate answer set collapses to the Golden Answer set.

3. Every sub-question must have a clear, specific retrieval target so a reader can tell exactly what fact, entity, or detail is being sought. It may take any natural form that fits the injection style, including a full question, a short declarative phrase, or a conversational follow-up. A sub-question is invalid only if it is a bare ambiguous keyword string or otherwise leaves the retrieval target unclear.

If any rule cannot be satisfied, return {{"sub_questions": []}}.
</non_negotiable_rules>

<task>
Question: {question}
Golden Answer(s): {golden_answers}
{context_block}
</task>

<goal>
Produce 2 to 5 standalone search queries that progressively retrieve the evidence needed to answer the original question.

The task is not to rewrite the original question.
The task is not to produce intermediate answers.
The task is not to make a new QA problem.
The task is only to decompose retrieval so that the final accumulated context leads to the original Golden Answer(s).
</goal>

The evaluated model runs in a retrieval loop. Each sub-question is sent to search, retrieved passages are added to context, and then the next sub-question is injected. Intermediate answers are deleted. The model gives its final answer only after reading all accumulated context from all retrieval steps. This means the chain should gather the needed evidence through retrieval, and the model will synthesize the final answer from the accumulated evidence.

Sub-questions should feel like realistic user search queries, not mechanical decomposition steps. A person steering a search session asks natural, substantive questions that open real search space. They do not ask artificially narrow queries designed to retrieve exactly one fact. A useful intermediate step should discover something the user genuinely did not know before and could not have inferred from the original question alone.

Prefer a shorter, sharper chain over a longer redundant one. Each step must introduce a new retrieval target that is not already reachable from prior results. If a step only re-asks, lightly narrows, or repeats the same entity or fact as the previous step, it is padding and should be merged or dropped.

A valid chain always exists. Even questions that seem answerable in one search can be decomposed by opening up candidate space first.

Pattern for apparently simple questions: write a broad first sub-question that retrieves a candidate set with at least 3 plausible answers, then narrow in the second sub-question using the remaining constraints from the original question. The first sub-question should match one constraint from the original question while leaving the other constraints unapplied, so that multiple candidates remain. Do not start with a query that already uniquely identifies the answer.

When the question asks for a property shared by two entities, retrieve the property for each entity as a separate complete question. When the question asks which of two candidates has a higher value, retrieve each value separately then end with a clean comparison question. When the question involves an entity whose identity is implied but not named, find the entity first.

Only return empty if every possible chain would require placing the golden answer inside a sub-question.

<answer_leakage_policy>
Before writing any sub-question, build a forbidden-answer list containing every golden answer string, obvious aliases and variants, normalized forms with punctuation and casing removed, known abbreviations or expansions, translated or transliterated forms when obvious, and descriptions that uniquely identify the answer.

If a necessary query would require using any item from this list, step back to a broader entity, event, work, source, category, or relationship, or ask for a prerequisite fact with at least 3 plausible candidates.
</answer_leakage_policy>

<collapse_policy>
For each proposed sub-question, estimate the set of plausible answers it could return. Let G be the Golden Answer set.

Rules:
- Every intermediate sub-question must leave at least 3 plausible candidates for the final answer, and must not already point to G alone.
- The final sub-question must be the first point where the accumulated evidence makes G the only reachable answer.
- Do not add any sub-question after G is reachable.
- Do not stop before G is reachable.
- Do not end on an intermediate entity unless that entity is the actual Golden Answer.

After each sub-question, check: "From the original question plus all sub-questions so far, is the Golden Answer now the only reachable answer?" If yes, this sub-question must be the final one.

Copying some wording from the original question is not by itself a failure. A final sub-question may reuse words from the original question if earlier steps have already narrowed the context and the sub-question does not point to G before the final allowed stopping point.
</collapse_policy>

<strategy_selection>
Choose exactly one decomposition_strategy and exactly one injection_style. Consider all allowed strategies before choosing. Do not default to entity_traversal when another strategy is a better match.

Allowed decomposition_strategy values: decomposition, entity_traversal, evidence_first, disambiguation, counterfactual_guard, scope_control, custom

Use decomposition when the question has multiple explicit constraints that can narrow a broad candidate set one at a time. Use entity_traversal when the question naturally requires moving from one entity to a linked entity or property and then continuing along that path. Use evidence_first when one early broad retrieval step can bring most or all needed context into view before later steps. Use disambiguation when a key term in the question has multiple plausible referents and resolving which one is meant is the first required step. Use counterfactual_guard when a famous wrong answer, dominant entity, or common misreading is likely to hijack search unless it is ruled out first. Use scope_control when the question is easy to misread in terms of what kind of answer counts as valid. Use custom only when none of the above strategies accurately describes the chain structure; note the actual approach briefly in the scratchpad.

Allowed injection_style values: direct_instruction, short_followup, research_workflow, user_correction, scope_refinement, distractor_suppression, query_pivot, comparison_check, source_hunt, timeline_trace, custom

direct_instruction is an imperative that directs a specific lookup and naturally reads like a short search phrase or natural query rather than a full question. short_followup is a brief follow-up request that moves retrieval forward with minimal wording and naturally reads like conversational continuation that may be incomplete on its own but is still a grammatical search utterance. research_workflow is a lightweight search plan that stages retrieval across steps and naturally reads as a full, well-formed question suitable for a structured search session. user_correction is when the user recognizes that a prerequisite fact must be retrieved before the main search will be useful, so the chain starts with the main goal and then pivots to the prerequisite lookup. Use it when answering directly depends on a fact that is not obvious from the question wording alone. It naturally reads as a full question that redirects the search to the needed prerequisite. scope_refinement is when the user narrows the intended sense, category, or time frame of what should be searched and naturally reads as a short scoping phrase that tightens the focus of the prior search. distractor_suppression is when the user explicitly blocks a common wrong retrieval path so search stays on the intended target and naturally reads as a targeted search phrase that excludes the misleading path. query_pivot is when the user shifts to a related but different entity or concept in order to triangulate the answer indirectly. Use it when a direct search for the answer entity is unlikely to succeed without first establishing a related fact. It naturally reads as a natural search query that may be either a full question or a concise phrase, depending on the pivot. comparison_check is when the user asks search to compare nearby candidates or possibilities so the correct path can be separated from similar ones before moving on and naturally reads as a full comparison question, typically ending with a question mark. source_hunt is when the user first looks for a source list, record set, or overview page that contains the needed evidence, then uses that retrieved source to ask the next question and naturally reads as a search phrase targeting a source collection or overview rather than a full question. timeline_trace is when the user searches through an ordered sequence of events, dates, or versions to move from an earlier fact to the target fact. It naturally reads as a full question that steps through chronology. Use custom only when none of the above styles accurately describes the phrasing pattern; note the actual style briefly in the scratchpad.
</strategy_selection>

<sub_question_requirements>
Every sub-question must:
- be a complete literal search query with no bracket placeholders, blanks, variables, or stand-ins
- be self-contained and searchable in isolation, with no dependence on prior retrieved results
- explicitly name its searchable subject
- ask for exactly one piece of information (the ending rules define the allowed exception for a final comparison question)
- retrieve a new fact or source needed for the original answer
- be written as a natural search query, not as a task instruction to a researcher
- avoid copying sentences from the reference context
- avoid paraphrasing the original question as an intermediate step
</sub_question_requirements>

<ending_rules>
For non-yes/no questions:
- Do not end with a generic synthesis, decision, or reasoning query.
- End at the last retrieval query that makes the Golden Answer uniquely reachable.
- If the original question asks which candidate wins a comparison and the Golden Answer is a named entity, the final sub-question must be a clean comparison question that makes the model output the winning entity name. It must not include the specific values or facts retrieved earlier.

For yes/no questions:
- The final sub-question must be a clean yes/no comparison question.
- It must not include the specific values, facts, or implied answer retrieved in earlier steps.
- It must require the model to use accumulated context to decide.
</ending_rules>

<pre_return_checks>
Before returning, validate all of the following. If any check fails, rewrite or return empty as stated in the non-negotiable rules above.

Leakage audit:
- Does any sub-question contain the golden answer, an alias, a translation, an abbreviation, or a uniquely identifying description?
- Does any sub-question make the answer obvious without retrieval?

Collapse audit:
- Does every intermediate step leave at least 3 plausible final-answer candidates?
- Does the final step make the original Golden Answer uniquely reachable?
- Is the final answer still exactly the original Golden Answer, not an intermediate entity?
- Did the chain stop immediately after the first collapse to G?

Standalone audit:
- Does any sub-question use bracket placeholders, blanks, or unfilled variables?
- Does any sub-question bundle more than one retrieval target or combine retrieval with comparison?
- Do the sub-questions sound like realistic user searches rather than formulaic decomposition or researcher instructions?
- Does the chain vary in phrasing and structure across steps rather than repeating the same framing?

Strategy audit:
- Is the chosen decomposition strategy a natural fit? If not, reconsider.
- Have you tried more than one plausible strategy before deciding no valid chain exists?
</pre_return_checks>

Return JSON only:
{{"scratchpad": "brief note on why this strategy and style fit; confirm no answer leakage; confirm the final sub-question is the first collapse to the original Golden Answer; for counterfactual_guard also note the likely wrong interpretation; if custom is chosen for either field, name the actual approach used", "decomposition_strategy": "one value from: decomposition, entity_traversal, evidence_first, disambiguation, counterfactual_guard, scope_control, custom", "injection_style": "one value from: direct_instruction, short_followup, research_workflow, user_correction, scope_refinement, distractor_suppression, query_pivot, comparison_check, source_hunt, timeline_trace, custom", "sub_questions": [...]}}
\end{SaveVerbatim}

\begin{SaveVerbatim}{BenchmarkQualityJudgePrompt}
You are a quality judge for sub-question chains used in a multi-hop search QA benchmark.

The evaluated model runs in a retrieval loop. Each sub-question is searched, retrieved passages are added to context, and the next sub-question is injected. Intermediate answers are deleted. The final answer is produced only after all accumulated retrieval context is available.

Your job is not to judge whether the chain is perfect.
Your job is to reject only clear failures that would corrupt training.

Inputs:
Question: {question}
Golden answers: {golden_answers}
Sub-questions: {sub_questions}

Return BAD only for one of the following clear failures.

1. Answer leakage.
A sub-question leaks the golden answer if it explicitly states the golden answer, a direct alias, abbreviation, alternate spelling, translation, nickname, or uniquely identifying label that functions as the answer, unless that same wording already appears in the original question.
Do not flag a sub-question merely because a knowledgeable person could solve it using outside knowledge. Flag only when the sub-question text itself contains the answer, a direct alias, or a uniquely identifying label.
Do not flag a sub-question that asks for the golden answer, such as "what nationality is X?" or "what character did Y voice?" Asking for information is retrieval, not leakage. Leakage is when the answer is stated inside the sub-question text.
Do not flag intermediate entities that are retrieval targets on the path to the answer, unless that entity is itself the golden answer or a direct alias of it.

2. Final collapse mismatch.
The chain must still answer the original question, not a related or intermediate question.

Flag BAD only when it is definitively clear that the final sub-question targets a different answer type or a different final fact than the golden answer. Do not flag when uncertain or when the mismatch is only possible — only flag when it is obvious.

Examples:
- Golden answer is a person, final sub-question asks for a date → BAD.
- Golden answer is a date, final sub-question asks for a person → BAD.
- Golden answer is a place, final sub-question asks for an event title → BAD.
- Golden answer is yes/no, final sub-question is not a yes/no verification or comparison → BAD.
- Original question asks which candidate wins a comparison, golden answer is a named entity, but final sub-question asks only for one candidate's property value → BAD.

Do not require the final sub-question to restate the original question. It only needs to retrieve the last needed fact or perform the required final comparison. For winner-selection questions where the original question asks which of two candidates has a property (born earlier, died later, older, etc.), a final comparison sub-question between those candidates is correct and must not be flagged — even if the comparison does not explicitly restate the film or entity name.

3. Unclear or unusable retrieval target.
Each sub-question must have a clear searchable target.

Flag BAD if a sub-question:
- is only a bare keyword dump with unclear intent;
- contains an unfilled placeholder such as [name], [entity], <person>, or ___;
- depends on an unnamed prior result using wording like "that person", "this film", "the previous result", "his birthplace", or "its release date";
- is written as an instruction to a human researcher rather than a search query;
- asks only for general background with no specific retrieval target.

Short search phrases are allowed if the intended retrieval target is clear.
Conversational follow-ups are allowed only when they still explicitly name the subject being searched.

4. Obvious premature collapse or padding.
Flag BAD only if it is obvious that:
- an intermediate sub-question already uniquely points to the golden answer, making later sub-questions unnecessary; or
- a later sub-question repeats the same retrieval target without adding a new needed fact.

Do not reject merely because the chain could be shorter or phrased more elegantly.

Return GOOD if none of the above clear failures are present.

Decision standard:
Be strict on answer leakage, final-answer mismatch, and unclear retrieval targets.
Be tolerant on style, wording, strategy choice, and whether the chain is the absolute best possible decomposition.

Before deciding, check:
- Does any sub-question reveal the golden answer or a recognizable alias not already present in the original question?
- Does the final sub-question lead to the same answer type and same final fact as the golden answer?
- Does every sub-question have a clear retrieval target that can be searched literally?
- Is there any obvious premature collapse before the final sub-question?

Return JSON only, with exactly one of these forms:
{{"verdict":"GOOD","reason":null}}
{{"verdict":"BAD","reason":"<one sentence naming the specific failure>"}}
\end{SaveVerbatim}

% Callable "variables"
\newcommand{\PrintDecompositionPrompt}{%
\UseVerbatim[
  fontsize=\tiny,
  breaklines=true,
  breakanywhere=true,
  frame=single
]{DecompositionPrompt}
}

\newcommand{\PrintHintPrompt}{%
\UseVerbatim[
  fontsize=\tiny,
  breaklines=true,
  breakanywhere=true,
  frame=single
]{HintPrompt}
}

\newcommand{\PrintDisambiguationPrompt}{%
\UseVerbatim[
  fontsize=\tiny,
  breaklines=true,
  breakanywhere=true,
  frame=single
]{DisambiguationPrompt}
}

\newcommand{\PrintBenchmarkGenerationPrompt}{%
\UseVerbatim[
  fontsize=\tiny,
  breaklines=true,
  breakanywhere=true,
  frame=single
]{BenchmarkGenerationPrompt}
}

\newcommand{\PrintBenchmarkQualityJudgePrompt}{%
\UseVerbatim[
  fontsize=\tiny,
  breaklines=true,
  breakanywhere=true,
  frame=single
]{BenchmarkQualityJudgePrompt}
}
\section{Data Generation Prompts}
\label{sec:appendix_genprompts}

All sub-question chains used for training and evaluation are generated by prompting
GPT-5 with one of the prompts below. The three training prompts
(Sections~\ref{app:decomp}--\ref{app:disambig}) are applied to HotpotQA questions
during the training data pipeline; one strategy is sampled per question. The
benchmark generation prompt (Section~\ref{app:benchmark}) is used to produce the
\textsc{SteerSearch} evaluation benchmark from HotpotQA, 2WikiMultiHopQA, Bamboogle, and
MuSiQue, with ten injection styles and an LLM judge filter.

\onecolumn

\subsection{Decomposition Prompt}
\label{app:decomp}
\PrintDecompositionPrompt

\subsection{Hint Prompt}
\label{app:hint}
\PrintHintPrompt

\subsection{Disambiguation Prompt}
\label{app:disambig}
\PrintDisambiguationPrompt

\subsection{Benchmark Generation Prompt}
\label{app:benchmark}
\PrintBenchmarkGenerationPrompt

\subsection{Benchmark Quality Judge Prompt}
\label{app:judge}
\PrintBenchmarkQualityJudgePrompt

\end{document}